\documentclass[11pt]{article}

\usepackage[final]{acl}

\usepackage[utf8]{inputenc} 
\usepackage[T1]{fontenc}    
\usepackage{times}
\usepackage{latexsym}
\usepackage{url}            
\usepackage{booktabs}       
\usepackage{amsfonts}       
\usepackage{nicefrac}       
\usepackage{microtype}      
\usepackage{xcolor}         
\usepackage{graphicx}
\usepackage{subcaption}
\usepackage{enumitem}
\usepackage{wrapfig}
\usepackage{listings}
\usepackage{amsmath}
\usepackage[export]{adjustbox}

\graphicspath{{./figures/}}

\title{A Survey of LLM Prompt Datasets: Taxonomy, Linguistic Patterns, and Practical Uses}

\author{%
  Yuanming Zhang$^1$\NoHyper\thanks{Equal contribution.}\endNoHyper \quad
  Yan Lin$^2$\NoHyper\footnotemark[1]\endNoHyper \quad
  Arijit Khan$^{3, 2}$\NoHyper\thanks{Corresponding author.}\endNoHyper \quad
  Huaiyu Wan$^1$ \\
  $^1$School of Computer Science and Technology, Beijing Jiaotong University, Beijing, China \\
  $^2$Aalborg University, Aalborg, Denmark \\
  $^3$Bowling Green State University, Ohio, USA \\
  \texttt{\{ymzhang23,hywan\}@bjtu.edu.cn, lyan@cs.aau.dk,} \\
  \texttt{arijitk@bgsu.edu}
}

\begin{document}

\maketitle

\begin{abstract}
    We compile 129 public LLM prompt datasets with more than 1.22\,TB and more than 673M instances and organise them into a unified taxonomy. We use seven datasets for detailed analysis and identify lexical, syntactic, and semantic patterns that distinguish prompts from general text. We evaluate these features in prompt filtering, source domain routing, and elicited response quality analysis. A 63 dimensional linguistic feature set that can be extracted on a CPU achieves over 91\% of the F1 of GPU sentence embeddings and reduces latency per request by nearly half. In cross dataset routing, sentence embeddings reach 0.74 Macro-F1 and a structural feature subset reaches 0.61, compared with the chance level of 0.20. Structural routing features remain negatively associated with elicited response quality after prompt length is controlled on UltraFeedback. The apparent positive association of type to token ratio becomes much smaller under the same control. These findings support prompt processing pipelines that combine efficient structural features with robust semantic routing. Our catalogue, annotations, and code are given at \url{https://github.com/ymzhang-cs/prompt-dataset-analysis}.
\end{abstract}

\section{Introduction}
\label{sec:intro}

Recent advances in large language models (LLMs) have led to many custom prompts across GitHub repositories such as f/awesome-chatgpt-prompts~\citep{awesome-chatgpt-prompts}, Reddit forums such as ChatGPTPromptGenius~\citep{ChatGPTPromptGenius}, commercial platforms such as PromptBase~\citep{PromptBase}, and research releases~\citep{DatabricksBlog2023DollyV2,chen2024huatuogpto1medicalcomplexreasoning}. Few studies examine this growing collection as a whole. The variation in prompt vocabulary, syntax, and semantics across sources and its effects on prompt filtering, source domain routing, and elicited response quality analysis remain unclear.

To address this gap, we compile and analyse 129 diverse prompt datasets with more than 1.22\,TB and more than 673M instances. We derive a hierarchical taxonomy and evaluate the extracted linguistic features in prompt filtering, source domain routing, and elicited response quality analysis. The 63 dimensional linguistic feature set can be extracted on a CPU and recovers more than 91\% of GPU embedding F1. Its latency per request is 1.9 times lower at 3.0\,ms compared with 5.7\,ms. We state separately which findings use the full catalogue, the seven datasets, and the data for each application.

\section{Related Work}
\label{sec:related}

\textbf{Datasets for LLMs.} \citet{liu2024datasets} survey a broad range of LLM datasets and focus on training and model tuning corpora. \citet{zhang2023instruction} and OpenCodeInstruct~\citep{ahmad2025opencodeinstruct} focus on instruction tuning collections. \citet{luo2025instructionquality} organise instruction dataset quality evaluation into human, statistical, model based, and LLM based methods. Work on individual resources includes AuToGen, which generates tool learning examples from domain specific structured data~\citep{zeng2025autogen}, and MSEED, which constructs preference data for safety alignment and multidimensional evaluation~\citep{wang2025mseed}. Adjacent reviews survey capabilities such as mathematical problem solving~\citep{he2025mathsurvey}, and topic analyses map themes in ChatGPT research~\citep{zhang2025chatgpttopics}. These studies evaluate training data, construct particular resources, or map adjacent research; they do not provide a linguistic analysis at multiple levels for prompt datasets at this scale.

\textbf{Prompt engineering tools.} \citet{schulhoff2024prompt} survey prompt engineering techniques. PromptAid~\citep{MDSAHKB25} and PromptLandscape~\citep{DBLP:conf/pacificvis/WangYFDZWZ24} provide visual tools. Promptware~\citep{chen2025promptware} applies software engineering principles, and PromptAgent~\citep{DBLP:conf/iclr/WangLW0LZJXH24} creates prompts for expert tasks. We characterise complete dataset collections, while these studies develop individual prompts or tools.

\section{Prompt Datasets Discovery and Refinement}
\label{sec:datasets}

\textbf{Data discovery guideline.} We target datasets that meet three conditions. First, they contain natural language instructions that describe a task and guide an LLM towards a desired output. Second, they cover domains that range from everyday tasks such as travel planning to professional and specialised tasks such as academic writing, healthcare, and finance. Third, they include several prompt forms, such as single instructions and conversations with multiple turns.

\textbf{Data discovery process.} We collect public datasets from four source types. Searches on Hugging Face~\citep{lhoest-etal-2021-datasets}, Kaggle, Google Dataset Search~\citep{noy2019google}, and Papers with Code yield 60 datasets. We query ``prompt dataset'' and ``instruction following dataset.'' Publications on prompt engineering and dialogue systems at NeurIPS, ICLR, and ICML from 2023 through 2025 contribute 73 datasets. GitHub repositories add 21 datasets. These repositories include {\sf Awesome Instruction Datasets}~\citep{awesome-instruction-datasets}, {\sf Prompt Engineering Guide}~\citep{Saravia_Prompt_Engineering_Guide_2022}, and {\sf LLMDataHub}~\citep{LLMDataHub}. Prompt sharing websites contribute 14 datasets. Examples include {\sf Prompt Genius}~\citep{PromptGenius} and {\sf BoredHumans}~\citep{boredhumans}.

\textbf{Data filtering.} We first remove duplicate catalogue entries. For example, CVQA~\citep{romero2024cvqaculturallydiversemultilingualvisual} appears on Hugging Face and in NeurIPS 2024. We then apply four criteria. Curated datasets must contain at least 1K prompts, and datasets shared by users must contain at least 50. We exclude samples with inconsistent formatting or unclear structure, including illustrative prompts scattered through {\sf Prompt Engineering Guide}. We retain prompts tied to clearly defined tasks and omit collections such as {\sf PersonaHub}~\citep{PersonaHub}, which mostly contains persona descriptions without task instructions. Each dataset must be publicly available under a license that allows research. This process yields 129 prompt datasets for the taxonomy in \S\ref{sec:taxonomy}.

\begin{figure*}[t!]
    \centering
    \includegraphics[width=\textwidth]{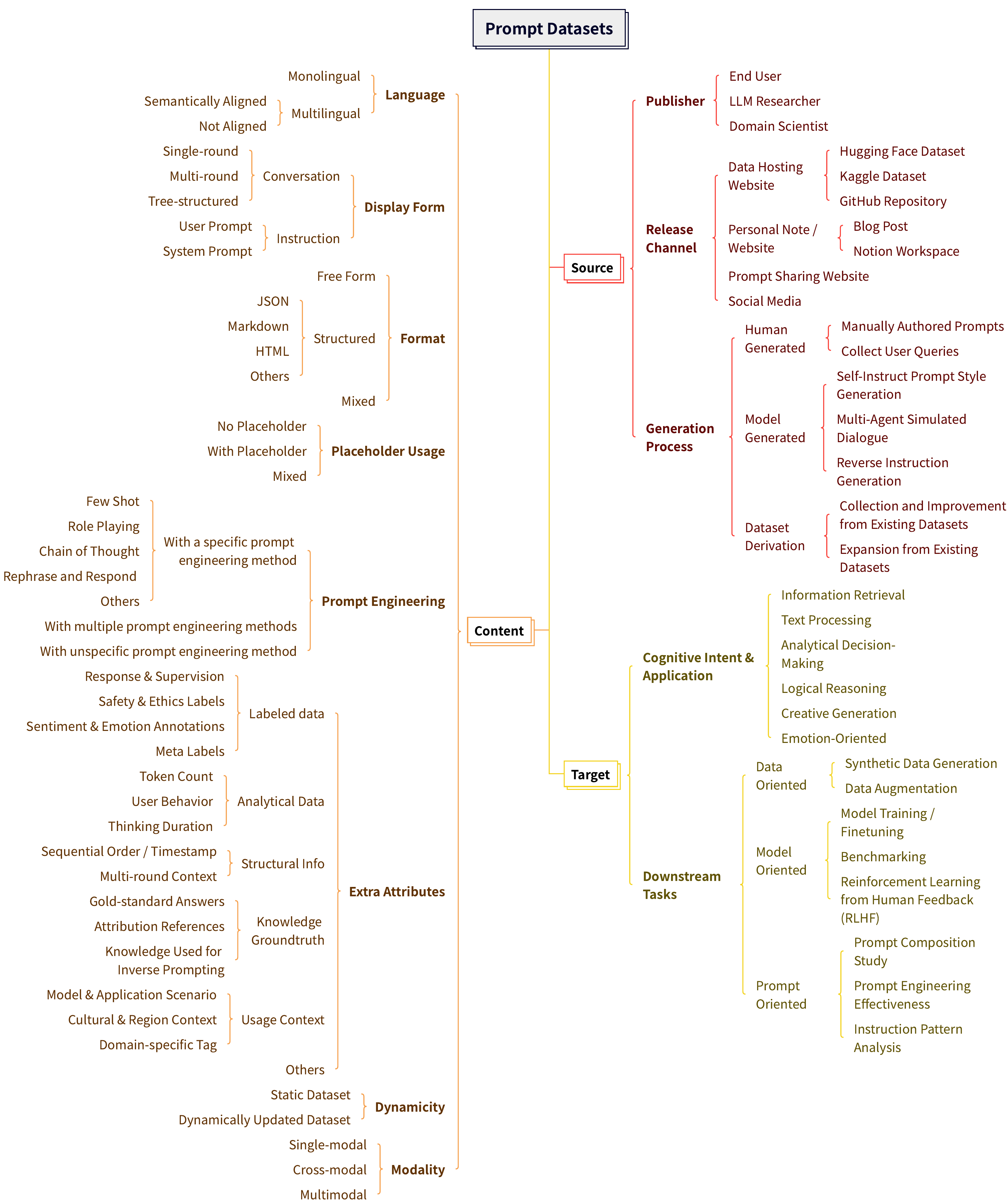}
    \caption{The hierarchical taxonomy of prompt datasets.}
    \vspace{-2mm}
    \label{fig:architecture}
\end{figure*}

\section{Dataset Taxonomy}
\label{sec:taxonomy}

We categorise the prompt datasets along the Source, Content, and Target axes. These axes form the hierarchical taxonomy in Figure~\ref{fig:architecture}. Appendix~\ref{sec:taxonomic-app} gives details for each category.

\textbf{Source} captures publisher, release channel, and generation process. Publishers include {\em end users}, {\em LLM researchers}, and {\em domain scientists}. Examples are {\sf Prompt Genius}, {\sf OpenMathReasoning}, and {\sf ChatGPT Data Science Prompts}~\citep{ChatGPT-Data-Science-Prompts}. Release channels include GitHub, Hugging Face, Kaggle, prompt sharing platforms, and social media. Prompt sharing platforms may provide open or commercial access. Generation processes include human writing, model generation, and dataset derivation. Examples include {\sf databricks-dolly-15k}, Self Instruct~\citep{wang2023selfinstructaligninglanguagemodels}, simulations with several agents~\citep{LAIKG23}, and {\sf Flan 2022}~\citep{Flan}.

\textbf{Content} captures linguistic and structural attributes. Language may be monolingual or multilingual. Display form covers instructions and conversations with one, several, or branching turns. Format may be free form, structured, or mixed, and it can affect response quality~\citep{liu2025promptcontentenhancingllm}. Placeholders support template substitution~\citep{shin-etal-2020-autoprompt}. Prompt engineering methods include few shot prompting~\citep{brown2020languagemodelsfewshotlearners}, role play~\citep{zhang-etal-2018-personalizing}, and chain of thought prompting~\citep{wei2023chainofthoughtpromptingelicitsreasoning}. Other attributes include safety labels, metadata, ground truth, update patterns, and modality. {\sf UltraFeedback} provides one example of additional labels~\citep{cui2023ultrafeedback}.

\textbf{Target} defines purpose and application. Cognitive intents include information retrieval, logical reasoning, creative generation, and tasks that involve emotion. {\sf DeepSeek-Prover-V1} supports logical reasoning~\citep{DeepSeek-Prover}, {\sf No Robots} supports creative generation~\citep{no_robots}, and {\sf empathetic-dialogues} supports emotional tasks~\citep{EmpatheticDialogues}. Downstream tasks focus on data, models, or prompts. Data tasks include synthesis and augmentation. Model tasks include SFT, benchmarking, and RLHF. Instruction tuning collections such as {\sf Alpaca}~\citep{alpaca} and {\sf OASST1}~\citep{KopfKRATSBNSNES23} form a major subset~\citep{zhang2023instruction}.

\textbf{Annotation protocol and reliability.} We assign catalogue labels at the dataset level. Two authors independently annotated all 129 datasets for publisher type, generation process, display form, domain, cognitive intent, and downstream task. They used dataset cards, official documentation, publisher descriptions of collection methods, and sampled prompts. Each scored facet receives one dominant label. Other documented facets permit \textit{Mixed}, multiple, and \textit{Unknown} values when needed. Cohen's $\kappa$ for the six facets is 0.829, 0.813, 0.862, 0.857, 0.606, and 0.838 in the listed order. Cognitive intent is the lowest and lies just above the conventional moderate--substantial boundary. Appendix~\ref{sec:taxonomy-annotation} gives decision rules, agreement details, and catalogue-wide distributions. The agreement study does not cover the remaining design facets.

\section{Prompt Data Analysis}
\label{sec:analysis}

We next analyse prompts from seven sources at the lexical, syntactic, and semantic levels. These sources form a sample for detailed analysis. The sample does not provide a statistical representation of all 129 datasets. We use statistical and machine learning methods to identify composition patterns and variation across sources.

\subsection{Dataset Selection}
\label{sec:selection}

We use three principles to select prompt datasets for analysis across sources. Each dataset uses English and provides accessible raw prompts with wording that supports linguistic analysis. We exclude benchmark prompts such as {\sf PHYBench}~\citep{qiu2025phybenchholisticevaluationphysical}. The selected datasets vary in publisher type, generation process, display form, domain, and target use. Table~\ref{tab:datasets_characteristics} summarises the seven datasets. Appendix~\ref{sec:datasets-descriptions} describes each dataset.

\begin{table*}[t]
    \caption{Key characteristics of seven datasets. Size gives the number of prompts retained after standard preprocessing and quality checks. See Appendix~\ref{sec:datasets-descriptions} for the mapping from native source units to retained prompt counts.}
    \label{tab:datasets_characteristics}
    \scriptsize
    \centering
    \resizebox{\textwidth}{!}{
        \begin{tabular}{@{}lcccccc@{}}
            \toprule
            \textbf{Dataset}       & \textbf{Size} & \textbf{Publisher Type} & \textbf{Generation Method} & \textbf{Display Form}        & \textbf{Domain} \\ \midrule
            \textbf{1.1k-business} & 1235          & End User                & Unknown                    & User Prompt                  & Business        \\
            \textbf{BoredHumans}   & 956           & End User                & Dataset Derivation         & User Prompt                  & General         \\
            \textbf{dolly-15k}     & 14779         & LLM Researcher          & Human Generated            & Single turn Conversation     & General         \\
            \textbf{medical-o1}    & 19679         & Domain Scientist        & Model Generated            & Single turn Conversation     & Medical         \\
            \textbf{OASST1}        & 22079         & LLM Researcher          & Human Generated            & Branching Conversation       & General         \\
            \textbf{Self-Instruct} & 81673         & LLM Researcher          & Model Generated            & Single turn Conversation     & General         \\
            \textbf{ShareGPT}      & 181577        & End User                & Human Generated            & Multiple turn Conversation   & General         \\
            \bottomrule
        \end{tabular}}
\end{table*}

\textbf{Evidence scope and representativeness check.} Claims about scale, coverage, and taxonomy use all 129 datasets. The detailed linguistic results use the seven selected datasets. We test one main group pattern on 11 more prompt sources---nine catalogue datasets and two extension sources---and three general text baselines with the same feature pipeline. We sample up to 400 texts from each source with seed 42 and require at least 10 characters. Mean unknown vocabulary rates measured by the \texttt{X} tag are 0.008 for the seven datasets, 0.020 for the additional prompt datasets, and 0.003 for general text. The distributions for individual datasets overlap. This check therefore supports a group mean pattern and does not show that the seven datasets represent the full catalogue. Appendix~\ref{sec:representativeness-app} gives the protocol and source lists.

\subsection{Token Level Analysis}
\label{sec:token}

We use $n$ gram models to analyse local text patterns~\citep{jurafsky2009speech,cavnar1994n,manning1999foundations}. After lemmatising all tokens, we extract sequences of three, four, and five words. Their frequency distributions show common instruction templates, keyword combinations, and syntactic patterns. Appendix~\ref{sec:token-app} gives the full results.

\textbf{Analysis of results.} Frequent word sequences show differences in domain and prompt engineering. {\sf OASST1} contains role cues such as ``you to act as,'' and {\sf medical-o1} contains reasoning phrases such as ``the most likely diagnosis.'' Sequences of three words capture general commands. Sequences of four and five words capture patterns for specific tasks, such as {\sf ShareGPT}'s ``please write in English language.'' Google Books contains sequences of five words such as ``at the end of the'' and ``in whole or in part.'' Prompt corpora contain more questions and commands, which separates them from narrative and descriptive text.

\subsection{Syntactic Analysis}
\label{sec:syntactic}

We examine prompt structure through dependency parsing~\citep{nivre-2003-efficient}, part of speech (POS) tagging~\citep{10.3115/974499.974526}, and term frequency inverse document frequency (TF-IDF) scoring~\citep{salton1988term}. These features describe the datasets and form vector representations for tasks such as prompt classification.

We compare the prompts with the English Universal Dependencies corpora EWT~\citep{silveira14gold} and ParTUT~\citep{sanguinetti2014partutUD}. EWT contains blogs, social media, reviews, email, and web text. ParTUT contains formal text from legal documents, news, and Wikipedia.

\begin{table*}[t]
    \centering
    \caption{The eight most common dependency types and their dataset proportions. \textbf{punct} marks punctuation, \textbf{prep} marks prepositions, \textbf{det} marks determiners such as ``the'' and ``a,'' \textbf{pobj} marks prepositional objects, \textbf{dobj} marks direct objects, \textbf{nsubj} marks nominal subjects, and \textbf{ROOT} marks the main verb or predicate. The spaCy model \texttt{en\_core\_web\_sm} uses some labels outside the Universal Dependencies standard. A dash marks these labels in comparisons across corpora. Table~\ref{tab:dependency-app} gives the full data.}
    \label{tab:dependency}
    \resizebox{\textwidth}{!}{\begin{tabular}{c|cc|ccccccc}
            \toprule
            \textbf{Dependency Type} & \textbf{EWT}     & \textbf{ParTUT}  & \textbf{1.1k-business} & \textbf{BoredHumans} & \textbf{dolly-15k} & \textbf{medical-o1} & \textbf{OASST1} & \textbf{Self-Instruct} & \textbf{ShareGPT}  \\
            \midrule
            punct                    & \underline{0.12} & \underline{0.12} & 0.1227                 & \textbf{0.1985}      & 0.1445             & 0.1216              & 0.1273          & 0.1863                 & 0.1540             \\
            prep                     & -                & -                & 0.0759                 & \underline{0.0672}   & 0.0866             & \textbf{0.1013}     & 0.0816          & 0.0676                 & 0.0764             \\
            det                      & 0.08             & 0.09             & \underline{0.0518}     & 0.0692               & \textbf{0.0961}    & 0.0906              & 0.0841          & 0.0838                 & 0.0693             \\
            pobj                     & -                & -                & 0.0718                 & \underline{0.0620}   & 0.0817             & \textbf{0.0979}     & 0.0760          & 0.0645                 & 0.0711             \\
            nsubj                    & 0.08             & 0.06             & 0.0596                 & 0.0545               & 0.0650             & \underline{0.0469}  & \textbf{0.0739} & 0.0596                 & 0.0562             \\
            ROOT                     & 0.07             & 0.04             & 0.0528                 & 0.0462               & 0.0768             & 0.0444              & 0.0604          & \textbf{0.0792}        & \underline{0.0437} \\
            amod                     & 0.05             & 0.06             & 0.0573                 & 0.0527               & 0.0469             & \textbf{0.1072}     & 0.0523          & \underline{0.0384}     & 0.0480             \\
            dobj                     & -                & -                & \textbf{0.0904}        & 0.0665               & 0.0447             & \underline{0.0315}  & 0.0594          & 0.0570                 & 0.0519             \\
            \bottomrule
        \end{tabular}}
\end{table*}

\begin{figure*}[t]
    \centering
    \begin{subfigure}[b]{0.33\textwidth}
        \centering
        \includegraphics[width=\linewidth]{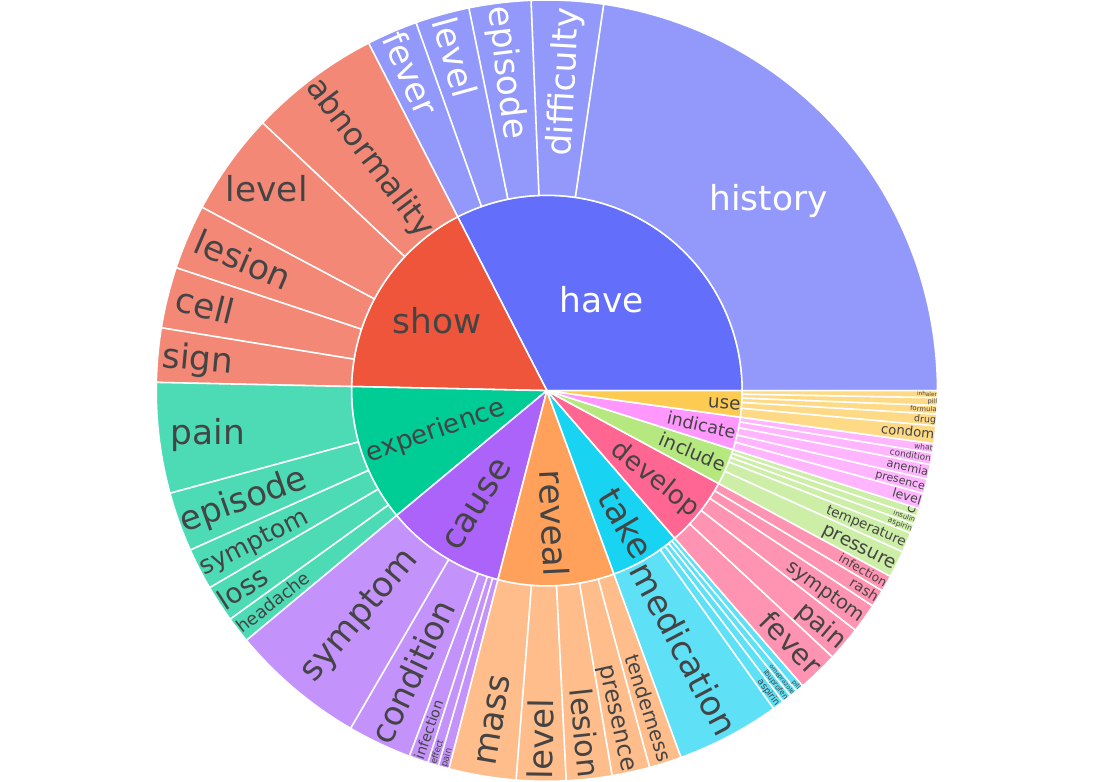}
        \caption{\scriptsize \textsf{medical-o1}}
        \label{fig:sunburst_medical-o1-reasoning-SFT}
    \end{subfigure}
    \begin{subfigure}[b]{0.33\textwidth}
        \centering
        \includegraphics[width=\linewidth]{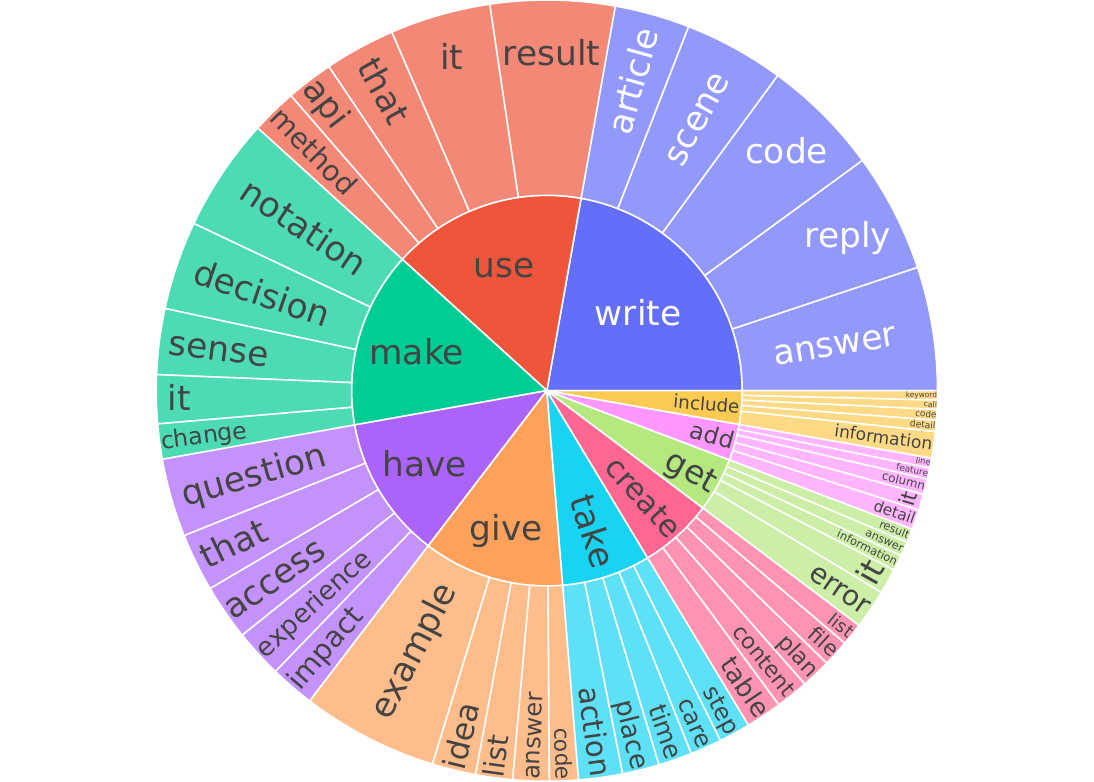}
        \caption{\scriptsize \textsf{ShareGPT}}
        \label{fig:sunburst_sharegpt}
    \end{subfigure}
    \begin{subfigure}[b]{0.32\textwidth}
        \centering
        \includegraphics[width=\linewidth]{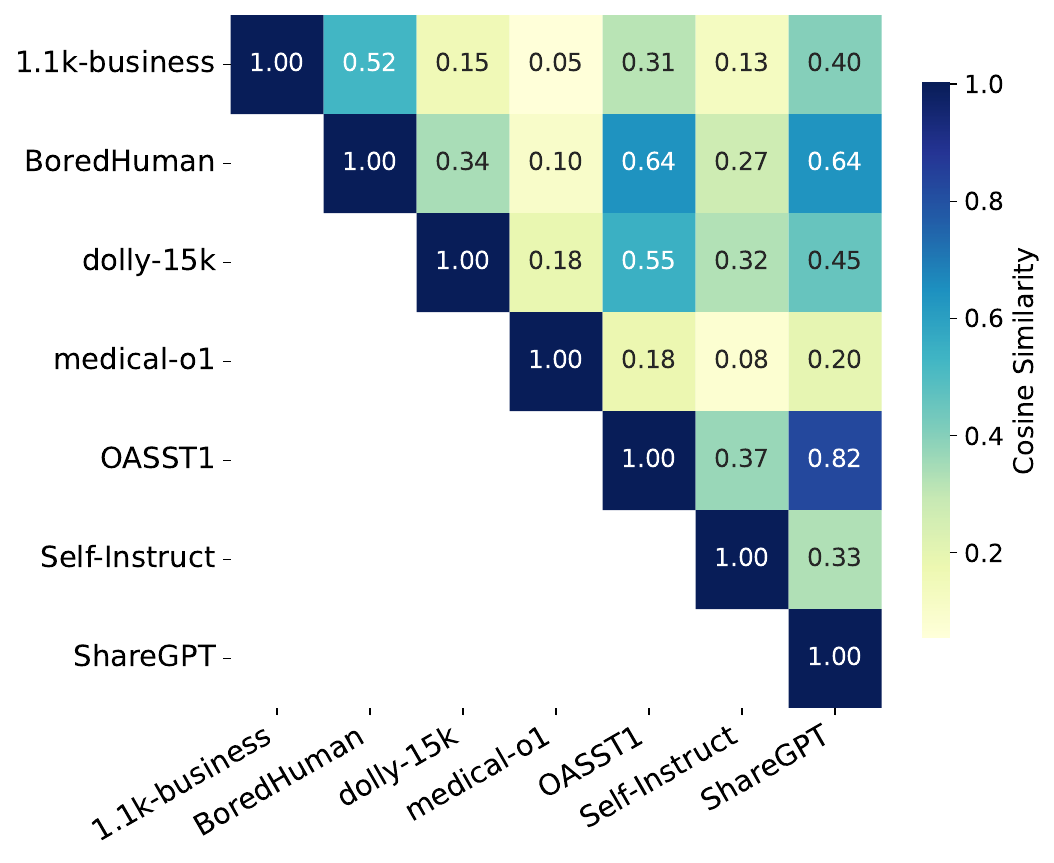}
        \caption{\scriptsize \textsf{TF-IDF}}
        \label{fig:tfidf}
    \end{subfigure}
    \caption{Panels (a) and (b) show the ten most common verbs and their five most common direct noun objects in two prompt datasets. Figure~\ref{fig:verb-noun-app} gives the other five datasets. Panel (c) shows cosine similarity between TF-IDF vectors for each dataset.
    }
    \label{fig:sunburst_main}
\end{figure*}

\subsubsection{Dependency Parsing}
\label{sec:dependency}
We use the spaCy \texttt{en\_core\_web\_sm} parser~\citep{spacy2} to extract syntactic dependencies and measure key grammatical relations in each dataset. We use the official dependency annotations for EWT and ParTUT. The results show consistent differences in linguistic style across prompt sources. We also track verb and object pairs to measure variation across prompt tasks. Figure~\ref{fig:sunburst_main} shows these pairs.

\textbf{Analysis of Results.}
Table~\ref{tab:dependency} shows eight common dependency types across seven prompt datasets and the EWT and ParTUT reference corpora. Three patterns stand out.

\begin{table*}[t]
    \centering
    \caption{The seven most common parts of speech and their dataset proportions. Table~\ref{tab:pos-app} gives the full data.}
    \label{tab:pos_frequency}
    \resizebox{\textwidth}{!}{
        \begin{tabular}{l|cc|ccccccc}
            \toprule
            \textbf{POS} & \textbf{EWT}     & \textbf{ParTUT} & \textbf{1.1k-business} & \textbf{BoredHumans} & \textbf{dolly-15k} & \textbf{medical-o1} & \textbf{OASST1} & \textbf{Self-Instruct} & \textbf{ShareGPT} \\
            \midrule
            NOUN         & \underline{0.17} & 0.21            & \textbf{0.2637}        & 0.2103               & 0.1899             & 0.2590              & 0.1946          & 0.2027                 & 0.1944            \\
            PUNCT        & 0.12             & 0.12            & \underline{0.1094}     & \textbf{0.1942}      & 0.1435             & 0.1158              & 0.1231          & 0.1839                 & 0.1450            \\
            VERB         & 0.11             & 0.10            & \textbf{0.1302}        & 0.1094               & 0.0871             & \underline{0.0775}  & 0.1069          & 0.0999                 & 0.0979            \\
            ADP          & 0.09             & 0.12            & 0.0758                 & \underline{0.0678}   & 0.0858             & \textbf{0.0998}     & 0.0851          & 0.0701                 & 0.0789            \\
            DET          & 0.08             & \textbf{0.11}   & \underline{0.0506}     & 0.0693               & 0.0949             & 0.0893              & 0.0839          & 0.0844                 & 0.0696            \\
            PRON         & 0.09             & 0.04            & \textbf{0.0912}        & 0.0708               & 0.0695             & \underline{0.0369}  & 0.0870          & 0.0701                 & 0.0583            \\
            ADJ          & 0.07             & 0.08            & 0.0588                 & 0.0543               & 0.0538             & \textbf{0.1104}     & 0.0632          & \underline{0.0498}     & 0.0563            \\
            \bottomrule
        \end{tabular}
    }
\end{table*}

First, \textsf{medical-o1} has a high rate of adjectival modifiers at 0.11 and a low rate of direct objects at 0.03. This pattern reflects descriptions of medical conditions and diagnoses that use linking verbs. Second, \textsf{1.1k-business} often uses imperatives that state goals. Its direct object rate is 0.09, and its determiner rate is 0.05. This pattern fits its focus on project planning. Third, verb and noun pairs separate domains. Medical prompts use pairs such as ``have history'' and ``experience pain,'' while \textsf{ShareGPT} uses general pairs such as ``write answer'' and ``use code.''

\subsubsection{Part of Speech Tagging}
\label{sec:pos}
We annotate each dataset with POS tags and measure the distribution of nouns, verbs, adjectives, and adverbs. Table~\ref{tab:pos_frequency} compares content words and function words. \textsf{1.1k-business} and \textsf{medical-o1} have noun proportions near 0.26, which exceed the proportion in the formal ParTUT corpus. These datasets therefore focus heavily on domain entities. \textsf{medical-o1} also has a high adjective ratio of 0.11, which fits the descriptive nature of clinical reasoning tasks.

\subsubsection{TF-IDF Analysis}
\label{sec:tfidf}

We analyse lexical patterns across prompt datasets with TF-IDF. We combine the prompts from each dataset into one document, which produces seven documents. The vectorizer uses at most 5{,}000 words and removes English stopwords. We measure lexical similarity across datasets with pairwise cosine similarity in Figure~\ref{fig:tfidf}. We also extract the three tokens with the highest weights in each dataset. Appendix Table~\ref{tab:tfidf} gives these tokens.

\textbf{Analysis of results.} Each corpus has a distinct lexical focus. \textsf{1.1k-business} emphasises business terms such as ``content'' and ``email.'' \textsf{BoredHumans} emphasises role prompts through words such as ``act.'' \textsf{Self-Instruct} emphasises structural tokens such as ``output'' with a weight of 0.766, which reflects its instruction and response format. Across datasets, \textsf{OASST1} and \textsf{ShareGPT} have the highest cosine similarity because both contain conversations between people and LLMs. \textsf{Self-Instruct} is lexically distant from \textsf{1.1k-business} and \textsf{medical-o1}, which reflects differences in domain and style.

\subsection{Semantic Analysis}
\label{sec:semantic}

We encode each prompt as a dense vector with 384 dimensions by using Sentence-BERT's \texttt{all-MiniLM-L6-v2} model~\citep{reimers-2019-sentence-bert}. We sample 500 prompts from each dataset and project them to two dimensions with PCA. Appendix Figure~\ref{fig:pca-app} shows a scatter plot for each dataset. \textsf{Self-Instruct} has a broad and even semantic distribution, which fits its use of many instruction types. \textsf{medical-o1} and \textsf{1.1k-business} form concentrated clusters for their domains. \textsf{dolly-15k}, \textsf{OASST1}, and \textsf{ShareGPT} overlap because they share styles from interactions between people and LLMs.

K-Means with $K{=}7$ on the 384 dimensional embeddings yields NMI\,=\,0.39, ARI\,=\,0.31, Purity\,=\,0.56, and silhouette\,=\,0.03. These values show real boundaries with substantial overlap. The mean cosine distance to the centroid is 0.50 for \textsf{1.1k-business} and 0.55 for \textsf{medical-o1}, which confirms their concentrated clusters. \textsf{OASST1} and \textsf{ShareGPT} have distances of 0.79, and \textsf{dolly-15k} has a distance of 0.75. These three datasets have the greatest dispersion.

\section{Application}
\label{sec:application}

We use the extracted linguistic features in three applications. Prompt filtering separates LLM prompts from general text. Source domain routing assigns prompts to five source domains. Our feature-based elicited response quality analysis estimates the judged quality of responses elicited by a prompt without LLM inference at prediction time. These applications show that the patterns in \S\ref{sec:analysis} support classification and association tasks. To connect them to the taxonomy, Appendix~\ref{sec:taxonomy-feature-app} groups source-level linguistic measurements by cognitive intent and downstream task. Appendix~\ref{sec:llm-baselines-app} reports zero- and few-shot LLM baselines and inference cost for all three applications.

\subsection{Prompt Filtering}
\label{sec:filtering}

\textbf{Task.} Given a text snippet, the classifier determines whether it is an LLM prompt or general text. General text includes blog posts, Reddit comments, and Wikipedia passages. This task tests whether the structural features in our analysis capture the characteristics of prompts.

\textbf{Setup.} We sample 3{,}000 prompts from seven analysed datasets and 3{,}000 general texts. The general texts contain 1{,}000 examples from each of the blog, Reddit, and Wikipedia corpora. We extract six representations for each text. They are an 18 dimensional POS distribution, a 45 dimensional dependency distribution, their 63 dimensional Syntactic concatenation, a 5{,}000 dimensional TF-IDF vector, a 384 dimensional sentence embedding, and a Combined vector with 5{,}447 dimensions. We fit TF-IDF in each training fold. We train MLP classifiers with five fold stratified cross validation. We also evaluate an Isolation Forest fitted on prompts.

\textbf{Results.} Table~\ref{tab:filtering} reports the results. The Combined representation reaches F1=0.902, and sentence embeddings reach 0.899. The 63 dimensional Syntactic representation reaches F1=0.822, which is over 91\% of embedding F1. It uses $\approx$ 6x fewer dimensions and requires no embedding inference. In our runtime benchmark, CPU linguistic features take 3.0\,ms per request, and GPU embeddings take 5.7\,ms. Section~\ref{sec:app_discussion} discusses this comparison. The Isolation Forest~\citep{liu2008isolation} performs near chance, which shows the value of supervised learning. Appendix Figure~\ref{fig:filtering_overview} gives the training curves and F1 for each feature set. Supervised F1 ranges from 0.789 to 0.902 across the six representations.

\begin{table*}[t]
    \centering
    \caption{Classification of prompts and general text across six feature representations. The table reports the mean and standard deviation from five fold stratified cross validation for the supervised MLP and results for an Isolation Forest baseline. \textbf{Bold} marks the best supervised value for each metric.}
    \label{tab:filtering}
    \scriptsize
    \resizebox{\textwidth}{!}{
        \begin{tabular}{@{}l|cccc|cccc@{}}
            \toprule
                              & \multicolumn{4}{c|}{\textbf{Supervised (MLP)}} & \multicolumn{4}{c}{\textbf{Unsupervised (Isolation Forest)}}                                                                                                                            \\
            \textbf{Feature}  & \textbf{Accuracy}                              & \textbf{Precision}                                           & \textbf{Recall}        & \textbf{F1}            & \textbf{Accuracy} & \textbf{Precision} & \textbf{Recall} & \textbf{F1} \\
            \midrule
            POS (18 dim)       & .789$\pm$.007          & .787$\pm$.007          & .792$\pm$.021          & .789$\pm$.010          & .507 & .505 & .700 & .587 \\
            Dependency (45 dim)& .793$\pm$.017          & .790$\pm$.012          & .798$\pm$.029          & .794$\pm$.019          & .532 & .524 & .700 & .599 \\
            Syntactic (63 dim) & .822$\pm$.012          & .822$\pm$.012          & .822$\pm$.013          & .822$\pm$.012          & .515 & .511 & .700 & .591 \\
            TF-IDF (5k dim)    & .848$\pm$.009          & .843$\pm$.019          & .855$\pm$.029          & .849$\pm$.010          & .561 & .547 & .700 & .614 \\
            Embedding (384 dim)& .900$\pm$.005          & .907$\pm$.002          & .892$\pm$.014          & .899$\pm$.006          & .599 & .582 & .700 & .636 \\
            Combined (5.4k dim)& \textbf{.903$\pm$.008} & \textbf{.912$\pm$.008} & \textbf{.893$\pm$.018} & \textbf{.902$\pm$.009} & .569 & .554 & .700 & .619 \\
            \bottomrule
        \end{tabular}}
\end{table*}

\subsection{Source Domain Routing}
\label{sec:domain}

\textbf{Task.} We use the same feature representations to route prompts into Business, Medical, Coding, Finance, and Creative Writing. This task tests whether linguistic features separate source domains.

\textbf{Datasets.} We use five training sources. \textsf{1.1k-business} covers Business, and \textsf{medical-o1-reasoning-SFT} covers Medical. \textsf{OpenCodeReasoning} is an NVIDIA corpus distilled for competitive programming and covers Coding~\citep{ahmad2025opencodereasoning}. \textsf{finance-alpaca} contains financial instructions and covers Finance. \textsf{WritingPrompts} contains Reddit story prompts and covers Creative Writing~\citep{fan-etal-2018-hierarchical}. Each domain contributes 1{,}200 prompts, for a total of 6{,}000.

\textbf{Evaluation protocol.} We conduct four evaluation phases. Phase A uses five fold stratified cross validation with MLP classifiers. Phase B uses unsupervised K-Means clustering to test whether domain structure appears without labels and supervised Nearest Centroid classification as a simple geometric baseline. We report NMI, ARI, Purity, and accuracy. Phase C uses an 80/20 stratified split. Phase D trains on the five original sources and evaluates transfer to one target collection for each domain. The target collections are the business subset of \textsf{2.5k-templates}, \textsf{UltraMedical}~\citep{zhang2024ultramedical}, \textsf{OpenCodeInstruct}~\citep{ahmad2025opencodeinstruct}, \textsf{FiQA}~\citep{maia2018fiqa}, and the generation subset of \textsf{No Robots}~\citep{no_robots}.

\begin{figure*}[t]
    \centering
    \begin{subfigure}[b]{0.46\textwidth}
        \centering
        \includegraphics[width=\linewidth]{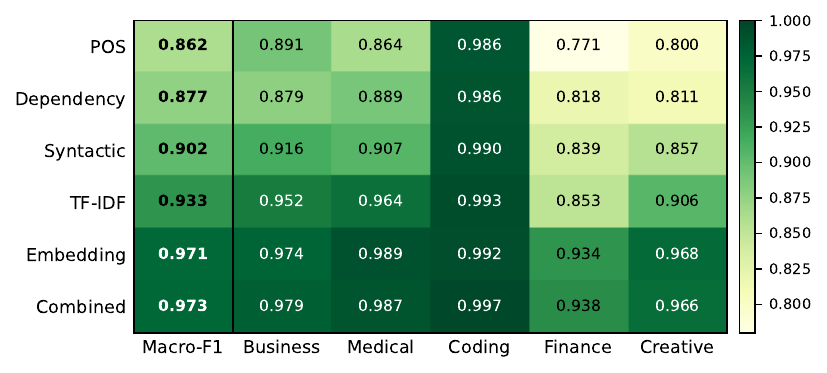}
        \caption{\scriptsize Phase A reports five fold Macro-F1 and F1 for each domain across six feature representations. The first column gives Macro-F1 in bold.}
        \label{fig:domain_phaseA_main}
    \end{subfigure}
    \hfill
    \begin{subfigure}[b]{0.26\textwidth}
        \centering
        \includegraphics[width=\linewidth]{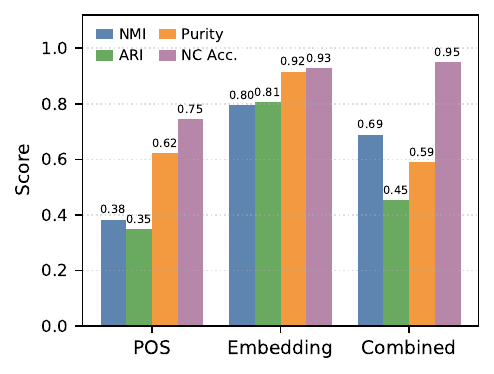}
        \caption{\scriptsize Phase B reports K-Means NMI, ARI, and Purity together with Nearest Centroid accuracy.}
        \label{fig:domain_phaseB_main}
    \end{subfigure}
    \hfill
    \begin{subfigure}[b]{0.26\textwidth}
        \centering
        \includegraphics[width=\linewidth]{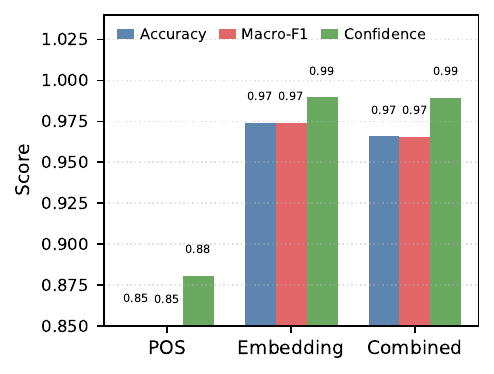}
        \caption{\scriptsize Phase C reports accuracy, Macro-F1, and average maximum softmax confidence on the 20\% test split.}
        \label{fig:domain_phaseC_main}
    \end{subfigure}
    \caption{Source domain routing under three protocols within each source. Panel (a) shows that Coding is recognised most consistently and Finance is generally the hardest domain. Panels (b)-(c) show that sentence embeddings lead the clustering metrics and reach 97.4\% Macro-F1 on the test split. The Appendix gives results for all six feature sets.}
    \label{fig:domain_classification}
\end{figure*}

\textbf{Phase A results.} Figure~\ref{fig:domain_phaseA_main} presents five fold cross validation results. Appendix Figure~\ref{fig:domain_phaseA} gives the diagnostics. The Combined representation reaches 97.3\% Macro-F1, and sentence embeddings reach 97.1\%. The Syntactic representation reaches 90.2\%, and TF-IDF 93.3\%. Coding is recognised most consistently. Finance is generally the hardest domain because its lexical and structural features overlap with general business text.

\textbf{Phase B and C results.} Figures~\ref{fig:domain_phaseB_main} and \ref{fig:domain_phaseC_main} report Phase B and Phase C metrics for three representative feature sets. Appendix Tables~\ref{tab:domain_unsup} and \ref{tab:domain_holdout} and Figure~\ref{fig:domain_unsup_fig} give results for all six feature sets. Sentence embeddings yield K-Means NMI\,=\,0.80 and Purity\,=\,0.92. The Combined representation yields a lower K-Means NMI of 0.69 because of its high dimensionality. On the 20\% test split, sentence embeddings reach 97.4\% Macro-F1. Most remaining errors occur in Finance, as shown in Appendix Figure~\ref{fig:domain_holdout}. Maximum softmax confidence separates 1{,}200 test prompts from the five domains and 1{,}000 general prompts with ROC-AUC\,=\,0.843. Appendix Figure~\ref{fig:domain_routing} gives this result.

\begin{table}[t]
    \centering
    \caption{Phase D source domain transfer. Pooled cross validation uses both collections. Cross dataset evaluation trains on the original collection and tests on the target collection. Chance Macro-F1 is 0.20.}
    \label{tab:domain_cross_dataset}
    \scriptsize
    \resizebox{\columnwidth}{!}{%
    \begin{tabular}{@{}lcc@{}}
        \toprule
        \textbf{Representation} & \textbf{Pooled CV} & \textbf{Cross dataset} \\
        \midrule
        Formatting syntactic (8 dim) & .677 & .329 \\
        Full syntactic (55 dim)       & .913 & .525 \\
        Structural syntactic (47 dim) & .893 & .605 \\
        Embedding (384 dim)            & \textbf{.964} & \textbf{.740} \\
        \bottomrule
    \end{tabular}
    }
\end{table}

\textbf{Phase D results.} All representations exceed chance on the target collections. The decrease from pooled cross validation shows that the representations capture signals specific to each collection. Removing eight formatting features improves syntactic transfer from 0.525 to 0.605. Sentence embeddings transfer best and reach 0.740. Creative recall ranges from 0.10 to 0.26 because the two creative collections contain different writing tasks. Syntactic features provide an efficient option within known distributions. Sentence embeddings provide stronger transfer across collections. Appendix~\ref{sec:domain-cross-app} gives the results for each domain.

\subsection{Feature Importance via Gradient Analysis}
\label{sec:importance}

We apply mean absolute gradient saliency~\citep{simonyan2014deep,baehrens2010explain} to the trained POS, Dependency, and Syntactic MLPs to identify the features that drive classification. For each input feature $x_i$ and output class $c$, the importance score is the expected absolute partial derivative of the class logit with respect to that feature. We average the score over all classes and the training distribution.
\begin{equation}
    \mathrm{Imp}(i) \;=\; \frac{1}{|C|}\sum_{c \in C} \mathbb{E}_{x}\!\left[\left|\frac{\partial\,\mathrm{logit}_c(x)}{\partial x_i}\right|\right].
\end{equation}
Larger values of $\mathrm{Imp}(i)$ mean that small changes to feature $i$ produce larger changes in the model logits. Appendix Table~\ref{tab:importance} gives the five most important features.

Appendix Figures~\ref{fig:importance_filtering} and \ref{fig:importance_domain} show gradient importance for both tasks.

\textbf{Structural signals dominate both tasks.} \texttt{SPACE} represents whitespace tokens, and \texttt{dep} represents unresolved dependency arcs. \texttt{SPACE} leads the POS-only prompt-filtering representation, while \texttt{dep} leads the separate Dependency-only representation. In the 63 dimensional Syntactic prompt-filtering representation, the five largest values are \texttt{nsubj} (.648), \texttt{SPACE} (.640), \texttt{attr} (.622), \texttt{PRON} (.594), and \texttt{ADP} (.561); \texttt{dep} is not among its five leading features. For source domain routing, \texttt{SPACE} and \texttt{dep} remain leading signals in the Syntactic and Dependency-only representations, respectively. \texttt{SPACE} captures numbered steps, code blocks, and bullet lists that are less common in linear prose. \texttt{dep} captures technical noun phrases and imperative fragments. Prompt filtering also uses an imperative and conversational register that is less common in general text.

\textbf{Domain lexical features and practical rules.} Source domain routing relies heavily on \texttt{X}, which represents unknown vocabulary tokens such as code identifiers, medical abbreviations, and financial tickers. Its importance is 0.87 in the POS-only representation. The \texttt{NUM} and \texttt{nummod} features identify quantitative content in Finance and Medical prompts. The \texttt{compound} feature captures domain terms such as \emph{blood pressure} and \emph{cash flow}. The \texttt{ROOT} feature captures action verbs such as \emph{write} and \emph{implement} in Coding, \emph{diagnose} and \emph{explain} in Medical, and \emph{calculate} in Finance. These results suggest three rules that do not require a trained model. High rates of \texttt{PRON}, \texttt{ADV}, \texttt{nsubj}, and \texttt{attr} help identify prompt text. The \texttt{X} tag ratio helps identify a domain without labels. The combined density of \texttt{NUM} and \texttt{nummod} separates quantitative and qualitative domains.

\subsection{Elicited Response Quality}
\label{sec:quality}

The prior experiments show that linguistic features identify the type and source domain of a prompt. We next test whether the same features correlate with the judged quality of responses elicited by a prompt. This target depends on the response models, task difficulty, and judge.

\textbf{Setup.} We use \textsf{UltraFeedback}~\citep{cui2023ultrafeedback}. Four LLMs answer each prompt, and GPT-4 rates every response from 1 to 10 with the \texttt{overall\_score} field. We average the four response scores for each prompt. We then draw a stratified sample of 10{,}000 unique prompts without adding annotations. The final scores range from 1.25 to 9.38 and have a mean of 6.41. We extract 55 linguistic features that contain 17 POS rates, 27 dependency rates, and 11 basic statistics. We report Spearman $\rho$ with Benjamini and Hochberg FDR at $\alpha{=}0.05$~\citep{benjamini1995controlling}. We also report Cohen's $d$ for Q4 compared with Q1~\citep{cohen1988statistical} and logistic regression AUC for Q1 compared with Q4 under five fold cross validation. We test judge dependence on 2{,}000 \textsf{HelpSteer2} prompts with human helpfulness ratings~\citep{wang2024helpsteer2}. For both datasets, we compute partial Spearman correlations that control for prompt word count.

\begin{table}[t]
    \centering
    \caption{Selected correlations with elicited response quality before length control on 10{,}000 UltraFeedback prompts. The columns report Spearman~$\rho$ and Cohen's~$d$ for Q4 compared with Q1.}
    \label{tab:quality_corr}
    \scriptsize
    \resizebox{\linewidth}{!}{%
        \begin{tabular}{@{}lrr@{}}
            \toprule
            \textbf{Feature}                   & $\rho$           & $d$              \\
            \midrule
            \multicolumn{3}{l}{\emph{Positive correlations}}           \\
            Lexical diversity                  & $+$.257          & $+$.721          \\
            \texttt{ADP} (preposition)         & $+$.168          & $+$.397          \\
            \midrule
            \multicolumn{3}{l}{\emph{Negative correlations}}            \\
            \texttt{dep} (syntactic ambiguity) & $-$.320          & $-$.678          \\
            \texttt{appos} (appositive)        & $-$.304          & $-$.598          \\
            \texttt{SPACE} (formatting tokens) & $-$.308          & $-$.607          \\
            Sentence count                     & $-$.294          & $-$.445          \\
            Word count                         & $-$.270          & $-$.379          \\
            \texttt{X} (OOV/technical tokens)  & $-$.248          & $-$.388          \\
            \texttt{NUM} (numerals)            & $-$.233          & $-$.384          \\
            \texttt{PROPN} (proper nouns)      & $-$.221          & $-$.448          \\
            \bottomrule
        \end{tabular}}
\end{table}

\textbf{Results before length control.} Table~\ref{tab:quality_corr} reports selected associations. After Benjamini and Hochberg FDR at $\alpha{=}0.05$, 49 of 55 features remain significant. Within UltraFeedback, prediction of Q1 compared with Q4 reaches AUC 0.799 with all features and 0.773 with Dependency alone. HelpSteer2 human ratings show the same directions before length control. Lexical diversity has $\rho{=}+.150$. The correlations for \texttt{dep}, \texttt{SPACE}, \texttt{NUM}, and \texttt{X} are $-.128$, $-.127$, $-.172$, and $-.077$.

\textbf{Prompt length confounds lexical diversity.} The reported diversity feature is type to token ratio, which depends on length~\citep{mcCarthy2010mtld}. After word count is controlled, its association changes from $+.257$ to $+.024$ in UltraFeedback and from $+.150$ to $-.017$ in HelpSteer2. In UltraFeedback, the routing features \texttt{dep}, \texttt{SPACE}, \texttt{X}, and \texttt{NUM} remain negative. Their partial correlations are $-.212$, $-.197$, $-.153$, and $-.136$. In HelpSteer2, only \texttt{NUM} remains negative after length control. We interpret these findings as observational associations that depend on the dataset and judge. They provide no evidence that lexical diversity causes better elicited responses. Appendix~\ref{sec:quality-robustness-app} gives the full robustness results.

\subsection{Discussion}
\label{sec:app_discussion}

The 63 dimensional linguistic features reach more than 91\% of GPU embedding F1. In our benchmark, their latency per request is 3.0\,ms compared with 5.7\,ms for embeddings. They also require no corpus vocabulary and support gradient importance analysis. The separate 55 dimensional linguistic representation used in Phase D transfers less reliably than sentence embeddings across target collections. Gradient importance identifies representation-specific structural signals: \texttt{nsubj} and \texttt{SPACE} lead the Syntactic prompt-filtering model, while \texttt{dep} leads the Dependency-only model and is important for routing. Several routing features remain negatively associated with UltraFeedback scores after length control. The positive association for type to token ratio becomes negligible after the same control. These results support a two stage pipeline that uses efficient structural features for known traffic and semantic features or more processing when the distribution changes.

\textbf{Comparison with LLM inference.} On matched but non-paired samples, zero- and few-shot \texttt{gpt-4o-mini} is slightly stronger for filtering (F1 .919--.943 vs.\ .902) and cross-dataset routing (Macro-F1 .921--.924 vs.\ .740), but our linguistic classifier is stronger for response-quality quartile prediction (AUC .799 vs.\ .457--.479; LLM $\rho\approx0$). LLM inference also takes 1.54--1.60\,s per prompt, 270--510 times the 3.0--5.7\,ms local pipelines, and incurs per-query API cost. Thus our approach offers competitive filtering, better within-dataset quality discrimination, interpretability, and substantially lower latency; LLM inference is most useful for semantic routing (Appendix~\ref{sec:llm-baselines-app}, Table~\ref{tab:llm-baselines}).

\section{Conclusion}
\label{sec:conclusion}

We compiled 129 LLM prompt datasets into a structured taxonomy and measured annotation reliability at the dataset level. We analysed seven diverse datasets in detail. The 63 dimensional linguistic features provide efficient and interpretable signals within known distributions. Sentence embeddings transfer more reliably across target collections. In UltraFeedback, several structural routing features remain negatively associated with elicited response quality after length control. The association between type to token ratio and quality becomes much smaller after the same control. These findings support pipelines that combine efficient structural processing, semantic routing, and checks for changes in the data distribution.

\newpage
\section*{Limitations}
Our corpus contains mainly English prompts from Western platforms. The reported linguistic patterns may therefore reflect technically skilled users who write in English. Cross dataset routing uses two collections for each domain and transfers weakly between the two Creative tasks. The elicited response quality study is observational. UltraFeedback inherits biases from its GPT-4 judge~\citep{liu2024lost}, HelpSteer2 uses one protocol for human ratings, and prompt length confounds type to token ratio. Taxonomy labels apply at the dataset level. Dominant labels and descriptions of mixed attributes preserve major forms of variation but omit some differences within a dataset. Appendix~\ref{sec:limitation} discusses classifier scope and licensing constraints.

\section*{Acknowledgments}
Arijit Khan acknowledges support from the Novo Nordisk Foundation Grant NNF22OC0072415.

\bibliography{references}

\newpage
\appendix

\section{Limitations and Discussion}
\label{sec:limitation}

\paragraph{Dataset coverage and language bias.}
Our survey covers 129 sources with more than 1.22\,TB and more than 673M instances. Most prompts are in English and come from Western platforms such as GitHub, Reddit, Hugging Face, and PromptBase. The catalogue has less coverage of prompts in other languages, private enterprise prompts, and traffic from commercial assistants such as ChatGPT and Claude. The reported linguistic patterns may therefore reflect technically skilled users who write in English.

\paragraph{Taxonomy scope and residual heterogeneity.}
Two independent annotators assign the taxonomy in \S\ref{sec:taxonomy} at the dataset level. They use source documents and sampled prompts. Five of the six scored facets have almost perfect agreement. Cognitive intent has the lowest agreement at $\kappa=.606$, just above the conventional moderate--substantial boundary. Broad or mixed collections still have uncertain category boundaries. For documented but unscored attributes, mixed and multiple descriptions preserve supported categories but omit some differences between prompts. Future work can extend the taxonomy to prompt-level annotation within the catalogue instances.

\paragraph{Elicited response quality depends on data and judges.}
Our primary analysis in \S\ref{sec:quality} uses \textsf{UltraFeedback} with GPT-4 as the reference judge. GPT-4 has style and length biases~\citep{liu2024lost}. HelpSteer2 adds human helpfulness ratings from one response set and one rating protocol. Type to token ratio depends on length. Its positive association becomes negligible after prompt word count is controlled. The reported AUC and associations describe these data and judging settings. They do not define a universal or intrinsic measure of elicited response quality.

\paragraph{Association and causal interpretation.}
Gradient importance in \S\ref{sec:importance}, Spearman correlations, and Cohen's~$d$ report associations between linguistic features and classifier decisions or elicited response quality. These results provide no evidence that removing a feature such as whitespace tokens would improve an answer. Length control addresses one measured confound. Prompt topic, task difficulty, user expertise, response model capability, and judge preferences may also affect the results. Causal interpretation requires controlled rewriting experiments.

\paragraph{Classifier scope.}
We evaluate prompt filtering as a binary task that separates prompts and general text. We evaluate source domain routing on balanced samples from Business, Medical, Coding, Finance, and Creative Writing. The transfer study uses two collections for each domain. The two Creative collections represent different writing tasks. Macro-F1 may decrease on the full taxonomy or highly imbalanced traffic. The linguistic pipeline may also become less competitive as the number of classes increases.

\paragraph{Licensing and redistribution constraints.}
Several sources have unclear or restrictive licenses. Our release contains catalogue metadata, taxonomy annotations, processed statistics, and code. It excludes raw prompts when the source terms prohibit redistribution. Researchers who reproduce token level statistics for these sources must download the data from the original hosts.

\paragraph{Prompt effect evaluation.}
The selected datasets cover diverse tasks, so this study does not evaluate causal prompt effects by generating and comparing new task responses. The LLM baselines in Appendix~\ref{sec:llm-baselines-app} instead predict task labels or existing response scores from prompt text. Future work could run the same prompts through several LLMs and compare their task responses. Prompt data continues to grow, and repeated analyses could track new patterns that affect prompt design.

\section{LLM Usage}
\label{sec:llm-usage}

LLMs enter this work in the following roles.

\paragraph{LLMs as analysis objects.}
Some analysed datasets contain prompts generated by LLMs. \textsf{Self-Instruct}~\citep{wang2023selfinstructaligninglanguagemodels} expands human seeds with GPT-3. \textsf{medical-o1-reasoning-SFT}~\citep{chen2024huatuogpto1medicalcomplexreasoning} contains medical questions reformulated by GPT-4o. We treat these datasets as observed data and generate no new training prompts or task responses for the linguistic and elicited-response-quality analyses.

\paragraph{Existing labels in the quality analysis.}
The elicited response quality study in \S\ref{sec:quality} uses scores released with \textsf{UltraFeedback}~\citep{cui2023ultrafeedback} and human helpfulness ratings released with \textsf{HelpSteer2}~\citep{wang2024helpsteer2}. We use both label sets as released and add no LLM annotations. The statistical analysis uses Spearman $\rho$, partial correlations, Benjamini and Hochberg FDR, Cohen's $d$, and logistic regression AUC.

\paragraph{Existing encoders in the feature pipeline.}
The semantic analysis in \S\ref{sec:semantic} and the Embedding representation in \S\ref{sec:filtering} and \S\ref{sec:domain} use the frozen \textsf{Sentence-BERT all-MiniLM-L6-v2} encoder~\citep{reimers-2019-sentence-bert}. The encoder maps each prompt to a vector with 384 dimensions. We use it as a fixed feature extractor and perform no model tuning, prompting, or generation. We use spaCy in the same way to extract POS and dependency tags.

\paragraph{LLMs as evaluation baselines.}
The zero- and few-shot baselines in Appendix~\ref{sec:llm-baselines-app} query \texttt{openai/gpt-4o-mini} to predict prompt-filtering labels, source-domain labels, or an existing elicited-response score from prompt text. These calls generate task-level labels or scores for evaluation; they do not generate answers to the source prompts or add examples to the analysed datasets.

\paragraph{During paper preparation.}
The authors used LLM assistants for English editing, code completion, and checks of the organisation and consistency of analysis files. The authors verified all experimental designs, numerical results, citations, interpretations, and final text. The authors remain responsible for the paper.

\section{Ethics Statement}
\label{sec:ethics-statement}

We compile and analyse existing prompt datasets and collect no new sensitive data. We document the datasets, respect their sources and licenses, and release code and derived artefacts for review and reproduction. The work analyses existing data and releases no new generative technology. We document the methods to support external review.

\paragraph{Personally identifying information and offensive content.}
Several analysed datasets, including \textsf{ShareGPT90K} and \textsf{OASST1}, come from real user interactions. They may contain personally identifying information or offensive content even though users shared them on public platforms. We do not redistribute raw prompts from users. The released artefacts contain catalogue metadata, annotations, code, aggregate statistics, and derived results. This work does not provide a full content audit of the source corpora. Users who download the raw data should screen it for identifying information and harmful content based on their use case.

\section{Reproducibility Statement}
\label{sec:reproducibility-statement}

The repository linked in the abstract contains the code, data processing instructions, and experiment commands. It also specifies dependency versions. The datasets are public or have documented sources. We set random seeds when the method permits. These resources support reproduction of the analyses and results.

\section{Compute Resources}
\label{sec:compute}

All local experiments can run on one workstation and require no cluster or distributed training. The zero- and few-shot LLM baselines additionally require access to the external OpenRouter API; their end-to-end latency, token counts, and estimated cost are reported separately.

\paragraph{Compute workers.} Linguistic feature extraction, TF-IDF and scikit-learn baselines, and statistical analyses run on a standard CPU with several cores. Sentence-BERT embedding and the PyTorch MLP experiments run on one consumer GPU. The encoder can also run in CPU batches at the reported scale of at most 10{,}000 prompts for each feature extraction job.

\paragraph{Memory use.} The largest feature objects are the sentence embedding matrix with $n{\times}384$ values and the TF-IDF representation with 5{,}000 dimensions. The sample size is at most $10^4$. Both objects fit in workstation memory at this scale.

\paragraph{Runtime.} The application experiments can run on the workstation described above. Runtime depends on data loading, feature caching, and hardware. Embedding extraction and repeated MLP folds use most of the computation.

\paragraph{Execution details.} The repository at \url{https://github.com/ymzhang-cs/prompt-dataset-analysis} provides commands, random seeds, and hyperparameters for each experiment.

\section{Broader Impacts}
\label{sec:broader-impacts}

We survey and analyse public LLM prompt datasets and train compact classifiers with spaCy, TF-IDF, and Sentence-BERT features. We release no generative foundation model, image generator, or new scraped corpus. The following paragraphs describe potential benefits and risks.

\paragraph{Positive impacts.} The Syntactic representation used in prompt filtering and Phases A through C of source domain routing has 63 dimensions and reaches over 91\% of embedding F1 for prompt filtering. Its features can be extracted on a CPU and inspected through gradient importance. This pipeline may reduce GPU cost, latency, and energy use in systems that process requests as they arrive. It may also make prompt research more accessible to groups with limited compute. The released taxonomy, statistics for each dataset, and code provide a reference for future studies and audits of changes in prompt distributions.

\paragraph{Potential negative impacts.} Formatting density, unresolved dependency arcs, and technical tokens correlate with writing style. Classifiers trained on these features may penalise prompts from people who learned English as another language, new users, or users of assistive technologies. Type to token ratio also depends on prompt length and should not be treated as an intrinsic quality measure. A deployment should include fairness audits across relevant user groups. The reported structural signals could also help attackers craft prompt injection or jailbreak inputs that match expected statistics. Practitioners should combine these features with other signals and avoid using them as a sole detector.

\paragraph{Mitigations.} Downstream users should audit performance across relevant user groups before deployment. They should combine structural classifiers with safety models that inspect content. They should also monitor calibration as prompt distributions change.

\section{Datasets}
\label{sec:taxonomic-app}

We collected 129 prompt datasets from four source channels. They are dataset platforms, publications at NeurIPS, ICLR, and ICML from 2023 through 2025, GitHub repositories, and prompt sharing websites. Dataset platforms include Hugging Face, Kaggle, Google Dataset Search, and Papers with Code. The complete catalogue and taxonomy annotations are at \url{https://github.com/ymzhang-cs/prompt-dataset-analysis}. Each catalogue entry provides its \texttt{Publisher}, \texttt{Size}, \texttt{License}, \texttt{Link}, and \texttt{Description}. The license labels reflect rights declared by the publishers when we collected the data. Original license terms still apply to collections included within another dataset.

\paragraph{Source units and analysed prompts.}
Catalogue cards report the native source unit, whereas Table~\ref{tab:datasets_characteristics} reports prompts retained after preprocessing. BoredHumans has 964 source instances and 956 retained prompts. Dolly has 15,011 source records and 14,779 retained prompts. The medical-o1 release reports 90,120 records across its configurations; selection of the analysed English prompt view, deduplication, and quality checks yield 19,679 retained prompts. For Self-Instruct, the analysed self-generated configuration has 82,612 source records and 81,673 retained prompts; the separate 52,445-example P3 configuration is not analysed. ShareGPT has 90,665 source threads, which expand to 226,625 user turns and yield 181,577 retained prompts. OASST1 has 161,443 messages; the 84,437/4,401 train/validation rows include both roles, while our user-role view contains 55,507 messages, 22,850 English candidates, and 22,079 retained prompts. The released catalogue records both native and retained units for these six cases.

\subsection{Taxonomy Annotation Protocol}
\label{sec:taxonomy-annotation}

\paragraph{Unit and evidence.}
Each annotation applies to one dataset. Two authors independently labelled all 129 catalogue entries. They used the dataset card, official documents, the collection method stated by the publisher, and a fixed sample of prompts. The six scored facets cover all three taxonomy dimensions. Source includes publisher type and generation process. Content includes display form and domain. Target includes cognitive intent and downstream task. We document language and modality without $\kappa$ because their labels are nearly deterministic. We exclude format, placeholder use, prompt engineering method, and extra attributes from the agreement statistics because they require more inspection at the source or prompt level. The released supplement records two additional dataset-level operational labels, RLHF use and dynamicity, but we do not retroactively include them in the reported six-facet agreement study.

\paragraph{Ambiguity rules.}
Annotators record a dominant value only when the evidence supports it. Generation process permits \textit{Unknown}. Format and placeholder use permit \textit{Mixed}. Prompt engineering method permits multiple or unspecified values. When source inspection provides enough evidence, the catalogue descriptions preserve supported parts of mixed and multiple attributes; the release does not claim exhaustive structured labels for these unscored attributes. For example, \textsf{BoredHumans} combines role play and chain of thought methods. It contains prompts with and without placeholders. \textsf{2.5k-templates} also combines templates with slots and prompts without slots. We do not convert remaining variation within a dataset into unsupported labels for individual prompts.

\begin{table}[t]
    \centering
    \caption{Independent taxonomy agreement at the dataset level across all 129 datasets.}
    \label{tab:taxonomy-kappa}
    \scriptsize
    \begin{tabular}{@{}llc@{}}
        \toprule
        \textbf{Dimension} & \textbf{Facet} & \textbf{Cohen's $\kappa$} \\
        \midrule
        Source  & Publisher type     & .829 \\
        Source  & Generation process & .813 \\
        Content & Display form       & .862 \\
        Content & Domain             & .857 \\
        Target  & Cognitive intent   & .606 \\
        Target  & Downstream task    & .838 \\
        \bottomrule
    \end{tabular}
\end{table}

Five facets reach almost perfect agreement. Cognitive intent is the lowest at $\kappa=.606$, only .006 above the conventional moderate--substantial boundary. Both annotation files contain values in all 774 cells and contain no invalid labels. The exact match across all six facets is 62 of 129, or 0.481. Cognitive-intent disagreements occur mainly in broad or multi-purpose prompt libraries whose prompts genuinely span overlapping intents, rather than indicating a mechanically identifiable annotator error. Other disagreements occur at the boundary between prompt instructions and records that contain one instruction and one response. The released repository includes confusion matrices for each facet and both annotation files.

\begin{table*}[t]
    \centering
    \caption{Catalogue-wide dataset counts for the six scored taxonomy facets and two review-requested auxiliary facets. Percentages use all 129 datasets as denominator and may not sum to exactly 100 because of rounding.}
    \label{tab:taxonomy-aggregate}
    \scriptsize
    \begin{tabular}{@{}ll@{}}
        \toprule
        \textbf{Facet} & \textbf{Count (percentage)} \\
        \midrule
        Publisher type & LLM researcher 81 (62.8\%); end user 29 (22.5\%); domain scientist 19 (14.7\%) \\
        Generation process & human 52 (40.3\%); model 41 (31.8\%); derivative 35 (27.1\%); unknown 1 (0.8\%) \\
        Display form & single-round 84 (65.1\%); instruction 28 (21.7\%); multi-round 15 (11.6\%); tree 2 (1.6\%) \\
        Domain & general 82 (63.6\%); other 30 (23.3\%); medical 7 (5.4\%); coding 5 (3.9\%); business 3 (2.3\%); creative 1 (0.8\%); finance 1 (0.8\%) \\
        Cognitive intent & text processing 64 (49.6\%); reasoning 35 (27.1\%); retrieval 14 (10.9\%); analytical 10 (7.8\%); creative 6 (4.7\%); emotion 0 (0.0\%) \\
        Downstream task & model-oriented 103 (79.8\%); prompt-oriented 26 (20.2\%); data-oriented 0 (0.0\%) \\
        RLHF use (auxiliary) & no 113 (87.6\%); yes 16 (12.4\%) \\
        Dynamicity (auxiliary) & static 118 (91.5\%); dynamic 11 (8.5\%) \\
        \bottomrule
    \end{tabular}
\end{table*}

\noindent\textit{Zero-count categories.}
The zero counts for Emotion and Data-oriented mean that no catalogue dataset receives either category as its dominant dataset-level label. They do not imply that mixed-purpose collections contain no emotion-related prompts or that no collection supports data synthesis or augmentation as a secondary use.

For the auxiliary RLHF label, \textit{yes} includes datasets intended for reinforcement learning and datasets with preference, DPO, rating, reward-model, or verifier signals suitable for preference optimisation. For dynamicity, \textit{dynamic} denotes a living collection or platform that accepts or advertises continuing updates; a versioned snapshot is \textit{static}. The released CSV exposes every decision, and the statistics script reconstructs Table~\ref{tab:taxonomy-aggregate}.

\subsection{Taxonomy-organised Linguistic Check}
\label{sec:taxonomy-feature-app}

We connect the taxonomy to the bounded linguistic extension by joining the source-level means with catalogue labels. The analysis contains 16 catalogue sources: the seven analysed datasets and nine additional catalogue sources. We exclude the two extension sources that are not catalogue entries and the three general-text controls. Each source receives equal weight, so a large corpus does not dominate a category mean.

\begin{table}[t]
    \centering
    \caption{Descriptive source-level linguistic means grouped by taxonomy labels. Small groups are not population estimates.}
    \label{tab:taxonomy-feature-groups}
    \scriptsize
    \resizebox{\columnwidth}{!}{%
    \begin{tabular}{@{}llrrrrr@{}}
        \toprule
        \textbf{Facet} & \textbf{Label} & \textbf{$n$ src.} & \textbf{SPACE} & \textbf{X} & \textbf{TTR} & \textbf{dep} \\
        \midrule
        Cognitive intent & Analytical      & 2  & .000 & .035 & .862 & .015 \\
                         & Reasoning       & 2  & .048 & .005 & .622 & .051 \\
                         & Text processing & 12 & .019 & .016 & .828 & .023 \\
        \midrule
        Downstream task  & Model-oriented  & 7  & .031 & .003 & .802 & .035 \\
                         & Prompt-oriented & 9  & .012 & .027 & .811 & .018 \\
        \bottomrule
    \end{tabular}}
\end{table}

The higher formatting and unresolved-arc rates in the two reasoning sources are driven by medical and coding reasoning formats. Model-oriented sources show the same direction relative to prompt-oriented libraries. These values demonstrate how the taxonomy can organise the linguistic measurements, but the small and unbalanced source groups do not establish a catalogue-wide difference.

\subsection{Bounded Representativeness Check}
\label{sec:representativeness-app}

We compare the seven analysed datasets with 11 more prompt sources and three general text baselines. Nine of the additional sources are in the current 129-dataset catalogue; \textsf{finance-alpaca} and \textsf{WritingPrompts} are explicitly marked extension sources rather than catalogue entries. The additional sources are \textsf{2.5k-templates}, \textsf{awesome-chatgpt}, \textsf{chatgpt-cheatsheet}, \textsf{chatgpt-datascience}, \textsf{finance-alpaca}, \textsf{lamini}, \textsf{OpenCodeReasoning}, \textsf{prompt-genius}, \textsf{prompt-hackers}, \textsf{the-prompt-index}, and \textsf{WritingPrompts}. The general text baselines are blog, Reddit, and Wikipedia. We sample up to 400 texts from each source with seed 42 and retain texts with at least 10 characters. We apply the same spaCy \texttt{en\_core\_web\_sm} feature pipeline as in the main analysis.

\begin{table}[t]
    \centering
    \caption{Group means in the bounded representativeness check. Type to token ratio is included as a description and depends on length.}
    \label{tab:representativeness}
    \scriptsize
    \begin{tabular}{@{}lrr@{}}
        \toprule
        \textbf{Group} & \textbf{OOV/\texttt{X}} & \textbf{TTR} \\
        \midrule
        Seven analysed prompt datasets    & .008 & .877 \\
        Eleven additional prompt datasets & .020 & .776 \\
        General text baselines             & .003 & .705 \\
        \bottomrule
    \end{tabular}
\end{table}

Both prompt groups have higher mean OOV/\texttt{X} rates than the general text baselines. Type to token ratio follows the same ordering of group means and depends on length. It does not provide an independent quality signal. The distributions for individual datasets overlap. Group means for formatting and dependency features do not support a stable distinction across the catalogue. We use this experiment as a bounded check of central tendency on the measured dimensions.

\subsection{Cards for the Seven Analysed Datasets}
\label{sec:datasets-descriptions}

The following cards give the Publisher, Size, License, Link, and Description fields for the seven datasets used in detailed analysis. Table~\ref{tab:datasets_characteristics} summarises these datasets.

\begingroup
\sloppy
\begin{enumerate}[leftmargin=*]
    \item {\sf 1100+ ChatGPT Prompts for Business}
        \begin{itemize}
            \item \textbf{Publisher}: Chris Porter
            \item \textbf{Size}: 1235 instances
            \item \textbf{License}: -
            \item \textbf{Link}: \href{https://chatgpt-business-prompts.notion.site/1100-ChatGPT-Prompts-for-Business-eea03b0bc9b84ae7a5bdbd76a67460f3}{\nolinkurl{https://chatgpt-business-prompts.notion.site/1100-ChatGPT-Prompts-for-Business}}
            \item \textbf{Description}: ``1100+ ChatGPT Prompts for Business'' is a Notion dataset with 1,235 prompts for business tasks. Topics include buyer personas, content strategy, digital marketing, email campaigns, market research, product development, and finance. The prompts provide guidance for professional and business users.
        \end{itemize}
    \item {\sf BoredHumans}
        \begin{itemize}
            \item \textbf{Publisher}: Impulse Communications, Inc.
            \item \textbf{Size}: 964 source instances; 956 retained prompts
            \item \textbf{License}: -
            \item \textbf{Link}: \url{https://boredhumans.com/prompts.php}
            \item \textbf{Description}: BoredHumans combines prompts from sources that include Awesome ChatGPT Prompts, Data Science Prompts, and Tree of Thought Prompting. It covers several domains and prompt styles.
        \end{itemize}
    \item {\sf databricks-dolly-15K}
        \begin{itemize}
            \item \textbf{Publisher}: Databricks
            \item \textbf{Size}: 15,011 source records; 14,779 retained prompts
            \item \textbf{License}: CC-BY-SA-3.0
            \item \textbf{Link}: \url{https://huggingface.co/datasets/databricks/databricks-dolly-15k}
            \item \textbf{Description}: Databricks Dolly 15K contains over 15,000 instruction and response pairs written by Databricks employees. Its eight categories include brainstorming, classification, closed and open question answering, generation, information extraction, and summarisation. Some records use context from Wikipedia. The English dataset supports model tuning, synthetic data generation, and data augmentation under the stated license.
        \end{itemize}
    \item {\sf medical-o1-reasoning-SFT}
        \begin{itemize}
            \item \textbf{Publisher}: The Chinese University of Hong Kong, Shenzhen et al.
            \item \textbf{Size}: 90,120 source records across configurations; 19,679 retained English prompts
            \item \textbf{License}: Apache-2.0
            \item \textbf{Link}: \url{https://huggingface.co/datasets/FreedomIntelligence/medical-o1-reasoning-SFT}
            \item \textbf{Description}: medical-o1-reasoning-SFT supports supervised model tuning for medical reasoning in HuatuoGPT-o1. It contains English and Chinese instruction and response pairs generated by GPT-4o for clinical problems and checked by a medical verifier. Separate configurations provide data in one language or a mix of languages. The dataset supports question answering and text generation under the stated license.
        \end{itemize}
    \item {\sf OASST1}
        \begin{itemize}
            \item \textbf{Publisher}: OpenAssistant
            \item \textbf{Size}: 161,443 source messages; 22,079 retained English prompt messages
            \item \textbf{License}: Apache-2.0
            \item \textbf{Link}: \url{https://huggingface.co/datasets/OpenAssistant/oasst1}
            \item \textbf{Description}: OpenAssistant Conversations contains 161,443 messages in 66,497 conversation trees. The 84,437 training and 4,401 validation rows include both user and assistant roles and therefore are not prompt counts. Our prompt view contains 55,507 user-role messages, 22,850 English candidates, and 22,079 prompts after final deduplication and length checks. The dataset includes quality ratings and supports supervised model tuning and reward-model development.
        \end{itemize}
    \item {\sf Self-Instruct}
        \begin{itemize}
            \item \textbf{Publisher}: University of Washington et al.
            \item \textbf{Size}: 82,612 source records; 81,673 retained prompts
            \item \textbf{License}: Apache-2.0
            \item \textbf{Link}: \url{https://huggingface.co/datasets/yizhongw/self_instruct}
            \item \textbf{Description}: The analysed configuration contains 82,612 self-generated prompt-completion records and yields 81,673 prompts after extraction, deduplication, and quality checks. The separate 52,445-example P3 configuration caused the earlier card mismatch and is not analysed. Other configurations include Super Natural Instructions and expert-written evaluation tasks.
        \end{itemize}
    \item {\sf ShareGPT90K}
        \begin{itemize}
            \item \textbf{Publisher}: RyokoAI
            \item \textbf{Size}: 90,665 source threads; 181,577 retained prompt turns
            \item \textbf{License}: CC0
            \item \textbf{Link}: \url{https://huggingface.co/datasets/liyucheng/ShareGPT90K}
            \item \textbf{Description}: ShareGPT90K contains 90,665 conversational threads. Expanding the threads yields 226,625 user prompt turns; language, deduplication, and quality checks retain 181,577 prompts. Thus this card reports both source threads and the processed prompt turns used in Table~\ref{tab:datasets_characteristics}.
        \end{itemize}
\end{enumerate}
\endgroup

\section{Additional Experimental Results}

\subsection{Token Level Analysis}
\label{sec:token-app}

\begin{figure*}[t]
    \centering
    \begin{subfigure}[t]{0.47\textwidth}
        \includegraphics[height=6.5cm]{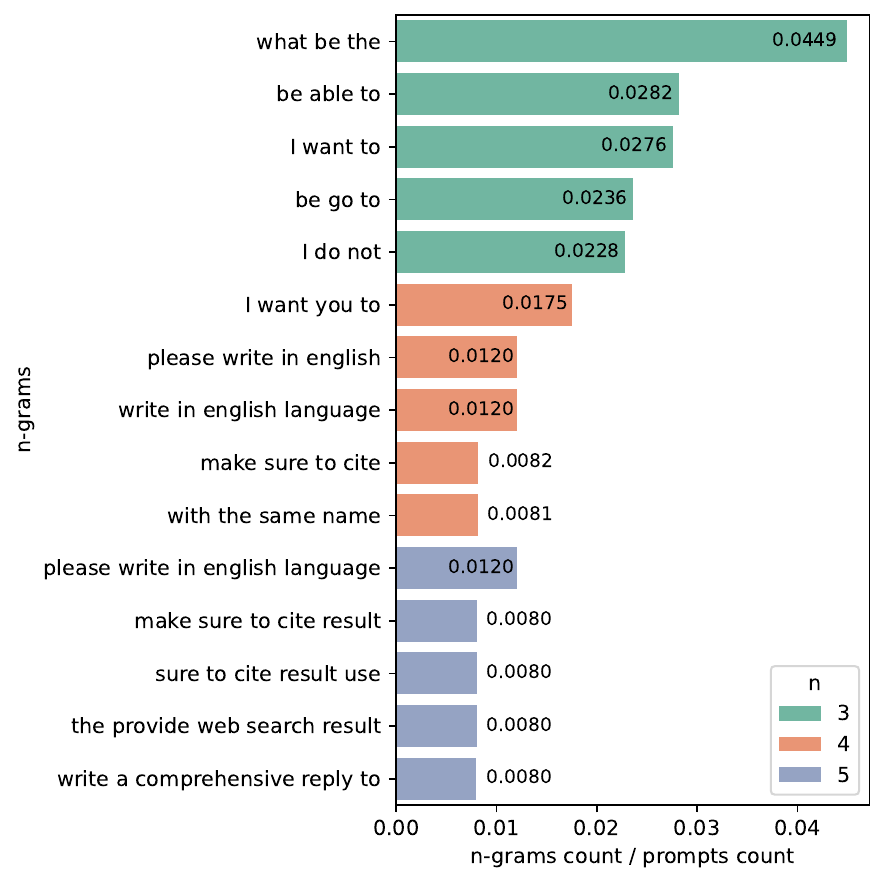}
        \caption{\scriptsize Five most common $n$ grams in \textsf{ShareGPT} for $n$ equal to 3, 4, and 5.}
    \end{subfigure}
    \hspace{.5cm}
    \begin{subfigure}[t]{0.47\textwidth}
        \includegraphics[height=6.5cm]{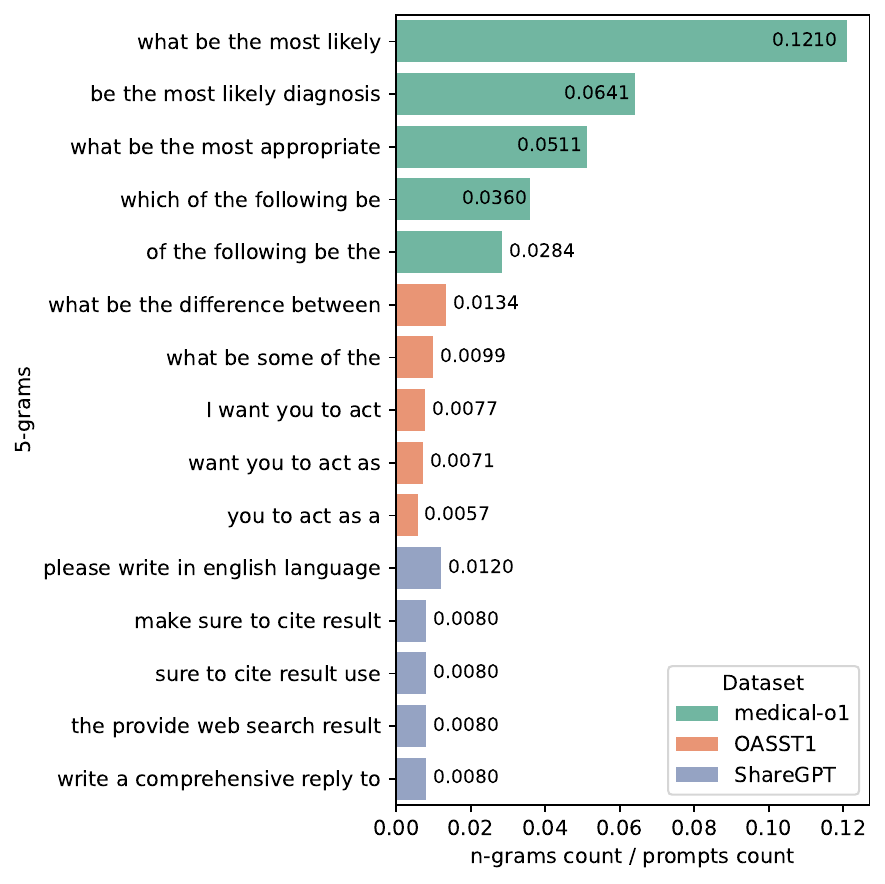}
        \caption{\scriptsize Five most common sequences of five words in \textsf{medical-o1}, \textsf{OASST1}, and \textsf{ShareGPT}.}
    \end{subfigure}
    \caption{\small Sequences of three, four, and five words within a dataset and across datasets. Each ratio is the count of a sequence divided by the number of prompts in the dataset.}
    \label{fig:ngram}
\end{figure*}

\begin{figure*}[t]
    \centering
    \begin{subfigure}[t]{0.31\textwidth}
        \includegraphics[height=4.2cm]{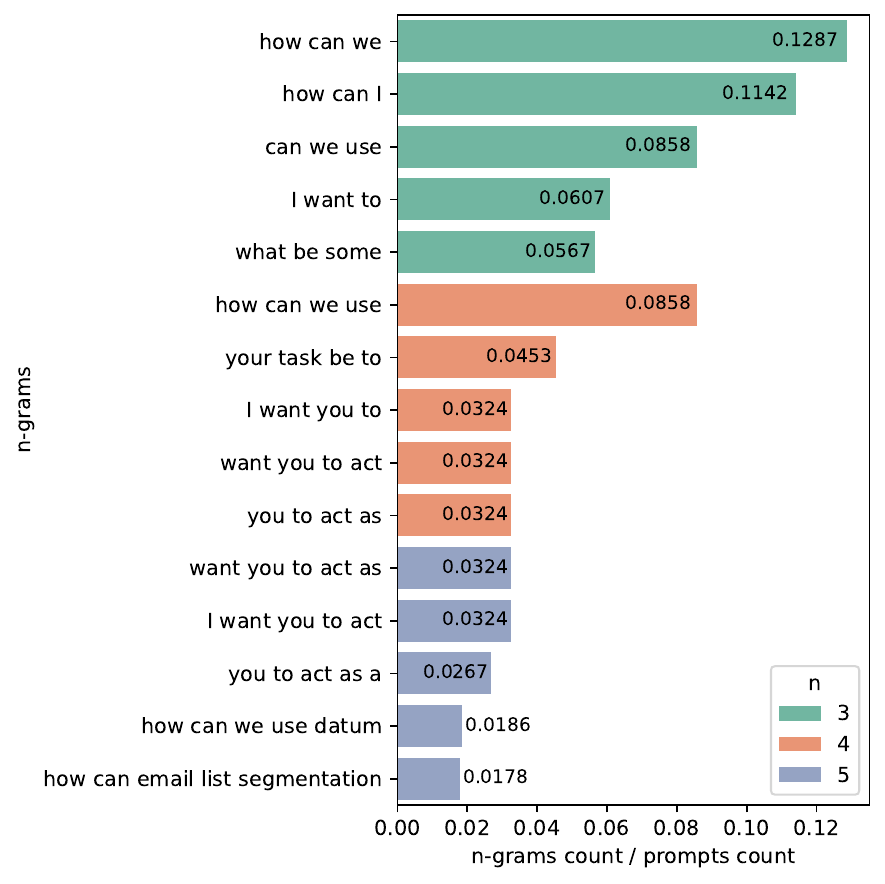}
        \caption{\scriptsize Five most common $n$ grams in \textsf{1.1k-business} for $n$ equal to 3, 4, and 5.}
    \end{subfigure}
    \hfill
    \begin{subfigure}[t]{0.31\textwidth}
        \includegraphics[height=4.2cm]{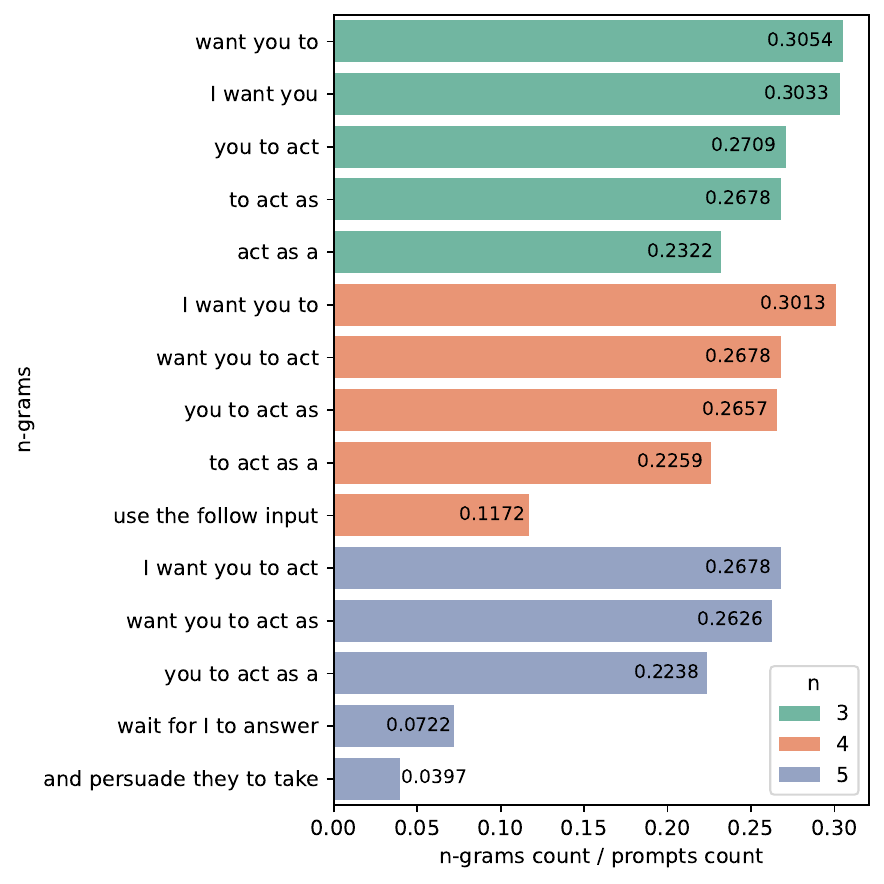}
        \caption{\scriptsize Five most common $n$ grams in \textsf{BoredHumans} for $n$ equal to 3, 4, and 5.}
    \end{subfigure}
    \hfill
    \begin{subfigure}[t]{0.31\textwidth}
        \includegraphics[height=4.2cm]{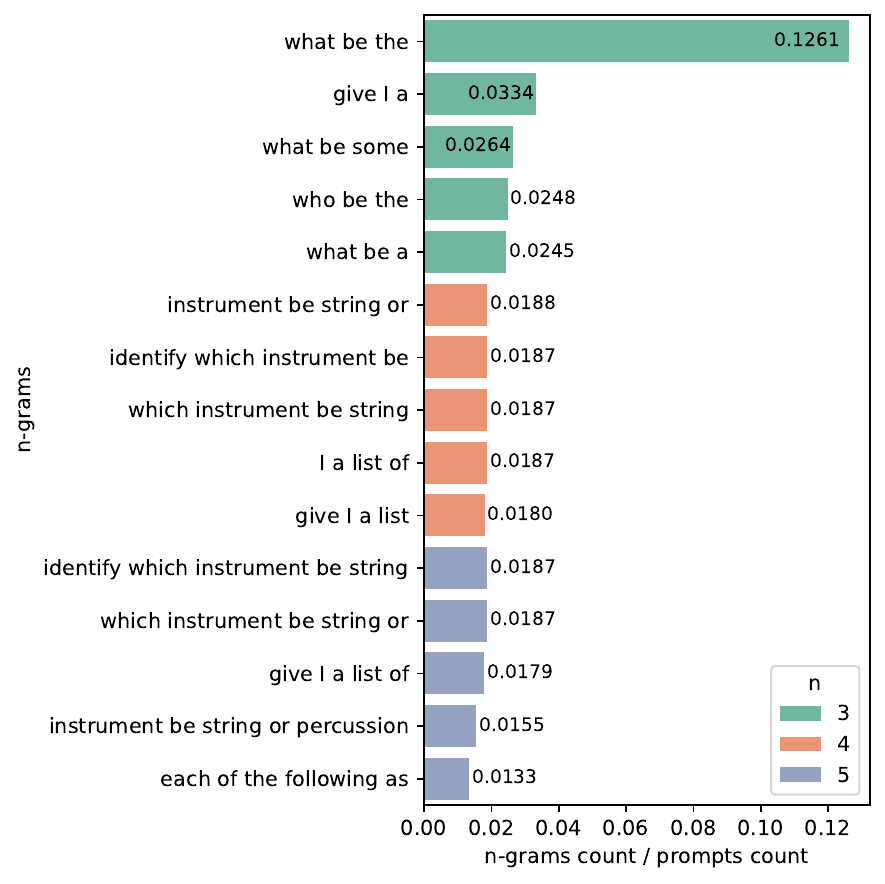}
        \caption{\scriptsize Five most common $n$ grams in \textsf{dolly-15k} for $n$ equal to 3, 4, and 5.}
    \end{subfigure}
    \begin{subfigure}[t]{0.31\textwidth}
        \includegraphics[height=4.2cm]{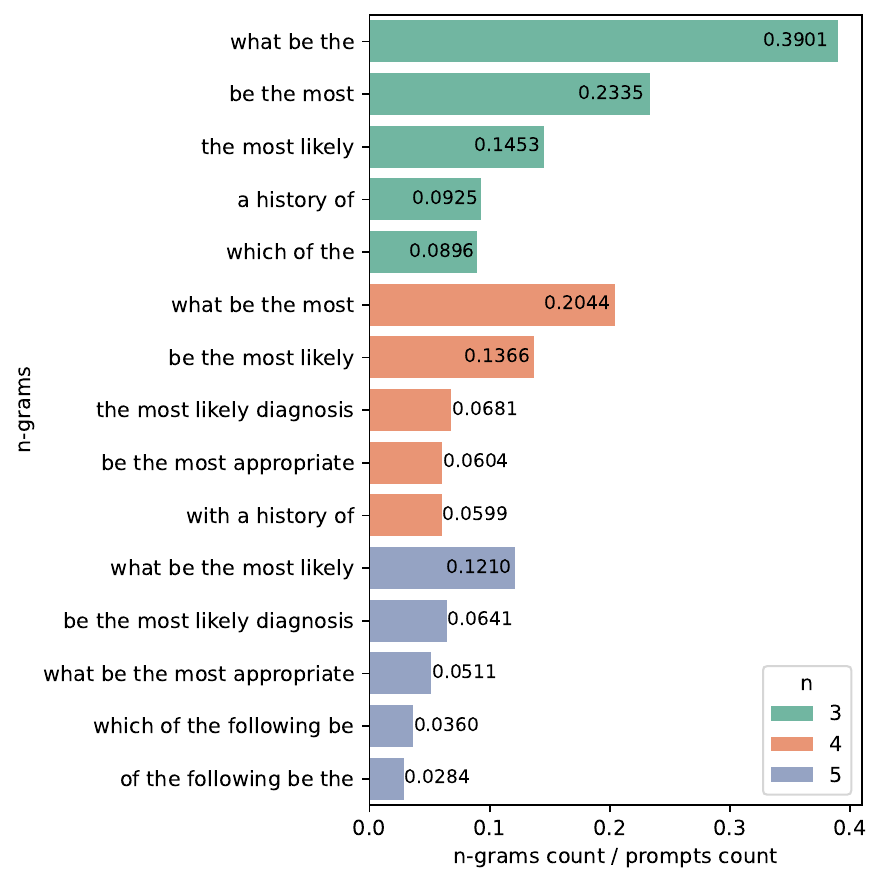}
        \caption{\scriptsize Five most common $n$ grams in \textsf{medical-o1} for $n$ equal to 3, 4, and 5.}
    \end{subfigure}
    \hfill
    \begin{subfigure}[t]{0.31\textwidth}
        \includegraphics[height=4.2cm]{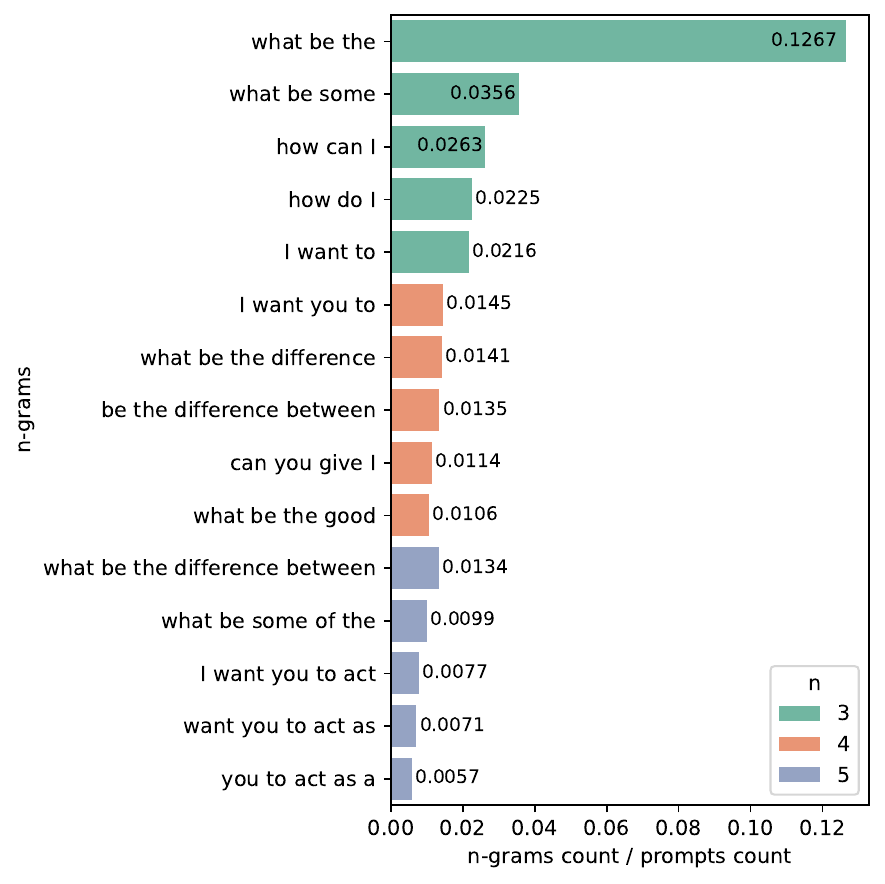}
        \caption{\scriptsize Five most common $n$ grams in \textsf{OASST1} for $n$ equal to 3, 4, and 5.}
    \end{subfigure}
    \hfill
    \begin{subfigure}[t]{0.31\textwidth}
        \includegraphics[height=4.2cm]{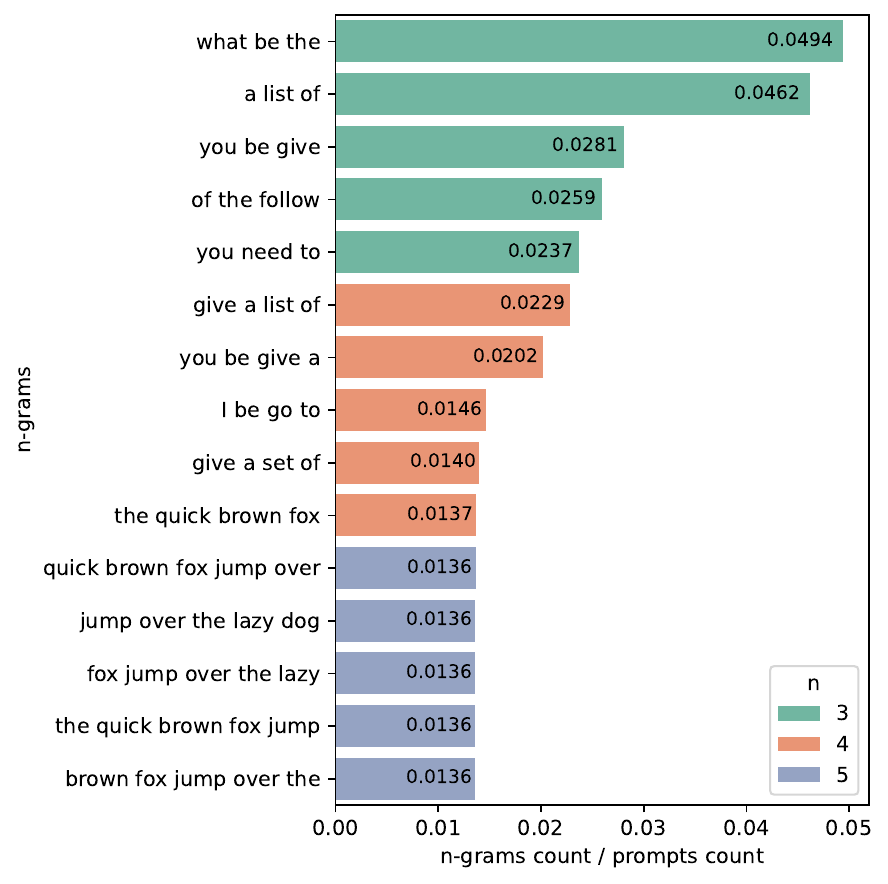}
        \caption{\scriptsize Five most common $n$ grams in \textsf{Self-Instruct} for $n$ equal to 3, 4, and 5.}
    \end{subfigure}
    
    \caption{\small Sequences of three, four, and five words within each dataset.}
    \label{fig:ngram-ds}
\end{figure*}

\begin{figure*}[t]
    \centering
    \begin{subfigure}[t]{0.3\textwidth}
        \includegraphics[width=\linewidth]{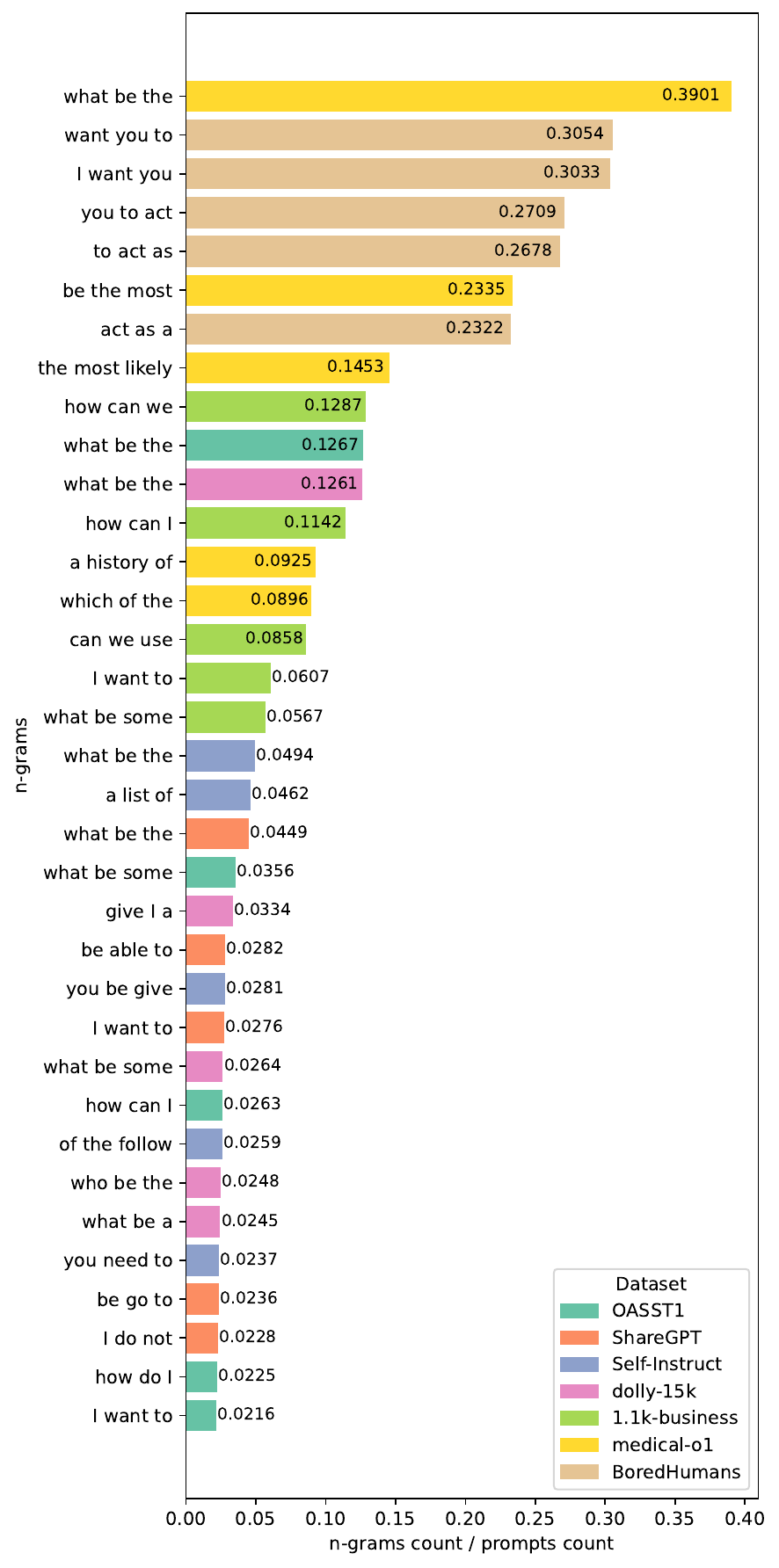}
        \caption{\scriptsize Sequences of three words.}
    \end{subfigure}
    \begin{subfigure}[t]{0.3\textwidth}
        \includegraphics[width=\linewidth]{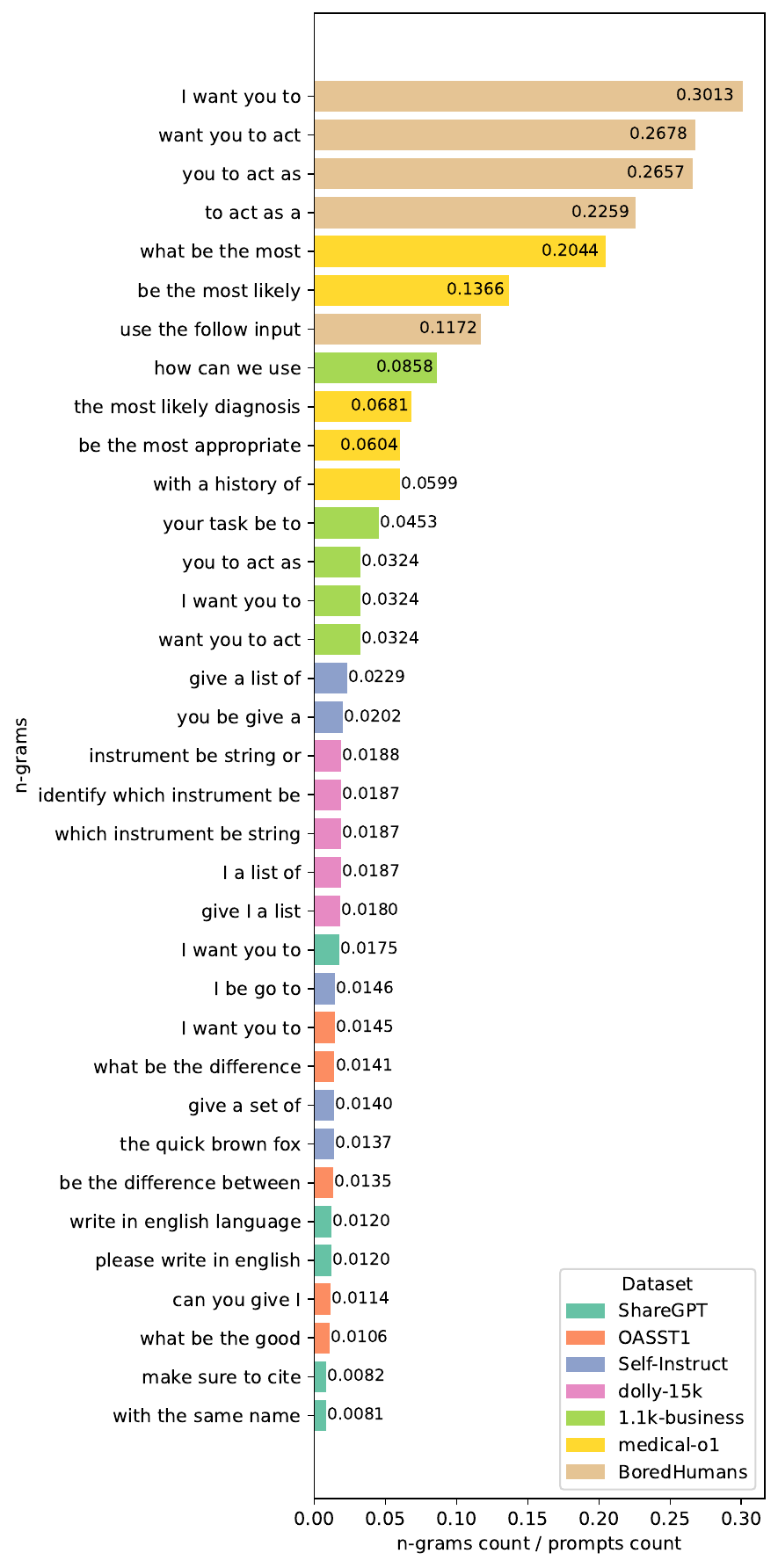}
        \caption{\scriptsize Sequences of four words.}
    \end{subfigure}
    \begin{subfigure}[t]{0.3\textwidth}
        \includegraphics[width=\linewidth]{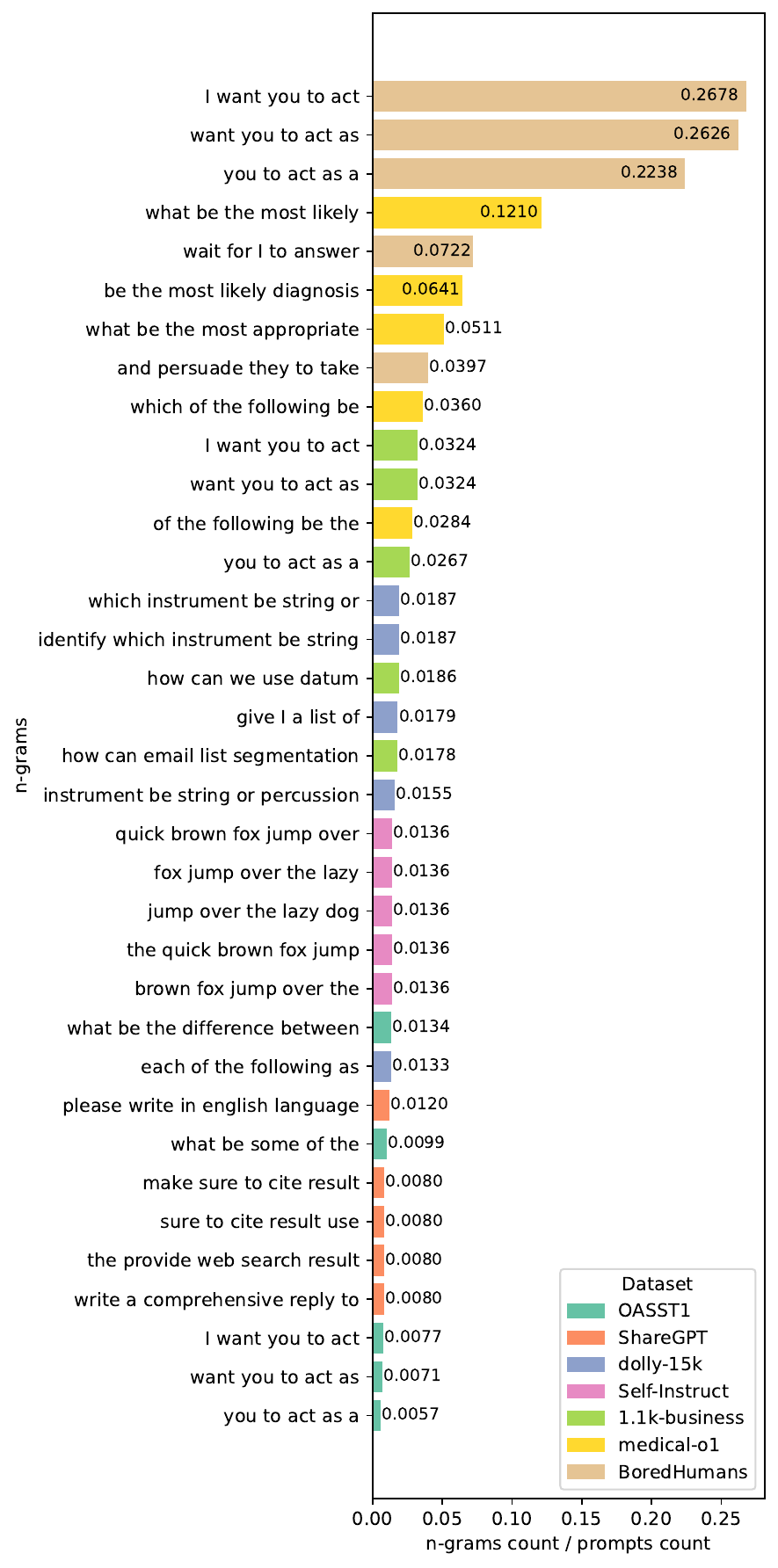}
        \caption{\scriptsize Sequences of five words.}
    \end{subfigure}
    \caption{The five most common $n$ grams across datasets.}
    \label{fig:ngram-n}
\end{figure*}

Figure~\ref{fig:ngram-ds} compares sequences of three, four, and five words for all datasets. Figure~\ref{fig:ngram} gives the ShareGPT results. Figure~\ref{fig:ngram-n} compares the five most common sequences across datasets.

The common sequences mainly contain commands and topic terms, which matches the main paper. The figures also show repetition and overlapping sequences.
\begin{enumerate} [leftmargin=*]
    \item Some datasets contain unusual common sequences. Examples include ``\textit{identify which instrument be string}'' in {\sf dolly-15} and ``\textit{The quick brown fox jumps over the lazy dog}'' in {\sf Self-Instruct}. These sequences show repeated content in template inputs or construction instructions. Such repetition may affect the balance of the content.
    \item Fixed sentences produce several overlapping sequences. For example, ``\textit{The quick brown fox jumps over the lazy dog}'' produces sequences such as ``\textit{quick brown fox jump over}'' and ``\textit{jump over the lazy dog}.''
\end{enumerate}

\subsection{Syntactic Analysis}
\label{sec:syntactic-app}

\begin{figure*}[t]
    \centering
    \begin{subfigure}[t]{0.45\textwidth}
        \includegraphics[width=\linewidth]{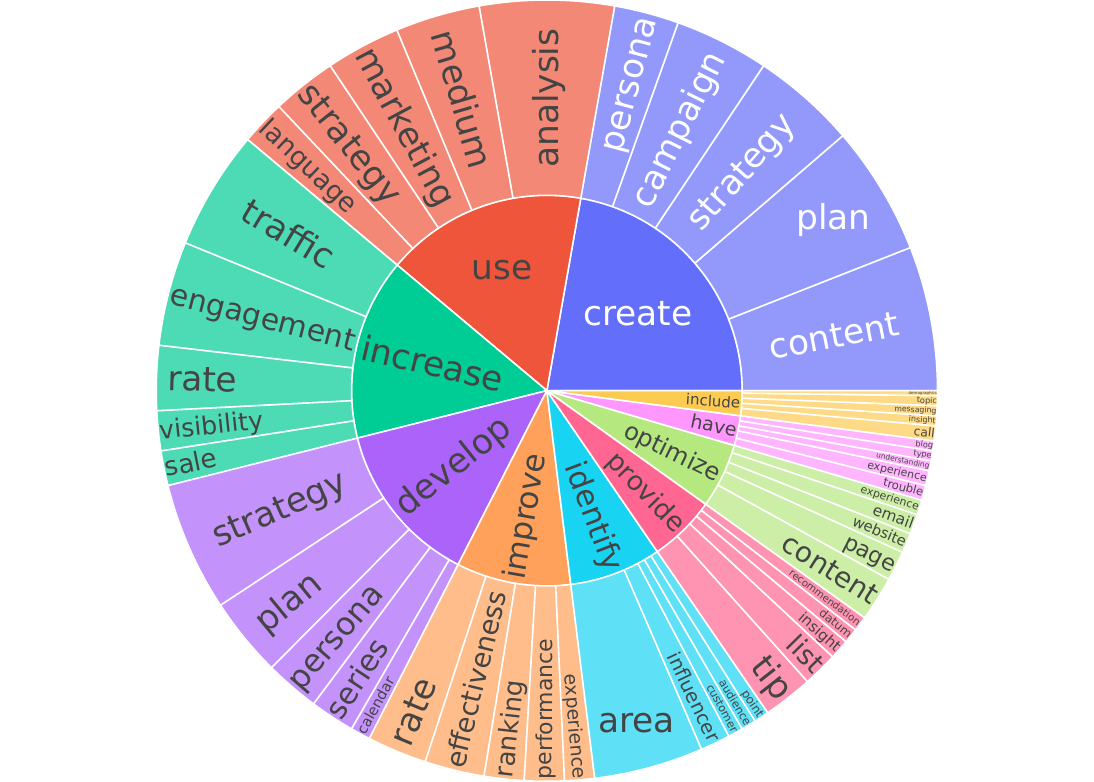}
        \caption{\scriptsize \textsf{1.1k-business}} 
    \end{subfigure}
    \hfill
    \begin{subfigure}[t]{0.45\textwidth}
        \includegraphics[width=\linewidth]{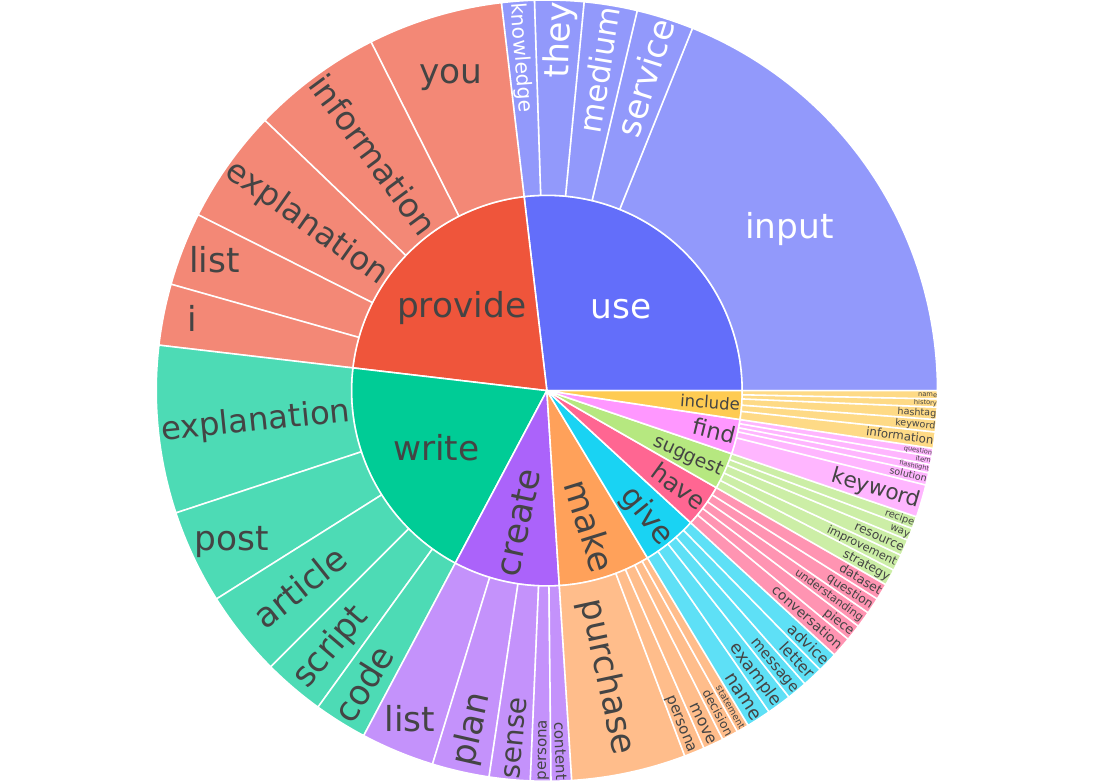}
        \caption{\scriptsize \textsf{BoredHumans}}
    \end{subfigure}
    \\[\baselineskip]
    \begin{subfigure}[t]{0.45\textwidth}
        \includegraphics[width=\linewidth]{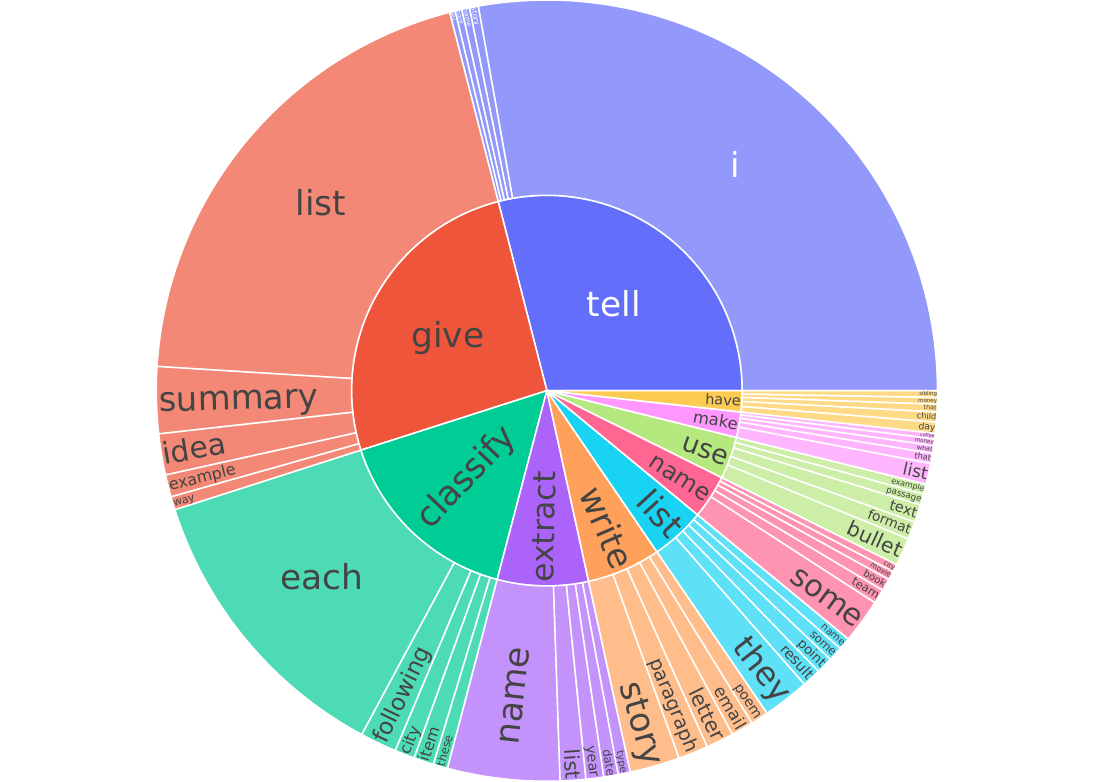}
        \caption{\scriptsize \textsf{dolly-15k}} 
    \end{subfigure}
    \hfill
    \begin{subfigure}[t]{0.45\textwidth}
        \includegraphics[width=\linewidth]{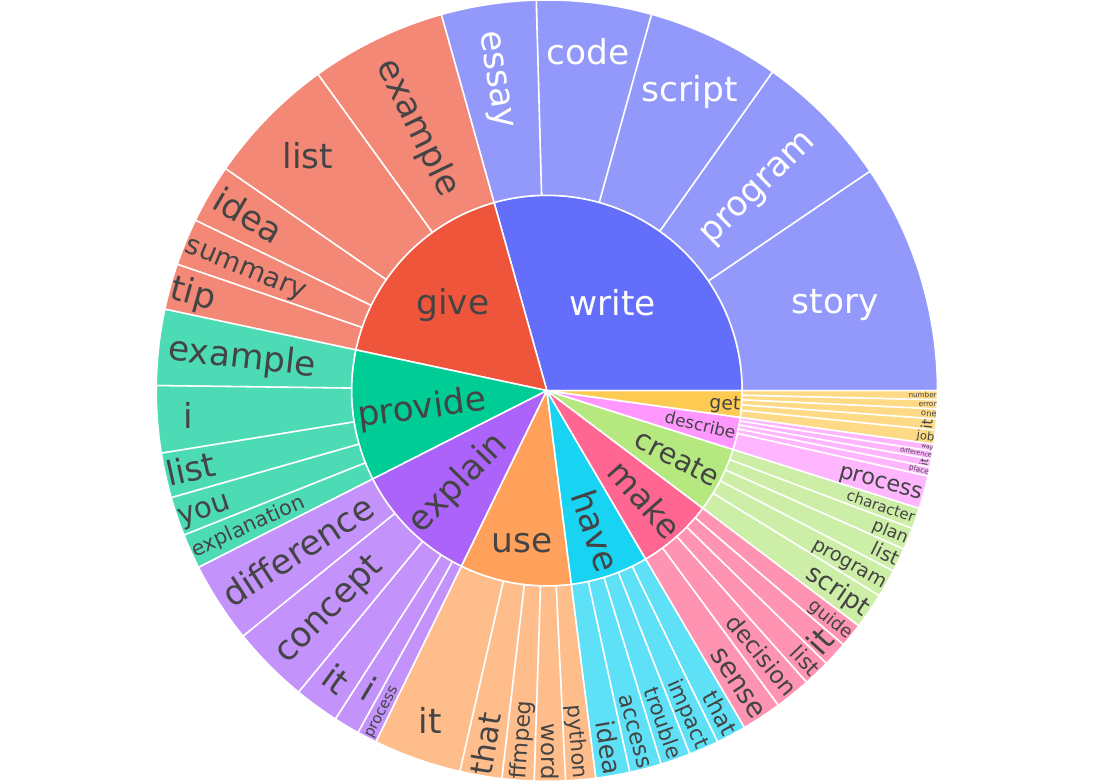}
        \caption{\scriptsize \textsf{OASST1}}
    \end{subfigure}
    \\[\baselineskip]
    \begin{subfigure}[t]{0.45\textwidth}
        \includegraphics[width=\linewidth]{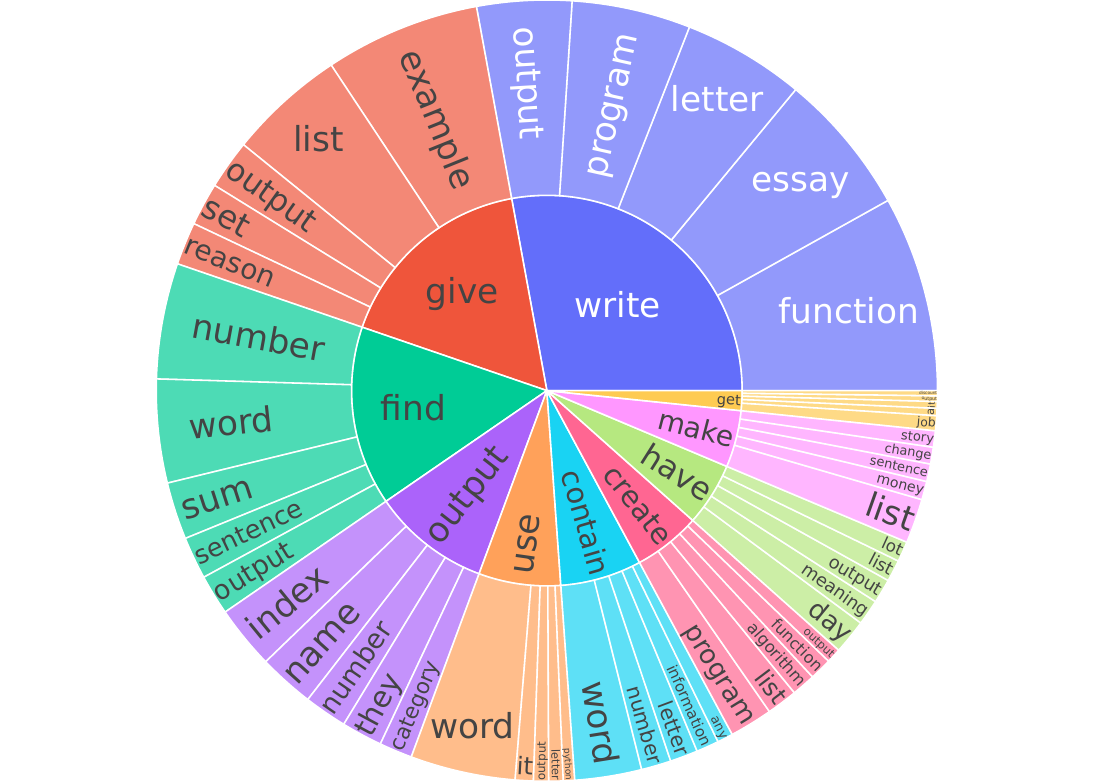}
        \caption{\scriptsize \textsf{Self-Instruct}} 
    \end{subfigure}
    \caption{\small The ten most common verbs and their five most common direct noun objects in prompt datasets.}
    \label{fig:verb-noun-app}
\end{figure*}

Table~\ref{tab:dependency-app} reports the proportion of every dependency type found in each dataset. Table~\ref{tab:pos-app} reports the proportion of every part of speech tag. Figure~\ref{fig:verb-noun-app} shows the ten most common verbs and their five most common direct noun objects. The main paper gives the corresponding figures for \textsf{medical-o1} and \textsf{ShareGPT}.

The additional data support two findings. The {\sf medical-o1} dataset has a high proportion of numerical modifiers at 0.0276 and passive auxiliaries at 0.0101. It also has a high numeral rate of 0.0309. These features reflect medical language that emphasises terms, precision, processes, and outcomes. The verb and noun pairs in {\sf 1.1k-business} include business phrases such as \textit{``create plan''} and \textit{``create strategy.''} The pairs in {\sf BoredHumans}, {\sf OASST1}, and {\sf Self-Instruct} cover more general uses.

In {\sf dolly-15k}, the most common verb and noun pairs have a skewed distribution. The most common nouns occur mainly with one or two verbs. Pairs such as \textit{``tell i,''} \textit{``give list,''} and \textit{``classify each''} also provide little information about the task. The writing habits of individual annotators may contribute to this pattern.

\begin{table}[t]
\centering
\caption{\small Proportions of all detected dependency types. A dash marks a type that was not detected in a dataset.}
\label{tab:dependency-app}
\resizebox{\columnwidth}{!}{\begin{tabular}{c|ccccccccc}
\toprule
\textbf{Dependency Type} & \textbf{1.1k-business} & \textbf{BoredHumans} & \textbf{dolly-15k} & \textbf{medical-o1} & \textbf{OASST1} & \textbf{Self-Instruct} & \textbf{ShareGPT} \\
\midrule
punct & 0.1227 & \textbf{0.1985} & 0.1445 & \underline{0.1216} & 0.1273 & 0.1863 & 0.154 \\
prep & 0.0759 & \underline{0.0672} & 0.0866 & \textbf{0.1013} & 0.0816 & 0.0676 & 0.0764 \\
det & \underline{0.0518} & 0.0692 & \textbf{0.0961} & 0.0906 & 0.0841 & 0.0838 & 0.0693 \\
pobj & 0.0718 & \underline{0.062} & 0.0817 & \textbf{0.0979} & 0.076 & 0.0645 & 0.0711 \\
nsubj & 0.0596 & 0.0545 & 0.065 & \underline{0.0469} & \textbf{0.0739} & 0.0596 & 0.0562 \\
ROOT & 0.0528 & 0.0462 & 0.0768 & 0.0444 & 0.0604 & \textbf{0.0792} & \underline{0.0437} \\
amod & 0.0573 & 0.0527 & 0.0469 & \textbf{0.1072} & 0.0523 & \underline{0.0384} & 0.048 \\
dobj & \textbf{0.0904} & 0.0665 & 0.0447 & \underline{0.0315} & 0.0594 & 0.057 & 0.0519 \\
compound & \textbf{0.0742} & 0.0471 & 0.0719 & 0.0716 & 0.0436 & \underline{0.023} & 0.0576 \\
conj & 0.0457 & 0.0494 & \textbf{0.0569} & 0.0391 & \underline{0.0343} & 0.0359 & 0.0371 \\
aux & \textbf{0.0642} & 0.0425 & 0.0257 & \underline{0.0143} & 0.0495 & 0.0302 & 0.0355 \\
dep & 0.0095 & 0.0306 & \underline{0.007} & 0.0183 & 0.0218 & \textbf{0.0611} & 0.0577 \\
cc & \textbf{0.04} & 0.0297 & \underline{0.0203} & 0.0291 & 0.0289 & \underline{0.0203} & 0.0287 \\
advmod & 0.0269 & 0.0273 & 0.0263 & 0.023 & \textbf{0.0383} & \underline{0.0224} & 0.0299 \\
poss & \textbf{0.0401} & 0.0183 & \underline{0.0084} & 0.0132 & 0.0118 & 0.0124 & 0.0116 \\
appos & \underline{0.003} & 0.0223 & 0.017 & 0.0099 & 0.0129 & 0.0207 & \textbf{0.0238} \\
attr & \underline{0.0044} & 0.005 & \textbf{0.0306} & 0.014 & 0.0163 & 0.0113 & 0.0083 \\
nummod & \underline{0.003} & 0.0073 & 0.0096 & \textbf{0.0276} & 0.0093 & 0.0136 & 0.0155 \\
nmod & \textbf{0.0252} & 0.0126 & \underline{0.0042} & 0.0095 & 0.0068 & 0.0043 & 0.0126 \\
ccomp & 0.0058 & 0.0129 & 0.0089 & \underline{0.0044} & 0.013 & \textbf{0.0142} & 0.013 \\
relcl & \textbf{0.0146} & 0.0088 & 0.0097 & \underline{0.0072} & 0.0109 & 0.0097 & 0.0094 \\
xcomp & \textbf{0.0194} & 0.0104 & \underline{0.0042} & \underline{0.0042} & 0.0106 & 0.0079 & 0.0099 \\
advcl & 0.0104 & 0.0109 & \underline{0.0056} & 0.0066 & 0.0115 & 0.0086 & \textbf{0.0122} \\
npadvmod & \underline{0.0026} & 0.0068 & 0.0052 & \textbf{0.0141} & 0.0051 & 0.0052 & 0.0066 \\
acomp & \underline{0.0034} & 0.0041 & 0.0059 & 0.007 & 0.0086 & \textbf{0.0091} & 0.0068 \\
mark & \underline{0.002} & 0.0057 & 0.0038 & 0.0033 & 0.0087 & \textbf{0.0114} & 0.0095 \\
acl & \underline{0.0038} & 0.0062 & 0.0055 & 0.0077 & 0.0056 & \textbf{0.0079} & 0.0063 \\
auxpass & \underline{0.0016} & 0.0017 & 0.0062 & \textbf{0.0101} & 0.0055 & 0.0052 & 0.0059 \\
pcomp & \textbf{0.008} & 0.0048 & 0.0036 & 0.0053 & 0.0057 & \underline{0.0035} & 0.0051 \\
nsubjpass & \underline{0.0014} & 0.0015 & 0.005 & \textbf{0.009} & 0.0046 & 0.0048 & 0.0051 \\
neg & \underline{0.0015} & 0.0031 & 0.0016 & 0.002 & 0.0043 & 0.0033 & \textbf{0.0045} \\
case & \textbf{0.0035} & 0.0013 & 0.0032 & 0.0024 & 0.002 & \underline{0.0011} & 0.0023 \\
dative & 0.0003 & 0.0029 & \textbf{0.0041} & \underline{0.0001} & 0.0035 & 0.0023 & 0.0016 \\
prt & 0.0014 & 0.0018 & 0.0014 & \underline{0.0006} & 0.0025 & \textbf{0.0047} & 0.0025 \\
intj & 0.0004 & \textbf{0.0038} & 0.0012 & \underline{0.0002} & 0.0033 & 0.0016 & 0.0025 \\
agent & \underline{0.0002} & 0.0003 & 0.001 & \textbf{0.002} & 0.001 & 0.0015 & 0.0012 \\
expl & \underline{0.0} & 0.0002 & 0.0004 & 0.0007 & \textbf{0.0016} & 0.0014 & 0.0011 \\
quantmod & \underline{0.0} & 0.0002 & 0.0005 & 0.0007 & 0.0009 & 0.0014 & \textbf{0.0016} \\
meta & 0.0001 & 0.0016 & 0.0001 & \underline{0.0} & 0.0003 & \textbf{0.0018} & 0.0012 \\
oprd & \underline{0.0002} & \textbf{0.0009} & \textbf{0.0009} & 0.0007 & \textbf{0.0009} & 0.0004 & 0.0008 \\
predet & - & 0.0003 & 0.0005 & \underline{0.0001} & 0.0006 & \textbf{0.0009} & 0.0005 \\
csubj & 0.0005 & \underline{0.0001} & 0.0004 & 0.0004 & 0.0005 & \underline{0.0001} & \textbf{0.0007} \\
parataxis & - & \textbf{0.0009} & 0.0001 & \underline{0.0} & 0.0003 & 0.0002 & 0.0006 \\
preconj & \underline{0.0} & 0.0001 & \textbf{0.0008} & 0.0002 & 0.0002 & 0.0002 & 0.0003 \\
csubjpass & \underline{0.0} & - & \textbf{0.0001} & \underline{0.0} & \underline{0.0} & \underline{0.0} & \underline{0.0} \\
\bottomrule
\end{tabular}}
\end{table}

\begin{table}[t]
\centering
\caption{\small Proportions of all detected part of speech tags. A dash marks a tag that was not detected in a dataset.}
\label{tab:pos-app}
\resizebox{\columnwidth}{!}{
\begin{tabular}{l|ccccccc}
\toprule
\textbf{POS} & \textbf{1.1k-business} & \textbf{BoredHumans} & \textbf{dolly-15k} & \textbf{medical-o1} & \textbf{OASST1} & \textbf{Self-Instruct} & \textbf{ShareGPT} \\
\midrule
NOUN & \textbf{0.2637} & 0.2103 & \underline{0.1899} & 0.259 & 0.1946 & 0.2027 & 0.1944 \\
PUNCT & \underline{0.1094} & \textbf{0.1942} & 0.1435 & 0.1158 & 0.1231 & 0.1839 & 0.145 \\
VERB & \textbf{0.1302} & 0.1094 & 0.0871 & \underline{0.0775} & 0.1069 & 0.0999 & 0.0979 \\
ADP & 0.0758 & \underline{0.0678} & 0.0858 & \textbf{0.0998} & 0.0851 & 0.0701 & 0.0789 \\
DET & \underline{0.0506} & 0.0693 & \textbf{0.0949} & 0.0893 & 0.0839 & 0.0844 & 0.0696 \\
PRON & \textbf{0.0912} & 0.0708 & 0.0695 & \underline{0.0369} & 0.0869 & 0.0701 & 0.0583 \\
ADJ & 0.0588 & 0.0543 & 0.0538 & \textbf{0.1104} & 0.0632 & \underline{0.0498} & 0.0563 \\
PROPN & \underline{0.0219} & 0.0372 & \textbf{0.1272} & 0.0515 & 0.0471 & 0.0294 & 0.0703 \\
AUX & 0.0458 & \underline{0.0379} & 0.0608 & 0.0382 & \textbf{0.0644} & 0.0453 & 0.0423 \\
CCONJ & \textbf{0.0399} & 0.0294 & 0.0209 & 0.0291 & 0.0288 & \underline{0.0204} & 0.0286 \\
SPACE & - & 0.0267 & \underline{0.0053} & 0.0175 & 0.019 & 0.0504 & \textbf{0.0517} \\
PART & \textbf{0.0358} & 0.0223 & 0.013 & \underline{0.01} & 0.0222 & 0.0172 & 0.0213 \\
NUM & \underline{0.0041} & 0.0146 & 0.014 & \textbf{0.0309} & 0.0153 & 0.0282 & 0.0273 \\
ADV & \underline{0.0097} & \textbf{0.0259} & 0.0107 & 0.0209 & 0.0238 & 0.0153 & 0.0247 \\
SCONJ & 0.0199 & 0.0105 & 0.0198 & \underline{0.007} & \textbf{0.0242} & 0.0197 & 0.0152 \\
X & \textbf{0.035} & 0.0128 & 0.0015 & \underline{0.0005} & 0.0044 & 0.0075 & 0.0082 \\
SYM & \textbf{0.008} & 0.0037 & \underline{0.0008} & 0.0049 & 0.0035 & 0.0031 & 0.0073 \\
INTJ & \underline{0.0002} & 0.003 & 0.0012 & 0.001 & \textbf{0.0037} & 0.0027 & 0.0028 \\
\bottomrule
\end{tabular}
}
\end{table}

\begin{table}[t]
    \centering
    \caption{Three tokens with the highest TF-IDF weights in each dataset.}
    \label{tab:tfidf}
    \scriptsize
    \resizebox{\linewidth}{!}{\begin{tabular}{l|l}
            \toprule
            \textbf{Dataset}    & \textbf{Three largest token weights}             \\
            \midrule
            {\bf 1.1k-business} & content (0.308), email (0.284), marketing (0.245) \\
            {\bf BoredHumans}   & act (0.269), want (0.261), write (0.217)          \\
            {\bf dolly-15k}     & list (0.338), given (0.246), following (0.241)    \\
            {\bf medical-o1}    & old (0.427), year (0.367), patient (0.256)        \\
            {\bf OASST1}        & write (0.307), like (0.216), does (0.207)         \\
            {\bf Self-Instruct} & output (0.766), input (0.292), task (0.243)       \\
            {\bf ShareGPT}      & write (0.190), use (0.180), data (0.157)          \\
            \bottomrule
        \end{tabular}
    }
\end{table}

\subsection{Semantic Analysis}
\label{sec:semantic-app}

\begin{figure*}[t]
    \centering
    \begin{subfigure}[t]{0.3\textwidth}
        \includegraphics[width=\linewidth]{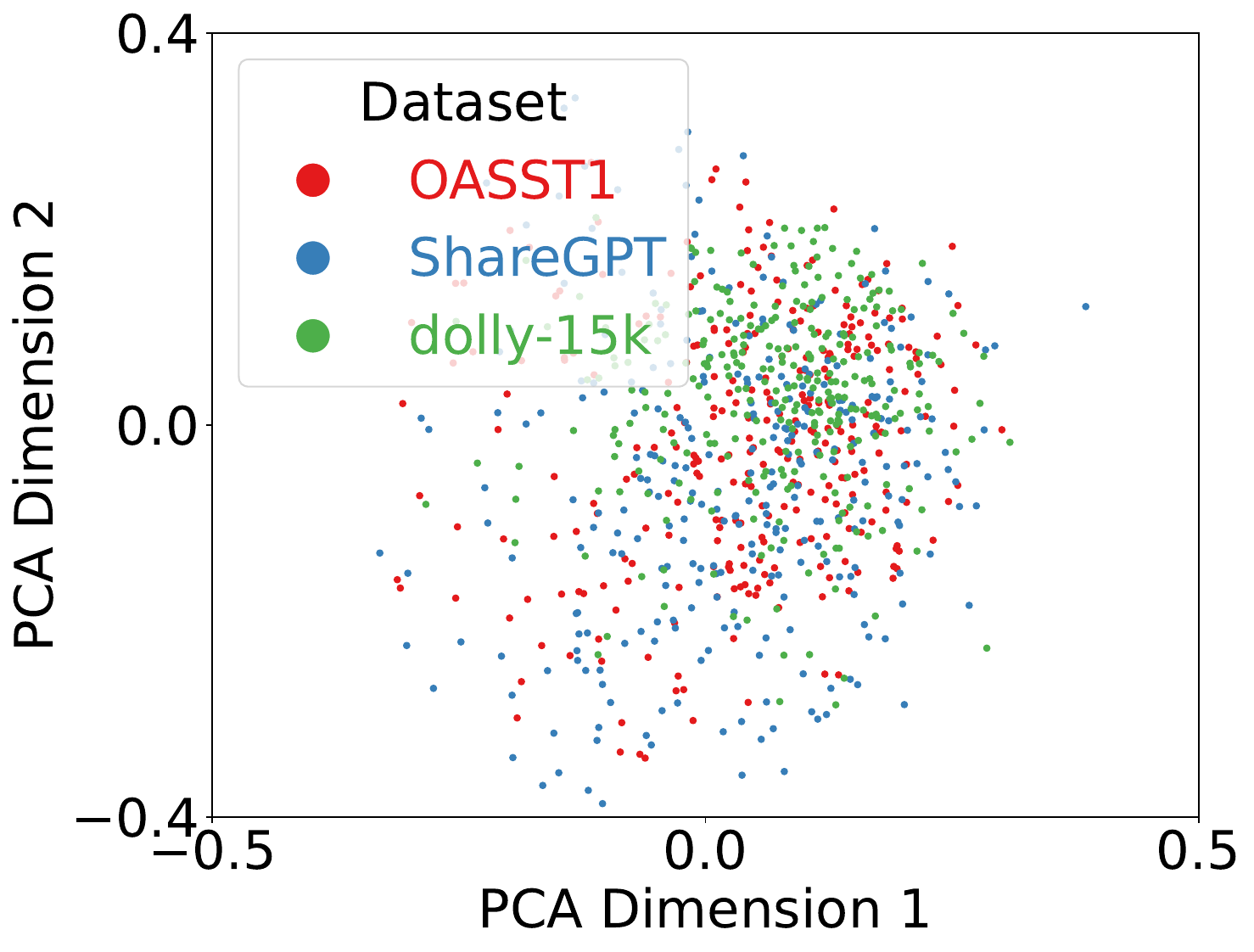}
        \caption{\scriptsize Overview of datasets written by people.}
    \end{subfigure}
    \hfill
    \begin{subfigure}[t]{0.3\textwidth}
        \includegraphics[width=\linewidth]{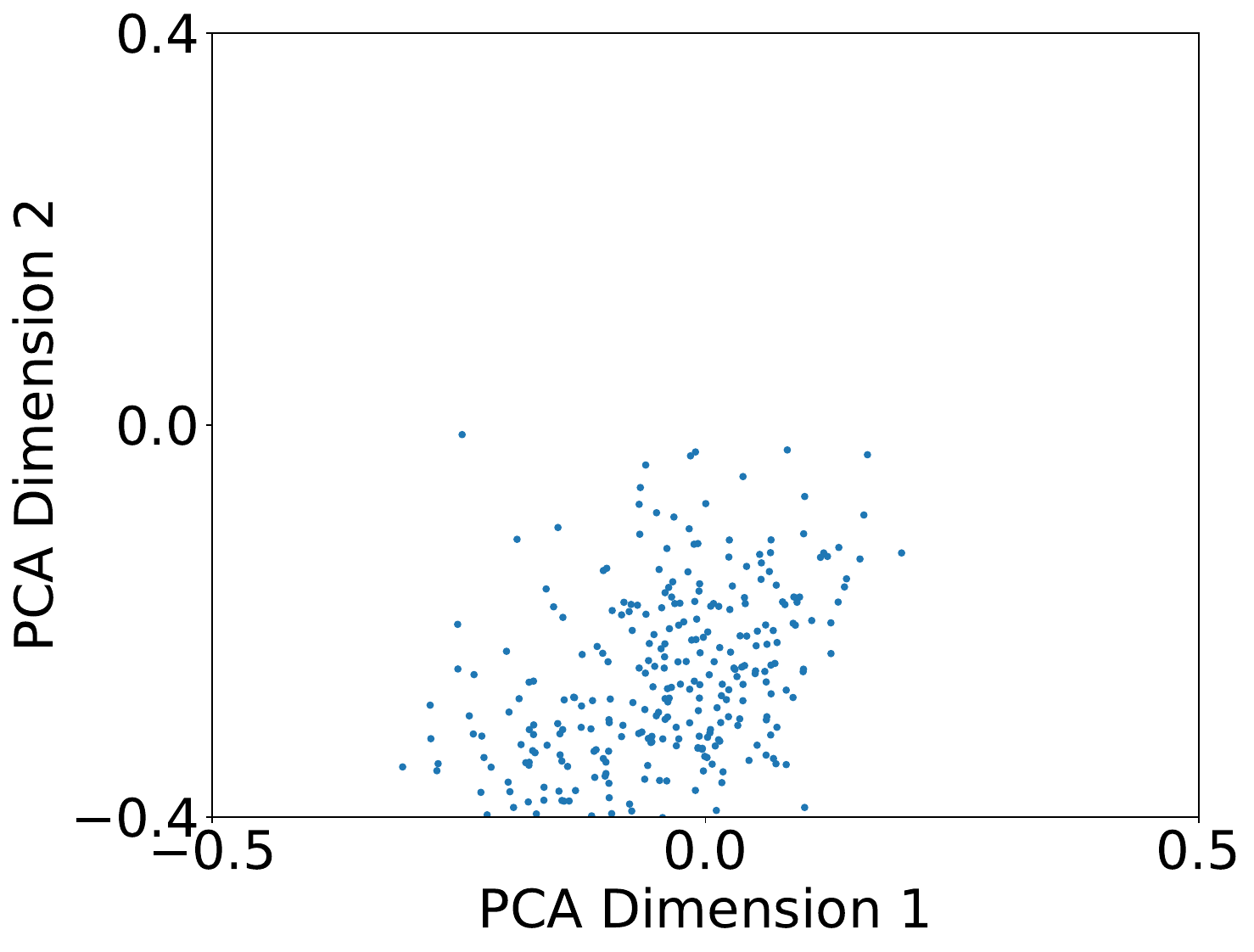}
        \caption{\scriptsize \textsf{1.1k-business}}
    \end{subfigure}
    \hfill
    \begin{subfigure}[t]{0.3\textwidth}
        \includegraphics[width=\linewidth]{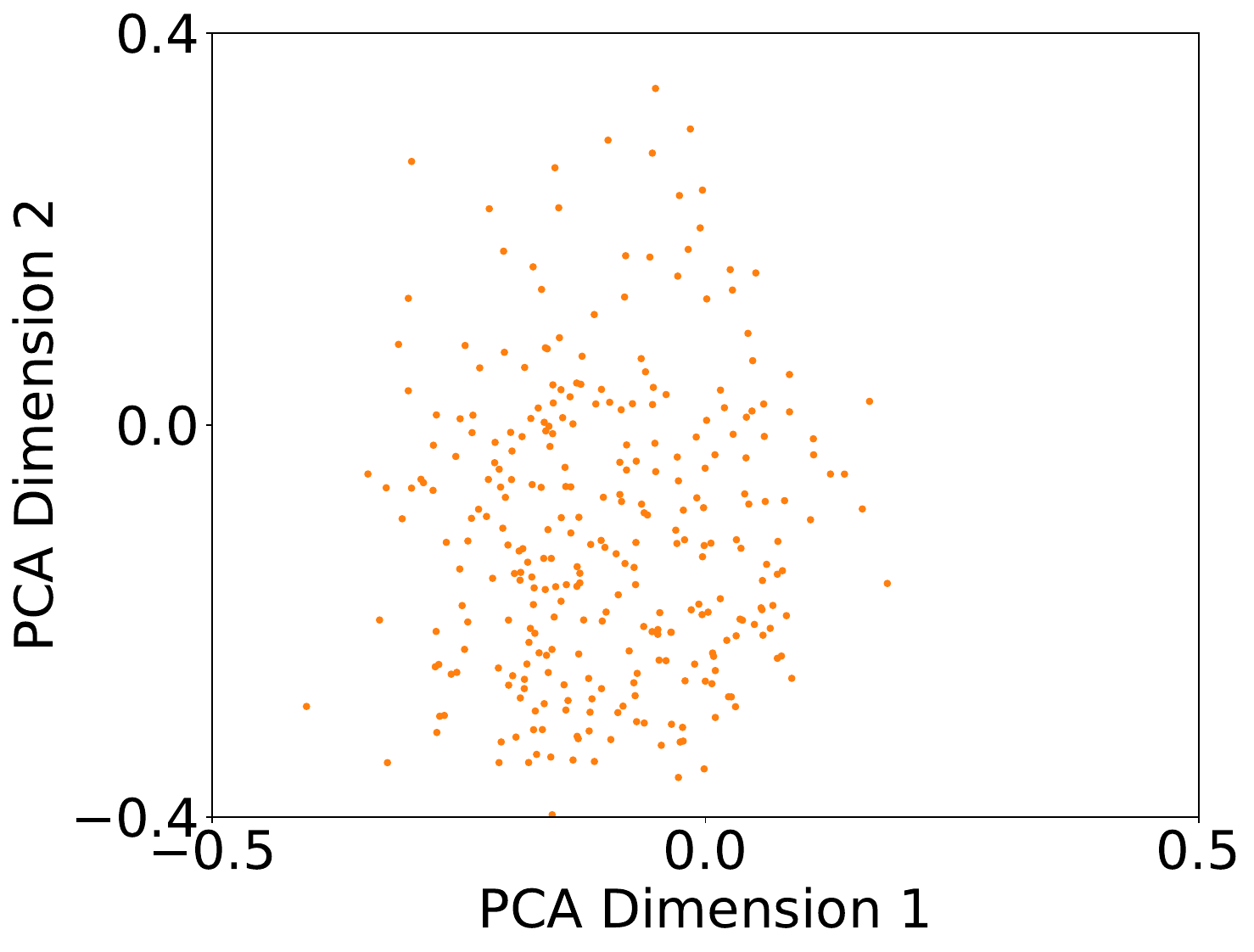}
        \caption{\scriptsize \textsf{BoredHumans}}
    \end{subfigure}
    \\[\baselineskip]
    \begin{subfigure}[t]{0.3\textwidth}
        \includegraphics[width=\linewidth]{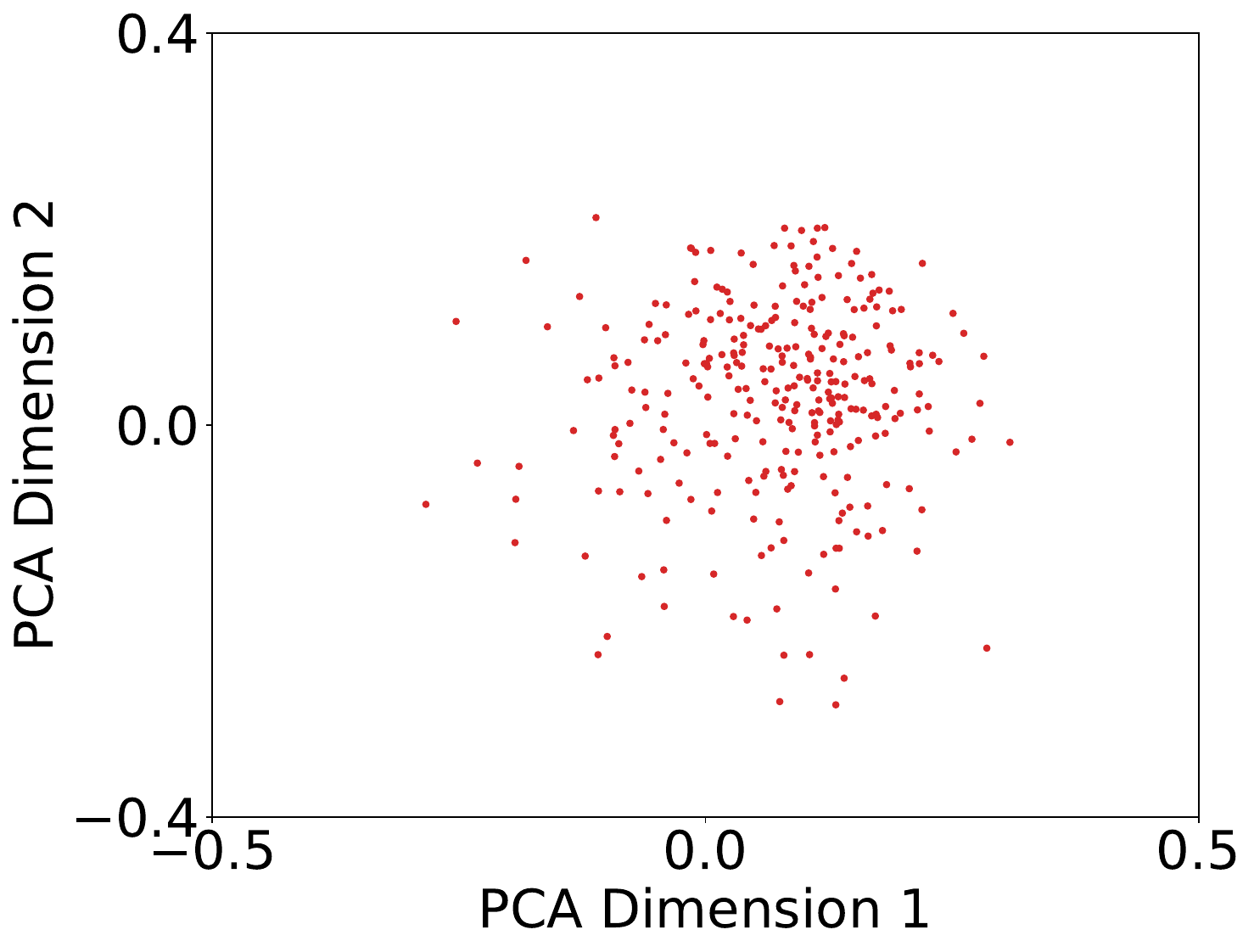}
        \caption{\scriptsize \textsf{dolly-15k}}
    \end{subfigure}
    \hfill
    \begin{subfigure}[t]{0.3\textwidth}
        \includegraphics[width=\linewidth]{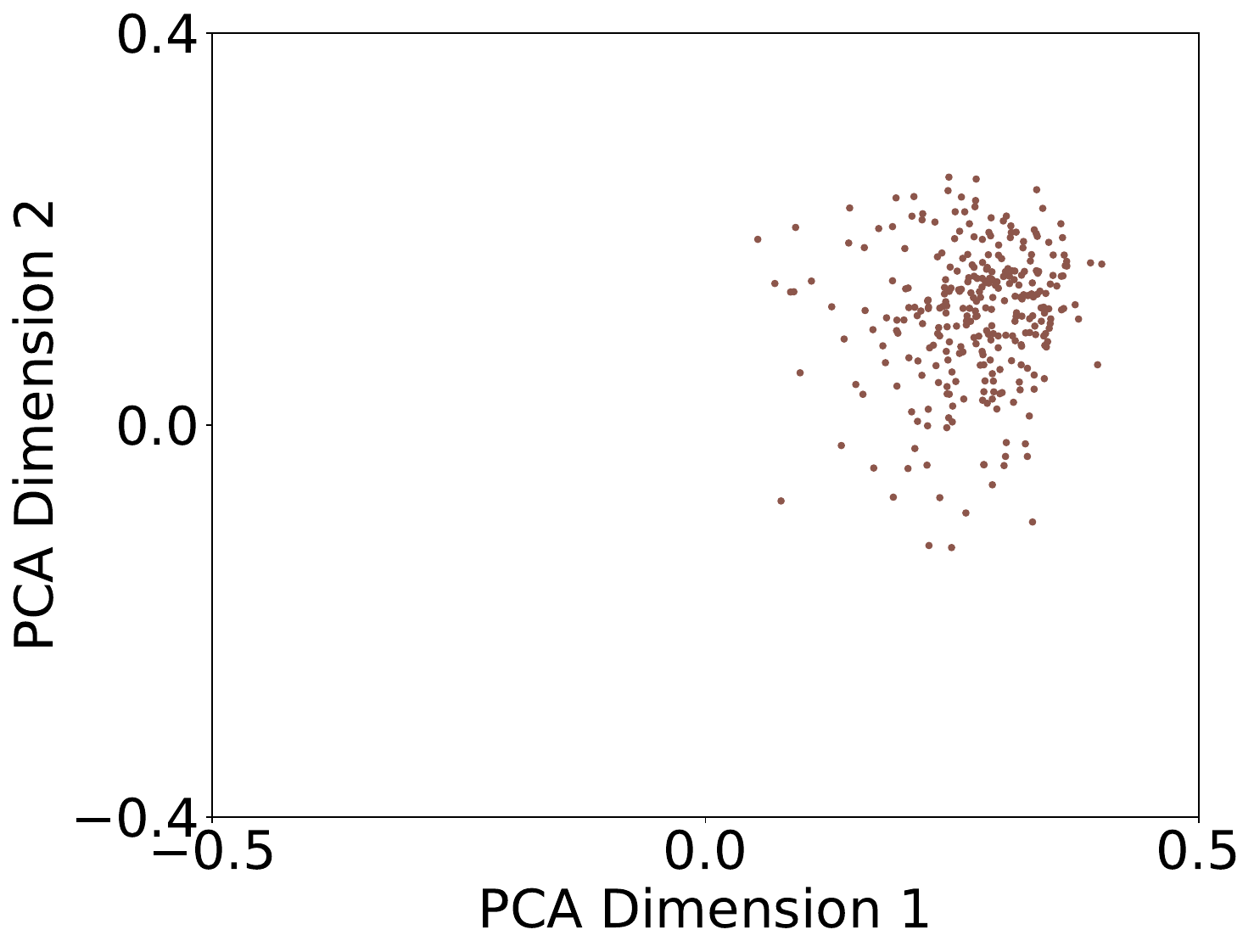}
        \caption{\scriptsize \textsf{medical-o1}}
    \end{subfigure}
    \hfill
    \begin{subfigure}[t]{0.3\textwidth}
        \includegraphics[width=\linewidth]{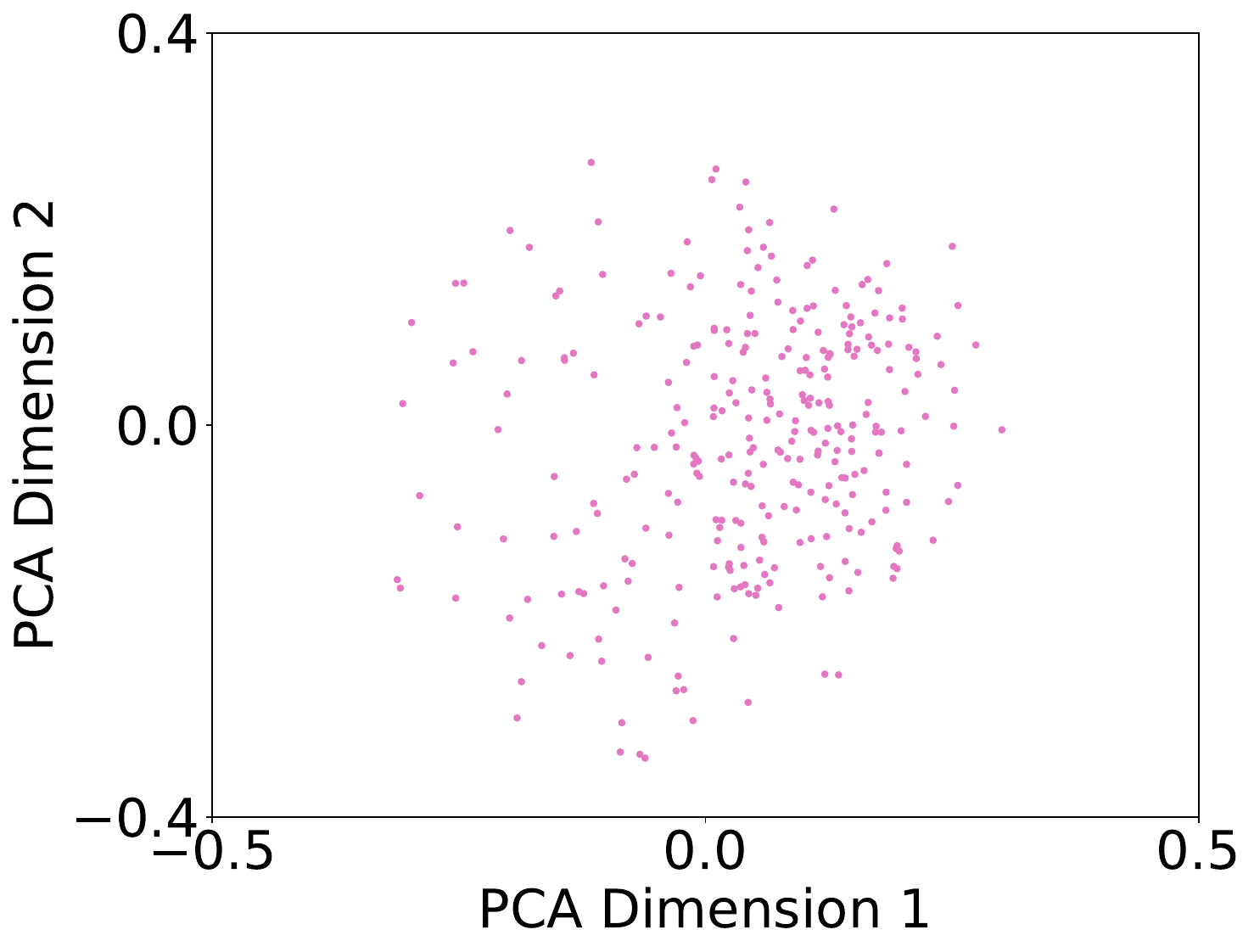}
        \caption{\scriptsize \textsf{OASST1}}
    \end{subfigure}
    \\[\baselineskip]
    \begin{subfigure}[t]{0.3\textwidth}
        \includegraphics[width=\linewidth]{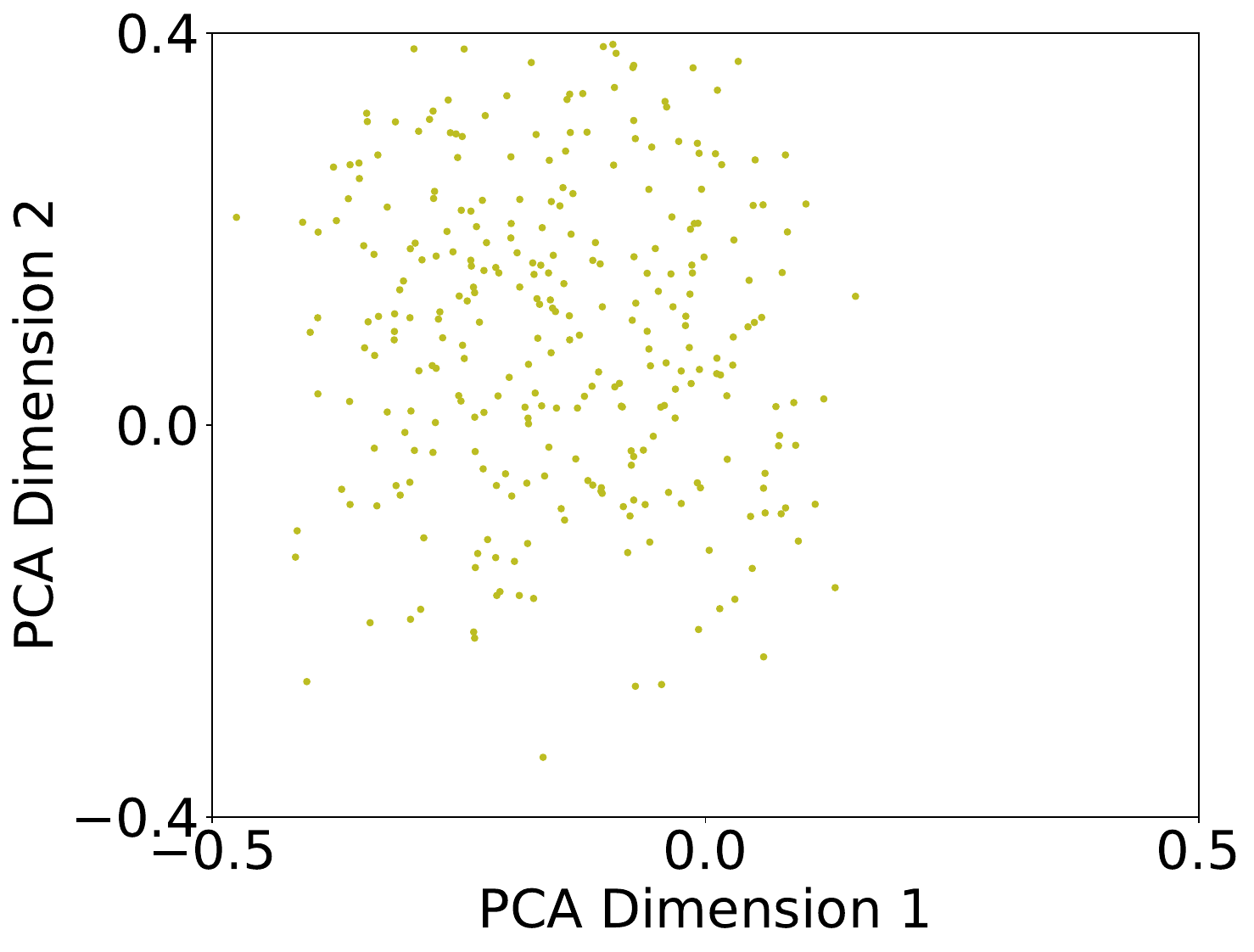}
        \caption{\scriptsize \textsf{Self-Instruct}}
    \end{subfigure}
    \hspace{3em}
    \begin{subfigure}[t]{0.3\textwidth}
        \includegraphics[width=\linewidth]{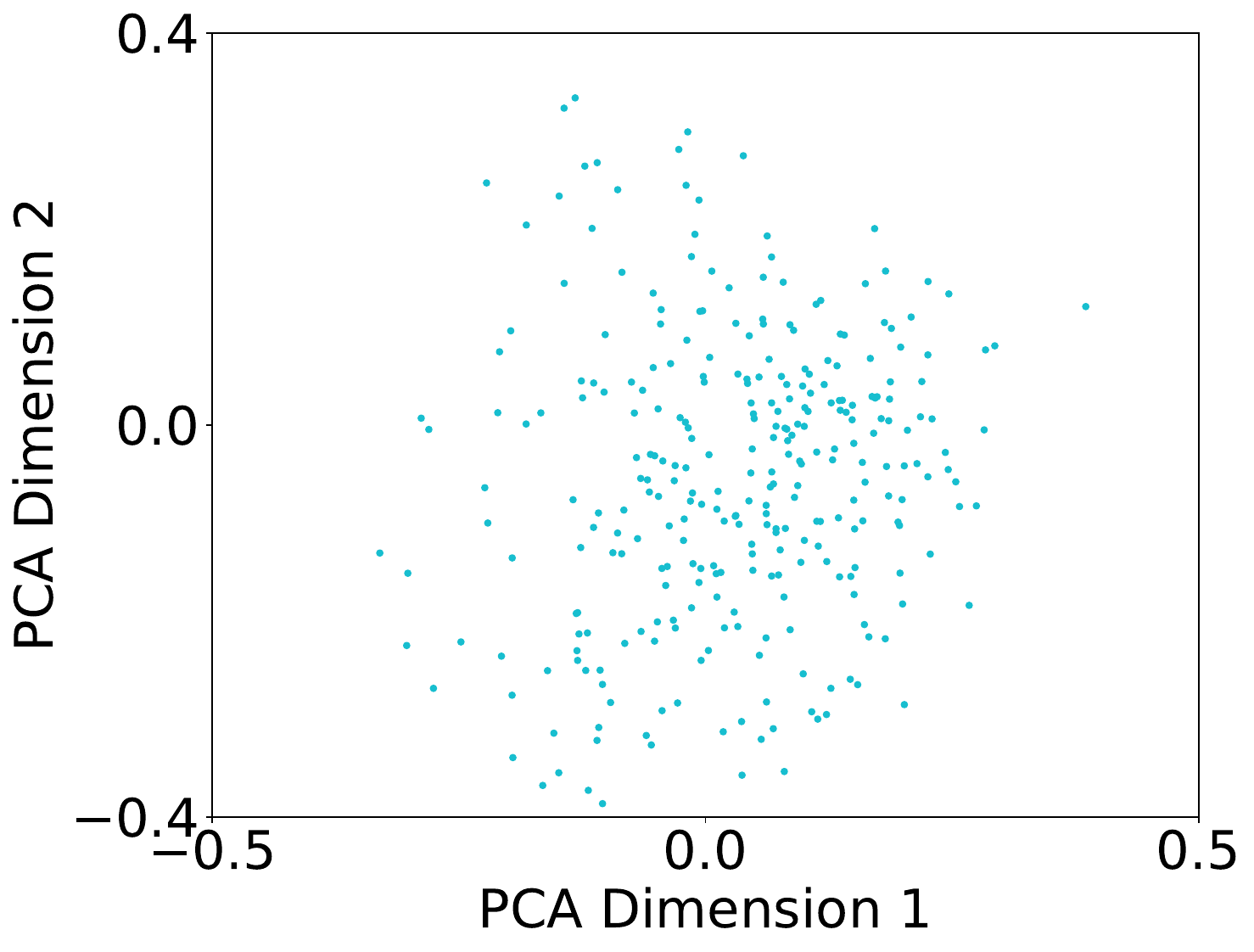}
        \caption{\scriptsize \textsf{ShareGPT}}
    \end{subfigure}
    \caption{\small PCA projection of prompt embeddings from 500 sampled prompts in each dataset. Panel (a) places all datasets written by people in a shared space. Panels (b) through (h) show each analysed dataset.}
    \label{fig:pca-app}
\end{figure*}

Figure~\ref{fig:pca-app} shows the sampled embedding points after PCA for all seven analysed datasets. The overview panel places the datasets written by people in a shared PCA space.

Datasets with focused topics form more concentrated groups. Examples include {\sf 1.1k-business} and {\sf medical-o1}. Datasets with broader topics form more dispersed groups. Examples include {\sf ShareGPT} and {\sf Self-Instruct}.

\clearpage

\subsection{Application}
\label{sec:application-app}

This section gives supplementary figures and tables for the application experiments in \S\ref{sec:application}. Sections~\ref{sec:filtering-app} and \ref{sec:domain-app} show MLP training and classification errors. Section~\ref{sec:importance-app} shows the input features that drive classifier decisions. Section~\ref{sec:quality-app} reports associations between the same linguistic features and elicited response quality under GPT-4 and human ratings.

\subsubsection{Prompt Filtering}
\label{sec:filtering-app}

\begin{figure*}[t]
    \centering
    \begin{subfigure}[t]{0.49\textwidth}
        \centering
        \includegraphics[width=\linewidth]{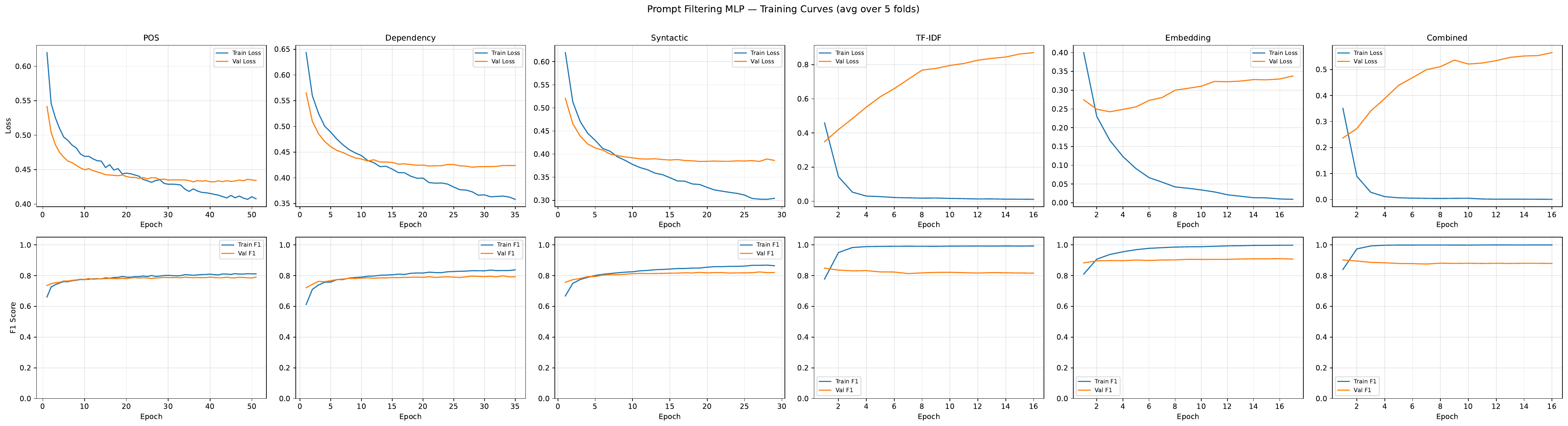}
        \caption{\small Training and validation curves for each feature set.}
        \label{fig:filtering_curves}
    \end{subfigure}
    \hfill
    \begin{subfigure}[t]{0.49\textwidth}
        \centering
        \includegraphics[width=\linewidth]{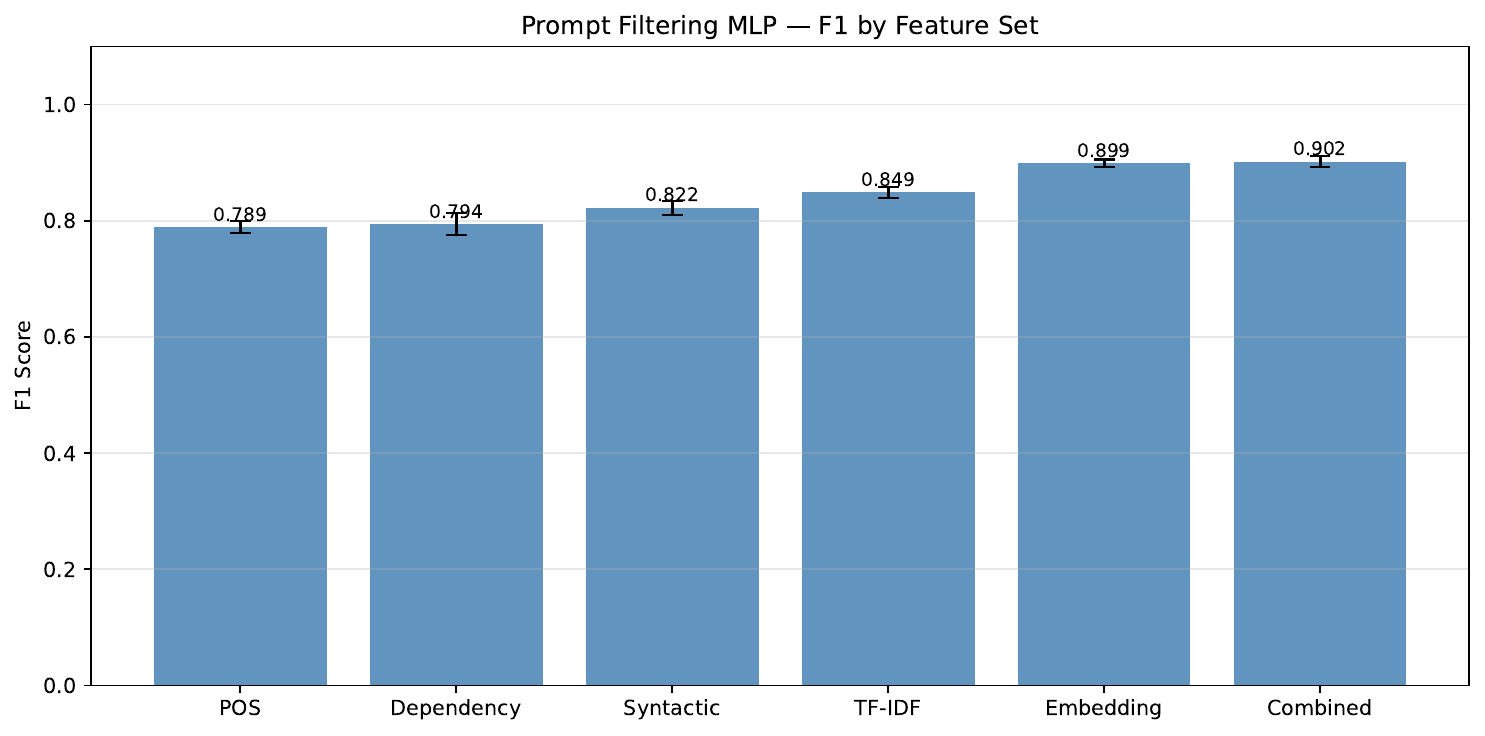}
        \caption{\small F1 comparison across feature sets.}
        \label{fig:filtering_f1}
    \end{subfigure}
    \caption{\small Prompt filtering results. Panel (a) shows MLP training. Panel (b) shows final F1 for each feature set. Combined reaches the highest F1, and sentence embeddings give a similar result. The Syntactic representation has 63 dimensions and reaches over 91\% of embedding F1 without embedding inference.}
    \label{fig:filtering_overview}
\end{figure*}

Figure~\ref{fig:filtering_overview} gives the prompt filtering training results that complement Table~\ref{tab:filtering}. Panel (a) shows training and validation loss and F1 for POS, Dependency, Syntactic, TF-IDF, Embedding, and Combined. These curves allow comparison of convergence, final loss, and the generalisation gap. The validation curves follow the training curves closely under a limit of 100 epochs and early stopping. This pattern shows that the 63 dimensional Syntactic representation has limited overfitting on the 6{,}000 samples. Panel (b) reports F1 under five fold cross validation. Sentence embeddings reach about 0.90. The Syntactic representation reaches 0.822, which is over 91\% of embedding F1, and requires no embedding inference or corpus vocabulary. The Combined representation with 5{,}447 dimensions provides a small improvement over embeddings. For the Isolation Forest, contamination is fixed at 0.30 and recall is evaluated on the fitted prompt examples; exactly 30\% are therefore designated outliers, mechanically producing recall 0.700 for every representation. This is a fixed operating point rather than an independently estimated recall.

\subsubsection{Source Domain Routing}
\label{sec:domain-app}

Tables~\ref{tab:domain_unsup} and \ref{tab:domain_holdout} extend the Phase B and Phase C results in Figures~\ref{fig:domain_phaseB_main} and \ref{fig:domain_phaseC_main}. The tables report POS, Dependency, Syntactic, TF-IDF, Embedding, and Combined. K-Means on the Sentence-BERT representation with 384 dimensions recovers the five domains without labels. It reaches NMI\,=\,0.80 and Purity\,=\,0.92. POS, Dependency, and Syntactic reach NMI at most 0.42, and TF-IDF reaches 0.45. The Combined representation with 5{,}447 dimensions reaches NMI\,=\,0.69 under K-Means and remains strong under supervised training.

\begin{table}[t]
\centering
\caption{\small Complete Phase B evaluation of source domain routing. The table reports K-Means NMI, ARI, and Purity and Nearest Centroid accuracy for all six feature representations. Sentence embeddings form the clearest five clusters. The high dimensional Combined representation gives weaker distance based clusters and remains strong under supervised classification.}
\label{tab:domain_unsup}
\scriptsize
\resizebox{\columnwidth}{!}{%
\begin{tabular}{@{}lcccc@{}}
\toprule
\textbf{Feature} & \textbf{K-Means NMI} & \textbf{K-Means ARI} & \textbf{Purity} & \textbf{NC Accuracy} \\
\midrule
POS         & .383 & .349 & .623 & .746 \\
Dependency  & .380 & .345 & .603 & .771 \\
Syntactic   & .421 & .406 & .675 & .794 \\
TF-IDF      & .450 & .200 & .396 & .915 \\
Embedding   & \textbf{.796} & \textbf{.805} & \textbf{.916} & .927 \\
Combined    & .689 & .453 & .591 & \textbf{.950} \\
\bottomrule
\end{tabular}
}
\end{table}

\begin{table}[t]
\centering
\caption{\small Complete Phase C evaluation of source domain routing on an 80/20 stratified split. POS with 18 dimensions reaches 85.0\% Macro-F1. Sentence embeddings reach 97.4\%, and Combined reaches 96.6\%. Confidence is the average maximum softmax probability for the predicted class on the test prompts.}
\label{tab:domain_holdout}
\scriptsize
\begin{tabular}{@{}lccc@{}}
\toprule
\textbf{Feature} & \textbf{Accuracy} & \textbf{Macro-F1} & \textbf{Avg Confidence} \\
\midrule
POS          & .851 & .850 & .881 \\
Embedding    & \textbf{.974} & \textbf{.974} & \textbf{.990} \\
Combined     & .966 & .966 & .990 \\
\bottomrule
\end{tabular}
\end{table}

\begin{figure*}[t]
    \centering
    \begin{subfigure}[t]{0.49\textwidth}
        \centering
        \includegraphics[width=\linewidth]{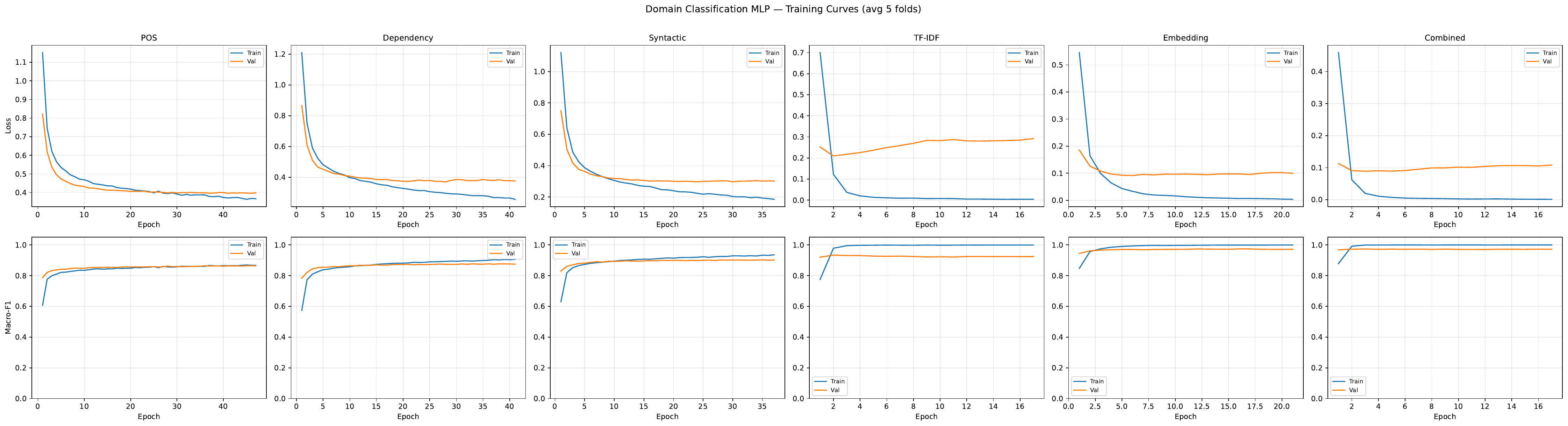}
        \caption{\small Training and validation curves.}
        \label{fig:domain_curves}
    \end{subfigure}
    \hfill
    \begin{subfigure}[t]{0.49\textwidth}
        \centering
        \includegraphics[width=\linewidth]{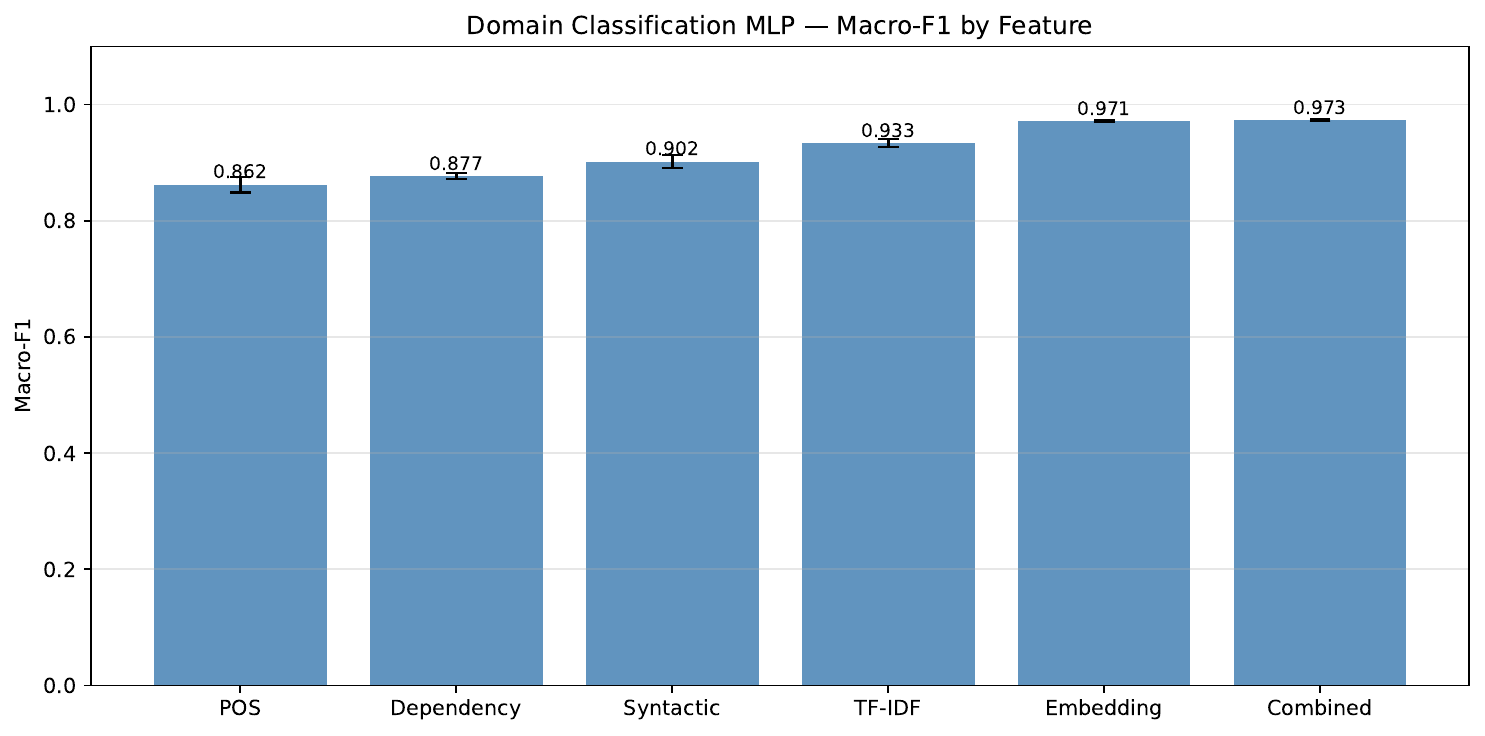}
        \caption{\small Macro-F1 comparison.}
        \label{fig:domain_f1}
    \end{subfigure}
    \\[.4em]
    \begin{subfigure}[t]{0.32\textwidth}
        \centering
        \includegraphics[width=\linewidth]{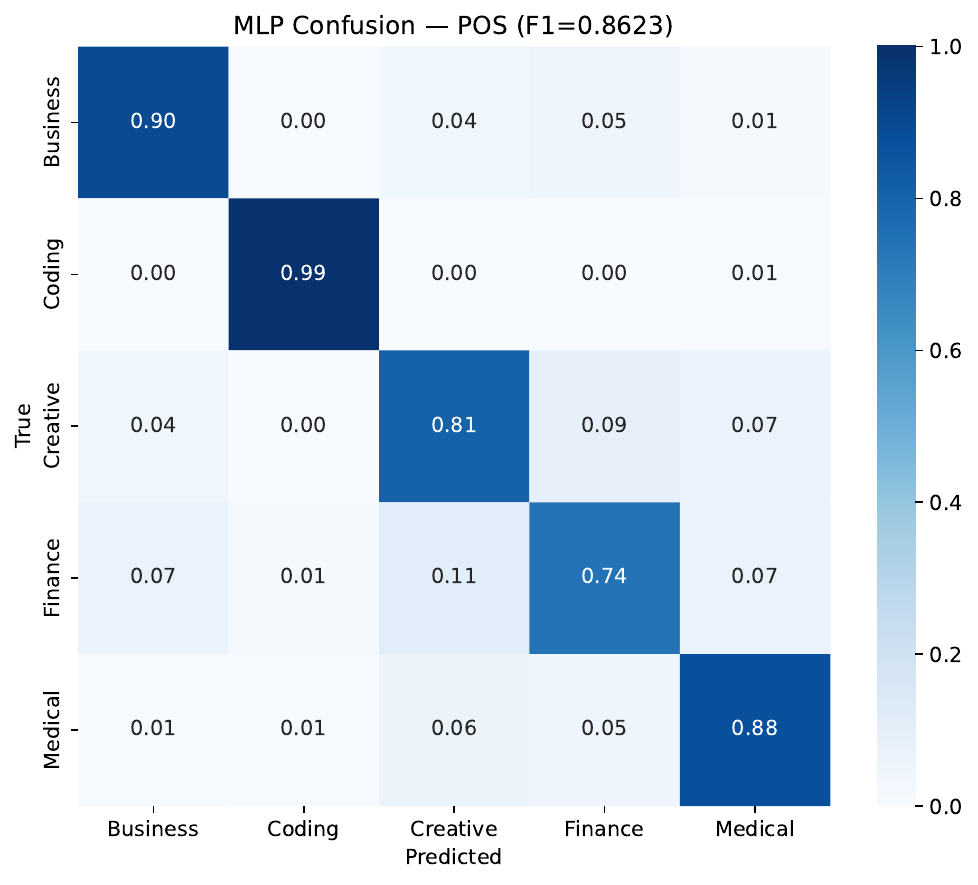}
        \caption{\small POS confusion with 18 dimensions.}
        \label{fig:domain_cm_pos}
    \end{subfigure}
    \hfill
    \begin{subfigure}[t]{0.32\textwidth}
        \centering
        \includegraphics[width=\linewidth]{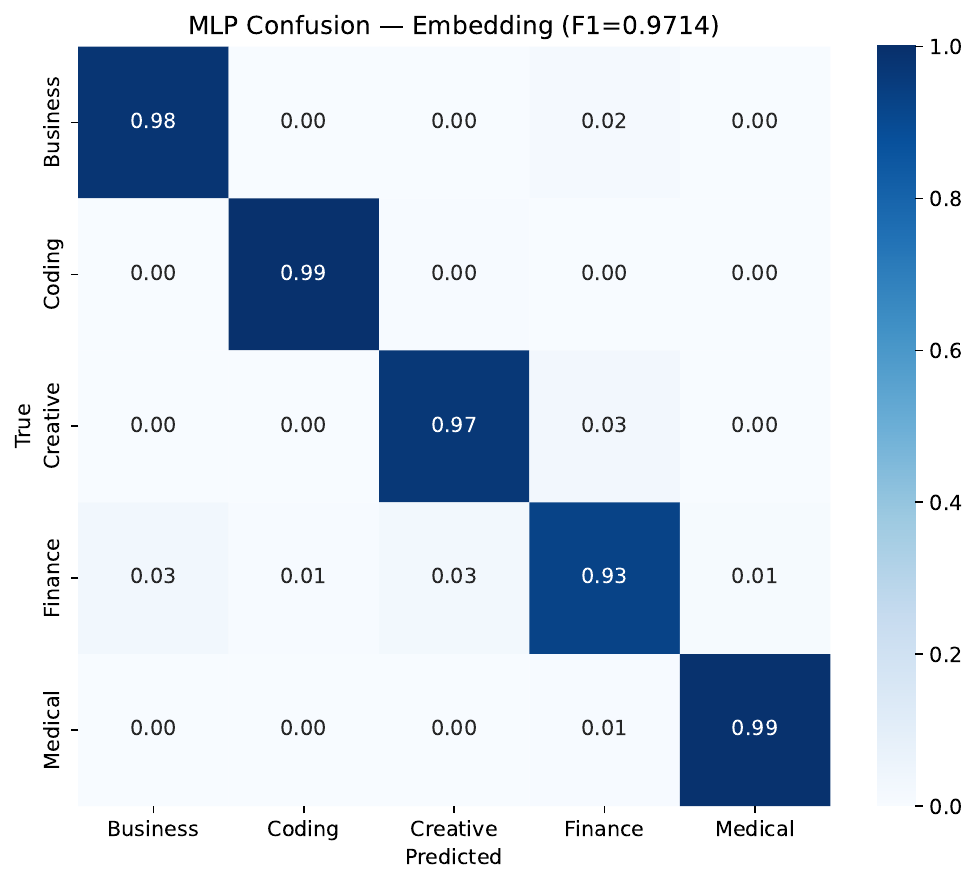}
        \caption{\small Embedding confusion with 384 dimensions.}
        \label{fig:domain_cm_emb}
    \end{subfigure}
    \hfill
    \begin{subfigure}[t]{0.32\textwidth}
        \centering
        \includegraphics[width=\linewidth]{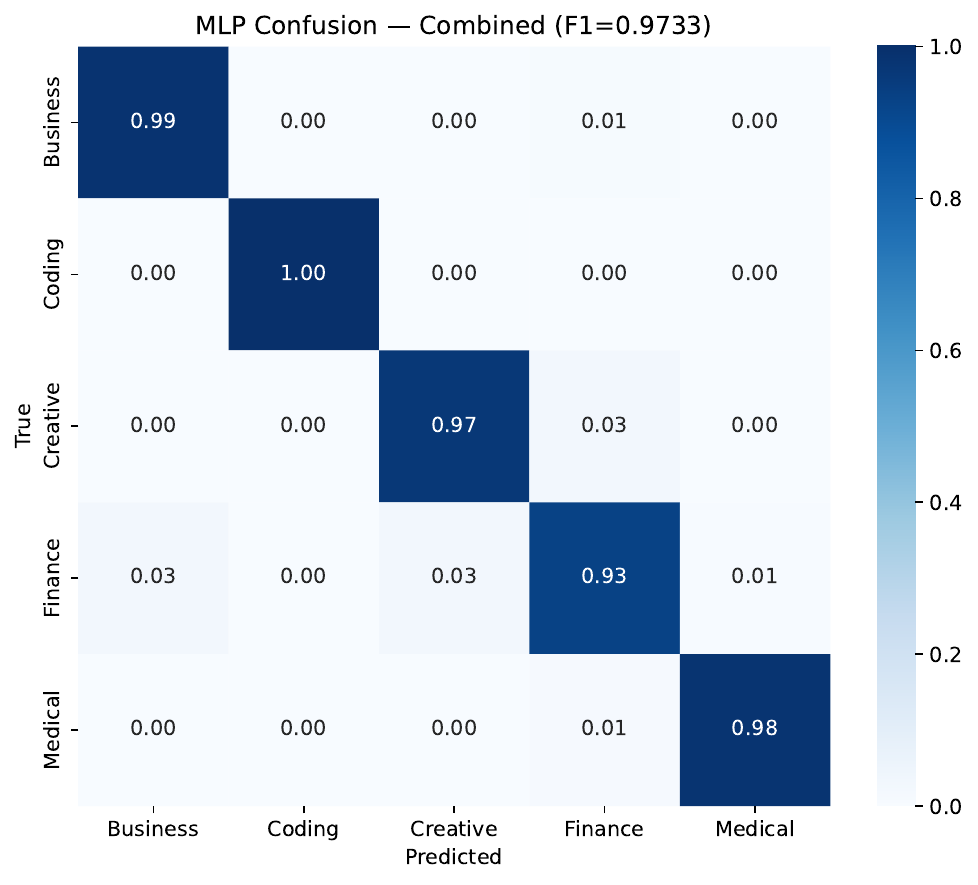}
        \caption{\small Combined confusion with 5.4k dimensions.}
        \label{fig:domain_cm_combined}
    \end{subfigure}
    \caption{\small Phase A source domain routing under five fold cross validation. Panels (a) and (b) show MLP training and Macro-F1 for each feature set. Panels (c) through (e) show confusion matrices for three feature sets. POS produces a strong diagonal pattern. Embedding and Combined further reduce confusion between Business and Finance and between Business and Creative.}
    \label{fig:domain_phaseA}
\end{figure*}

Figure~\ref{fig:domain_phaseA} shows the supervised Phase A results. The top row shows training, and the bottom row shows confusion matrices. Panel (a) shows stable convergence under early stopping. The Macro-F1 ranking in panel (b) matches Figure~\ref{fig:domain_phaseA_main}. Panels (c) through (e) compare POS, Embedding, and Combined. POS produces a strong diagonal pattern, especially for Medical and Coding. Embedding and Combined further reduce confusion between Business and Finance and between Business and Creative. These pairs account for nearly all remaining errors.

\begin{figure*}[t]
    \centering
    \begin{subfigure}[t]{0.32\textwidth}
        \centering
        \includegraphics[width=\linewidth]{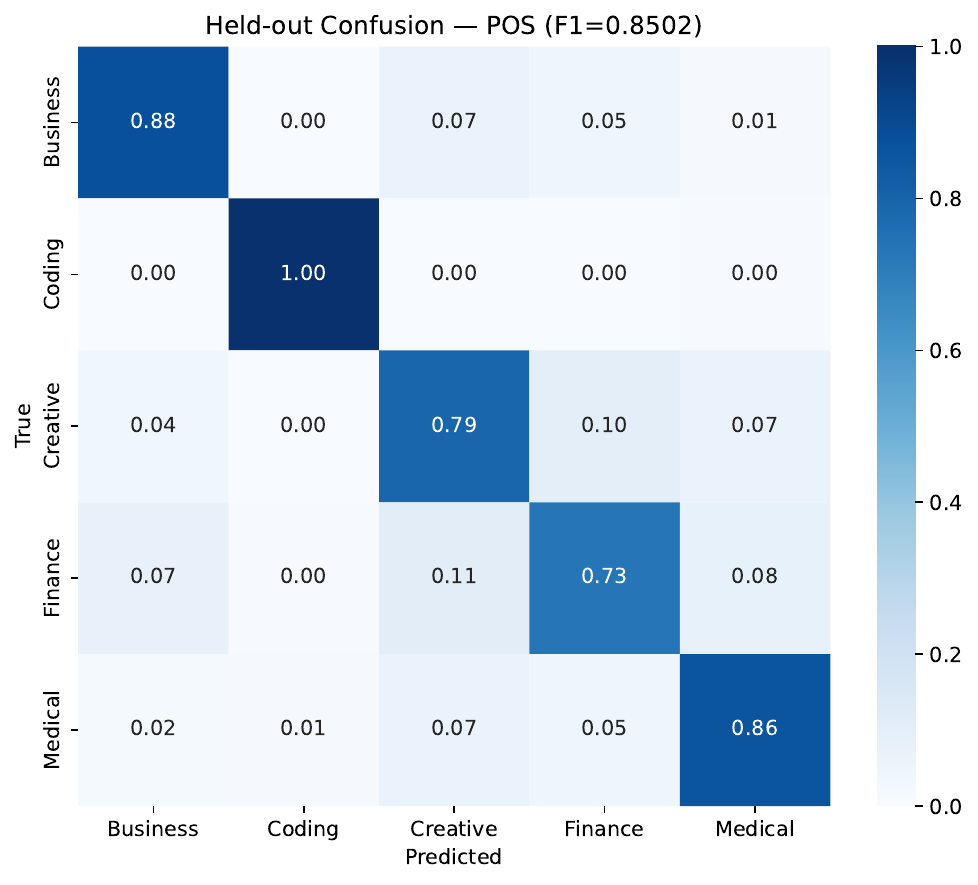}
        \caption{\small POS on the test split with 18 dimensions.}
        \label{fig:domain_hold_pos}
    \end{subfigure}
    \hfill
    \begin{subfigure}[t]{0.32\textwidth}
        \centering
        \includegraphics[width=\linewidth]{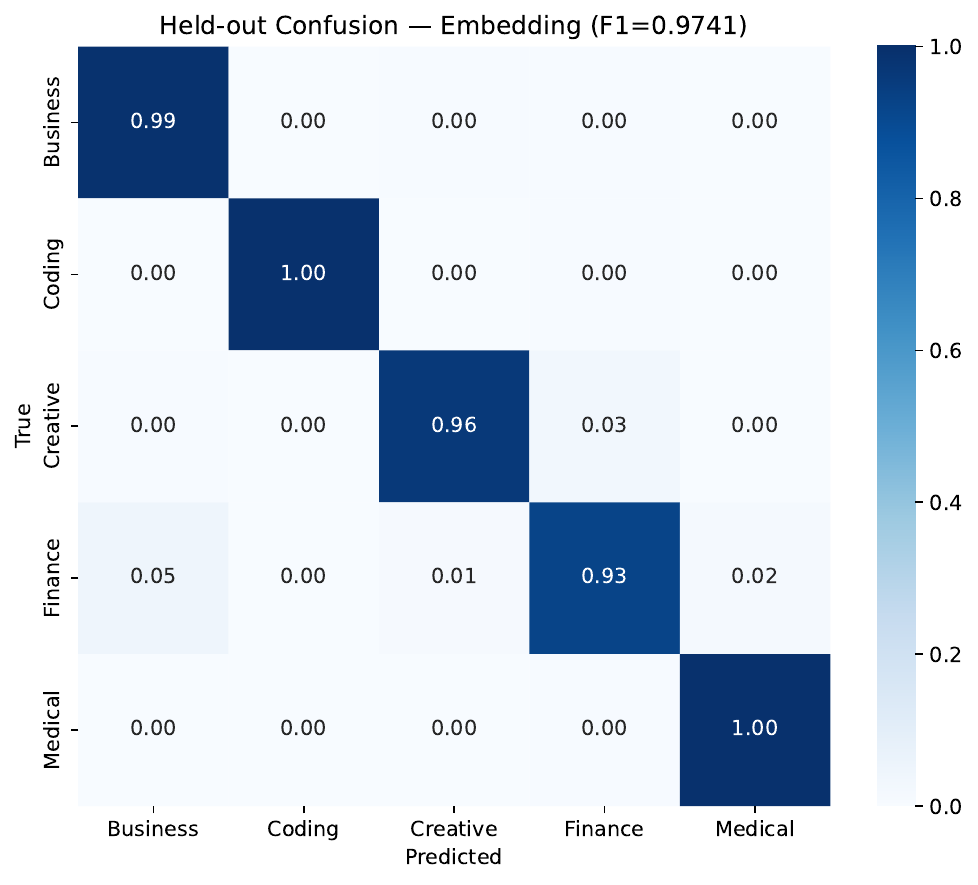}
        \caption{\small Embedding on the test split with 384 dimensions.}
        \label{fig:domain_hold_emb}
    \end{subfigure}
    \hfill
    \begin{subfigure}[t]{0.32\textwidth}
        \centering
        \includegraphics[width=\linewidth]{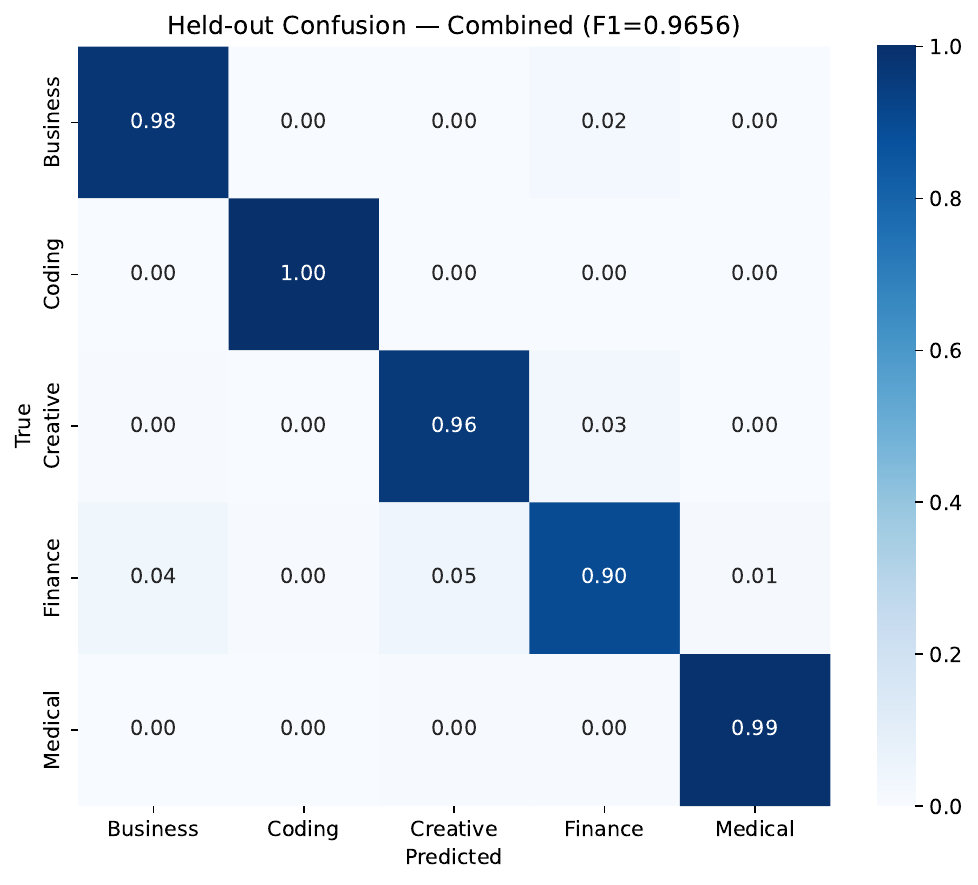}
        \caption{\small Combined on the test split with 5.4k dimensions.}
        \label{fig:domain_hold_combined}
    \end{subfigure}
    \caption{\small Phase C source domain routing on the 20\% test split. POS in panel (a) maintains an almost diagonal confusion matrix. Embedding in panel (b) and Combined in panel (c) further reduce confusion, especially between Business and Creative.}
    \label{fig:domain_holdout}
\end{figure*}

Figure~\ref{fig:domain_holdout} repeats the confusion analysis on the 20\% test split in Phase C. All three feature sets preserve the diagonal pattern. This result confirms that Phase A performance does not result from leakage across cross validation folds. POS has the most confusion between Business and Creative because both use command verbs such as ``write'' and ``draft.'' Embedding and Combined produce almost perfect confusion matrices.

\paragraph{Cross dataset transfer in Phase D.}
\label{sec:domain-cross-app}
Each domain and collection contributes up to 800 prompts with seed 42. Pooled cross validation uses five stratified folds over collections A and B. Transfer trains on all available examples from A and evaluates on target collection B. The full linguistic representation has 55 dimensions. It contains 17 POS rates, 27 dependency rates, and 11 basic statistics. The structural representation has 47 dimensions after removing eight features that depend on formatting. These representations differ from the Syntactic representation with 63 dimensions used in Phases A through C.

\begin{table}[t]
    \centering
    \caption{Phase D recall for each target collection. Table~\ref{tab:domain_cross_dataset} gives Macro-F1.}
    \label{tab:domain-cross-recall}
    \scriptsize
    \resizebox{\columnwidth}{!}{%
    \begin{tabular}{@{}lccccc@{}}
        \toprule
        \textbf{Representation} & \textbf{Bus.} & \textbf{Med.} & \textbf{Code} & \textbf{Fin.} & \textbf{Creative} \\
        \midrule
        Formatting (8 dim) & .000 & .561 & .860 & .249 & .101 \\
        Full (55 dim)       & .121 & .973 & .549 & .770 & .264 \\
        Structural (47 dim) & .192 & .938 & .995 & .751 & .243 \\
        Embedding (384 dim) & .938 & .968 & .821 & .945 & .201 \\
        \bottomrule
    \end{tabular}}
\end{table}

The large decrease from pooled cross validation confirms that Phase A performance partly reflects conventions specific to each collection. Removing formatting features improves overall syntactic transfer. Sentence embeddings transfer best. Creative recall is low because \textsf{WritingPrompts} contains story premises and the \textsf{No Robots} generation subset contains broader generation instructions. Syntactic features also give weak Business transfer, and sentence embeddings give strong Business transfer.

\begin{figure*}[t]
    \centering
    \includegraphics[width=\textwidth]{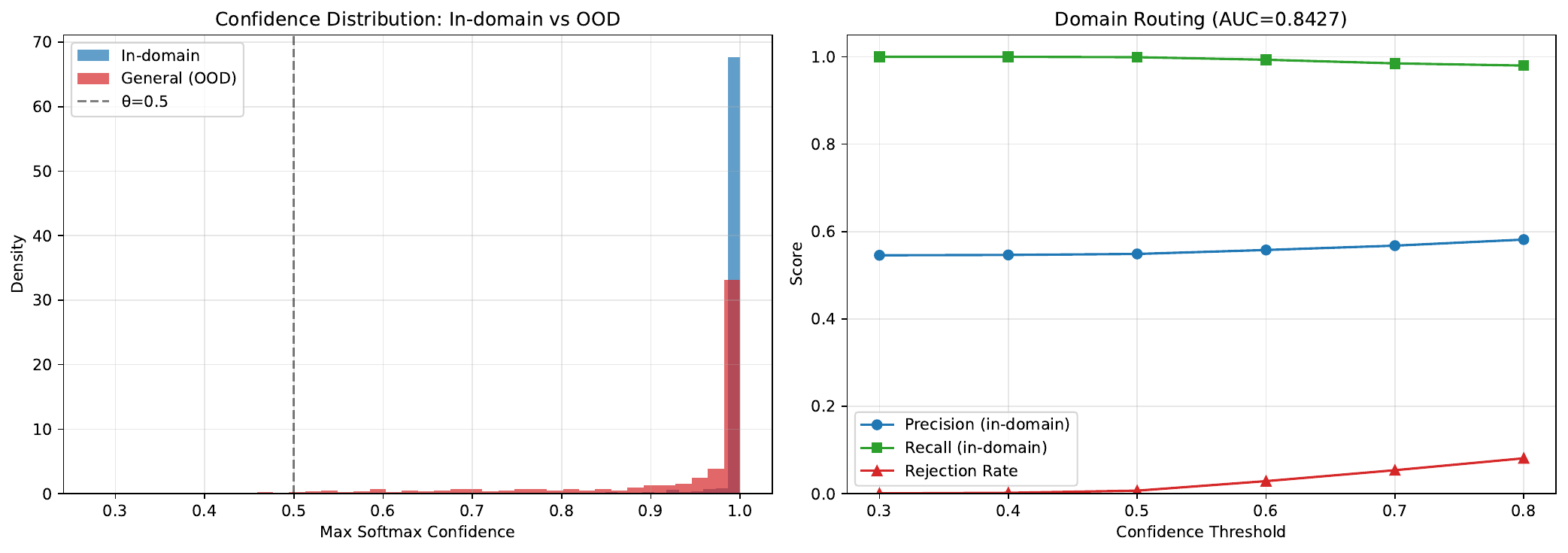}
    \caption{\small Routing gate based on confidence in the Phase C extension. The left panel shows maximum softmax confidence on prompts from the five domains and general prompts. The right panel shows precision, recall, and rejection rate across thresholds. Confidence separates the two groups with ROC-AUC\,=\,0.843~\citep{ong2024routellm,chen2023frugalgpt}.}
    \label{fig:domain_routing}
\end{figure*}

Figure~\ref{fig:domain_routing} extends Phase C with a routing gate based on confidence. The left panel compares maximum softmax confidence for 1{,}200 test prompts from the five domains and 1{,}000 general prompts from \textsf{ShareGPT} and \textsf{OASST1}. The right panel changes the decision threshold $\theta$ and reports precision, recall, and rejection rate. Confidence gives ROC-AUC 0.843. This value shows useful separation with some remaining errors and no additional distribution detector~\citep{ong2024routellm,chen2023frugalgpt}.

\begin{figure*}[t]
    \centering
    \includegraphics[width=0.75\textwidth]{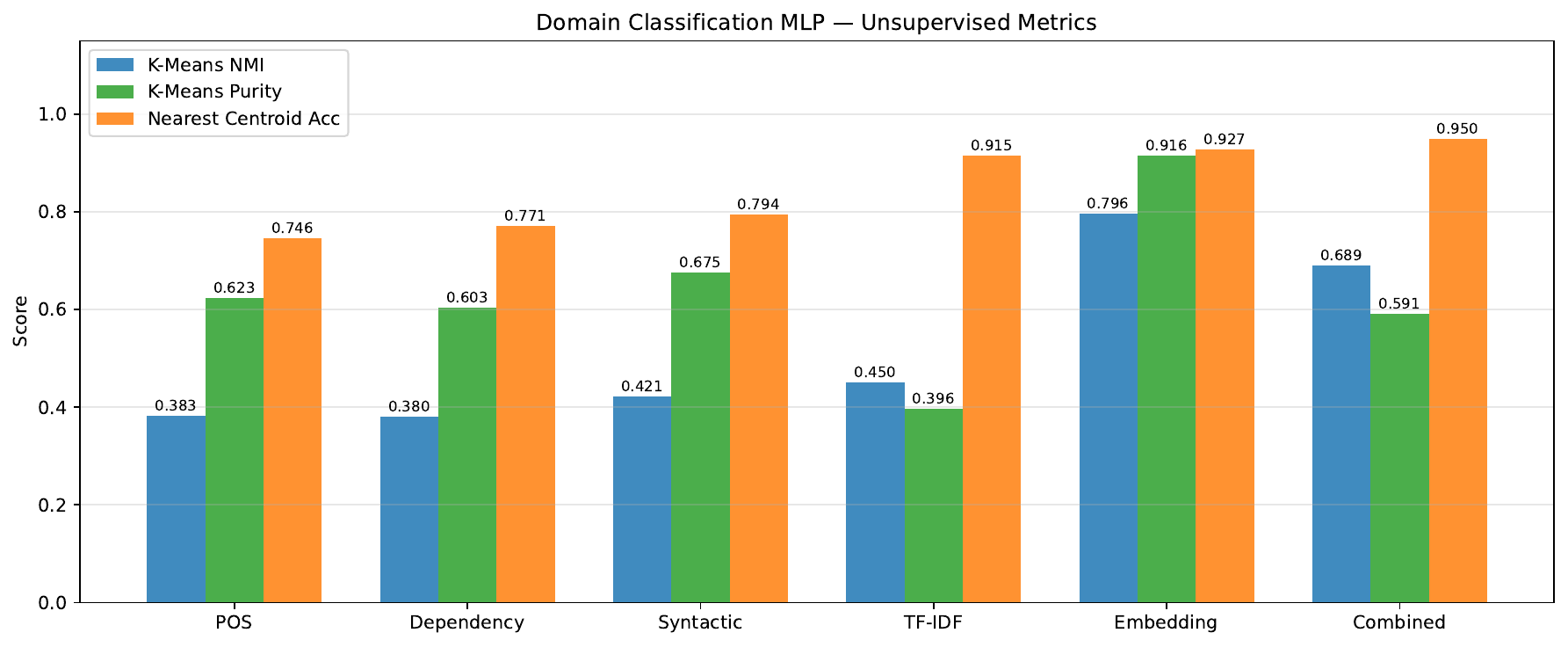}
    \caption{\small Phase B source domain clustering and centroid classification. Unsupervised K-Means with sentence embeddings forms clear clusters for the five domains and reaches high NMI, ARI, and Purity. The supervised Nearest Centroid accuracy is reported as a simple geometric baseline. The high dimensional Combined representation gives weaker distance based clusters.}
    \label{fig:domain_unsup_fig}
\end{figure*}

Figure~\ref{fig:domain_unsup_fig} shows the Phase B metrics from Table~\ref{tab:domain_unsup} for all six feature sets. Sentence embeddings lead every clustering metric. TF-IDF and Combined give lower scores even though they include the full vocabulary with 5{,}000 words. POS, Dependency, and Syntactic form a lower group. This result supports the claim in \S\ref{sec:domain} that semantic embedding geometry captures domain identity more strongly than lexical overlap.

\subsubsection{Feature Importance}
\label{sec:importance-app}

Table~\ref{tab:importance} summarises mean absolute gradient importance. The score $\mathrm{Imp}(i)$ is the expected absolute partial derivative $\mathbb{E}_x[|\partial\,\mathrm{logit}_c(x)/\partial x_i|]$ averaged across classes. A large value means that small changes to feature $i$ produce large changes in classifier logits. Each task has a global ranking, a heatmap for each class, and a radar chart for the leading features.

\begin{table*}[t]
    \centering
    \caption{Five largest gradient importance values for Prompt Filtering and Source Domain Routing.}
    \label{tab:importance}
    \scriptsize
    \resizebox{\textwidth}{!}{
        \begin{tabular}{@{}ll|ll@{}}
            \toprule
            \multicolumn{2}{c|}{\textbf{Prompt Filtering}} & \multicolumn{2}{c}{\textbf{Source Domain Routing}} \\
            \textbf{Feature set} & \textbf{Five largest values} & \textbf{Feature set} & \textbf{Five largest values} \\
            \midrule
            POS & SPACE(.766), PRON(.468), ADV(.461), NOUN(.394), PROPN(.346) & POS & SPACE(2.07), X(.871), PROPN(.829), NUM(.812), NOUN(.661) \\
            Dependency & dep(.616), meta(.544), attr(.529), aux(.492), nsubj(.474) & Dependency & dep(1.85), ROOT(1.26), compound(.747), nummod(.634), aux(.507) \\
            Syntactic & nsubj(.648), SPACE(.640), attr(.622), PRON(.594), ADP(.561) & Syntactic & SPACE(1.69), ROOT(1.11), X(.961), dep(.639), NOUN(.602) \\
            \bottomrule
        \end{tabular}}
\end{table*}

\begin{figure*}[t]
    \centering
    \begin{subfigure}[t]{0.32\textwidth}
        \centering
        \includegraphics[width=\linewidth]{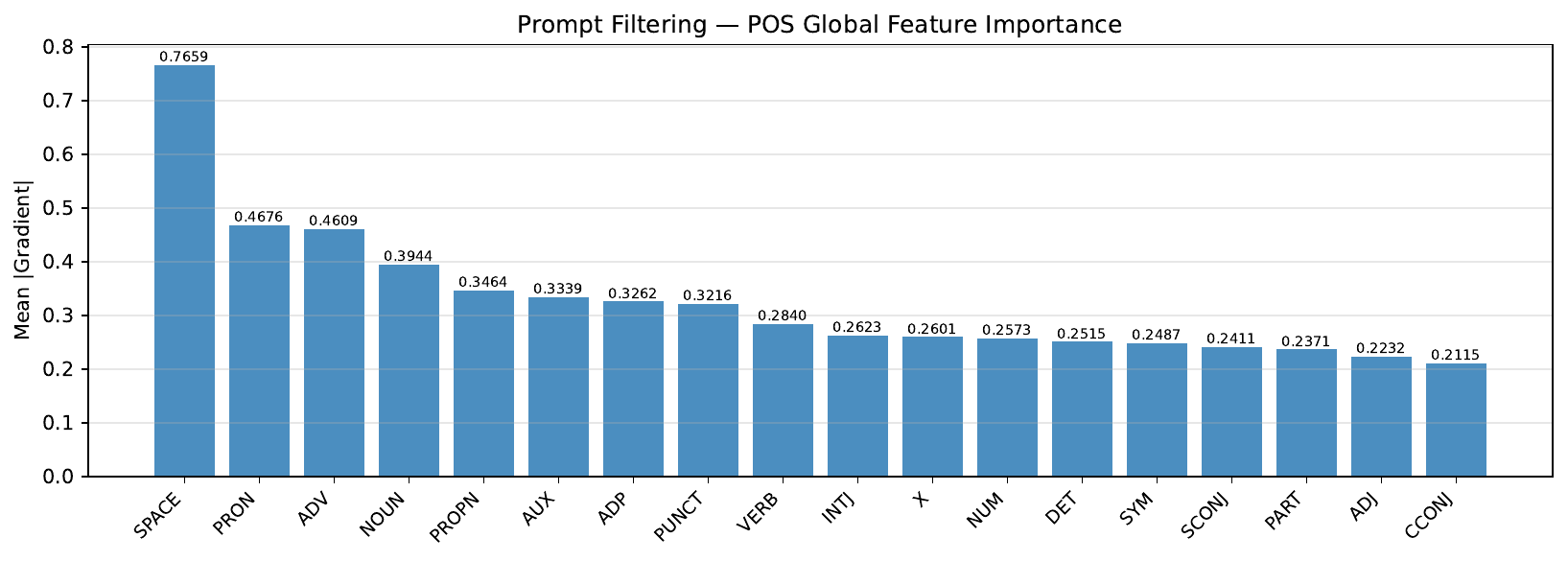}
        \caption{\small Global POS importance for filtering.}
        \label{fig:imp_filt_pos}
    \end{subfigure}
    \hfill
    \begin{subfigure}[t]{0.32\textwidth}
        \centering
        \includegraphics[width=\linewidth]{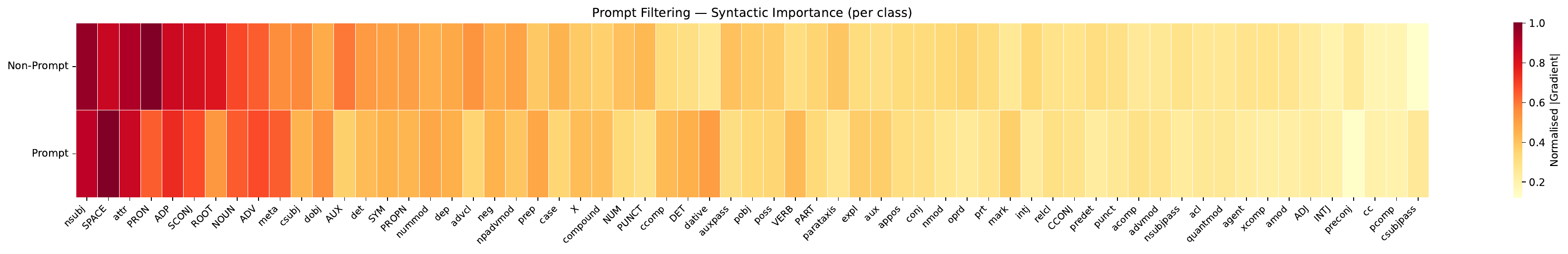}
        \caption{\small Syntactic heatmap for filtering.}
        \label{fig:imp_filt_heat}
    \end{subfigure}
    \hfill
    \begin{subfigure}[t]{0.32\textwidth}
        \centering
        \includegraphics[width=\linewidth]{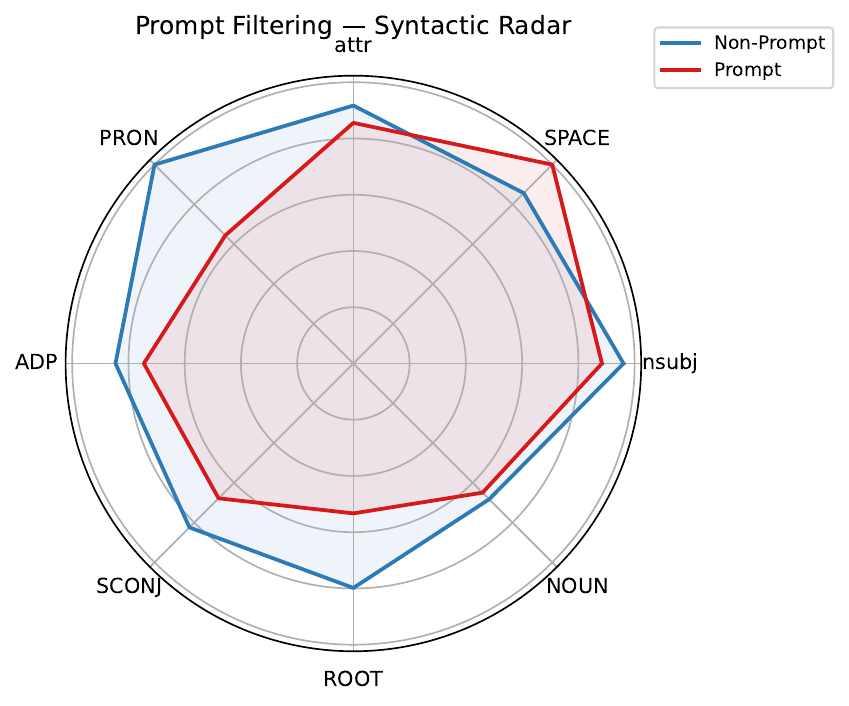}
        \caption{\small Syntactic radar chart for filtering.}
        \label{fig:imp_filt_radar}
    \end{subfigure}
    \caption{\small Gradient importance for \textbf{prompt filtering}. Panel (a) shows that \texttt{SPACE}, \texttt{PRON}, and \texttt{ADV} lead the POS ranking. Panel (b) shows the per-class gradients in the 63 dimensional Syntactic representation; its global leaders are \texttt{nsubj}, \texttt{SPACE}, and \texttt{attr}. Panel (c) summarises the leading Syntactic features.}
    \label{fig:importance_filtering}
\end{figure*}

Figure~\ref{fig:importance_filtering} reports importance for prompt filtering. Panel (a) ranks the 18 POS tags by global gradient magnitude. \texttt{SPACE} captures whitespace in numbered steps, code blocks, and bullet lists. \texttt{PRON} captures second person pronouns in commands, and \texttt{ADV} captures directives such as ``step by step.'' These three tags have the largest values in the POS-only representation. Panel (b) shows importance for each class in the Syntactic representation with 63 dimensions. Its global top five are \texttt{nsubj} (.648), \texttt{SPACE} (.640), \texttt{attr} (.622), \texttt{PRON} (.594), and \texttt{ADP} (.561). The general \texttt{dep} arc instead leads the separate Dependency-only representation at .616. Panel (c) shows the leading Syntactic features in a radar chart.

\begin{figure*}[t]
    \centering
    \begin{subfigure}[t]{0.245\textwidth}
        \centering
        \includegraphics[width=\linewidth]{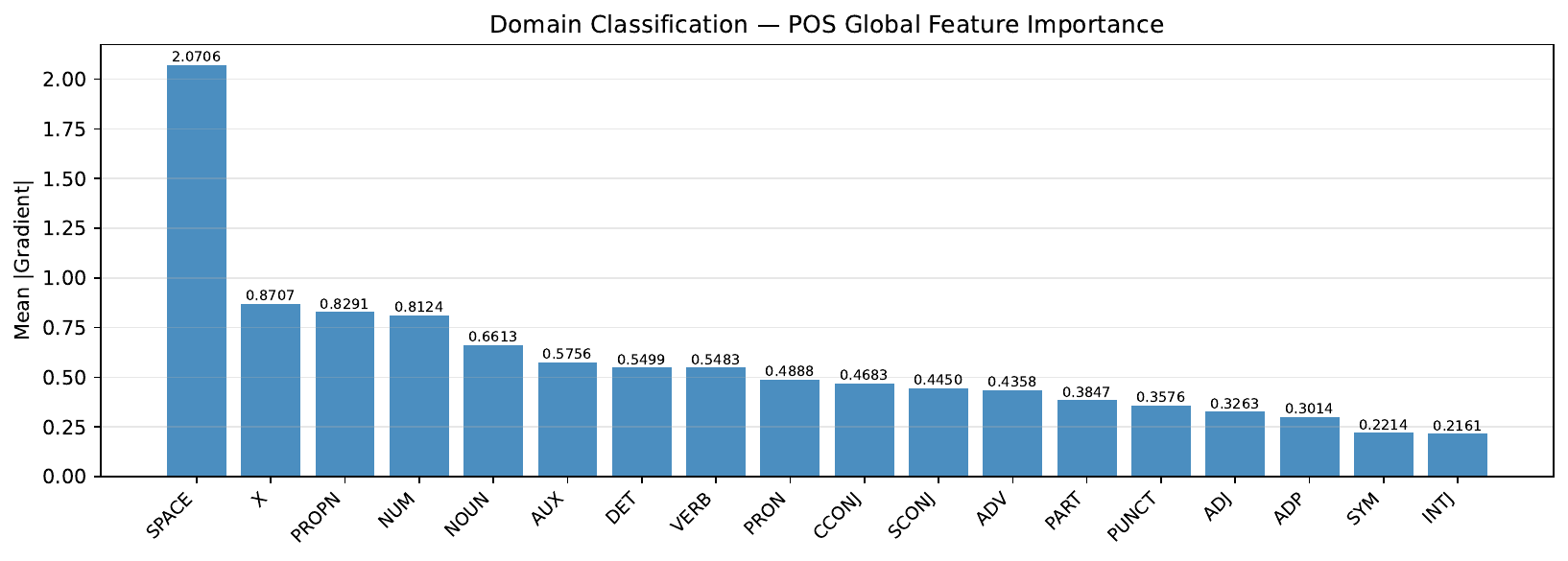}
        \caption{\small Global POS importance for routing.}
        \label{fig:imp_dom_pos}
    \end{subfigure}
    \hfill
    \begin{subfigure}[t]{0.245\textwidth}
        \centering
        \includegraphics[width=\linewidth]{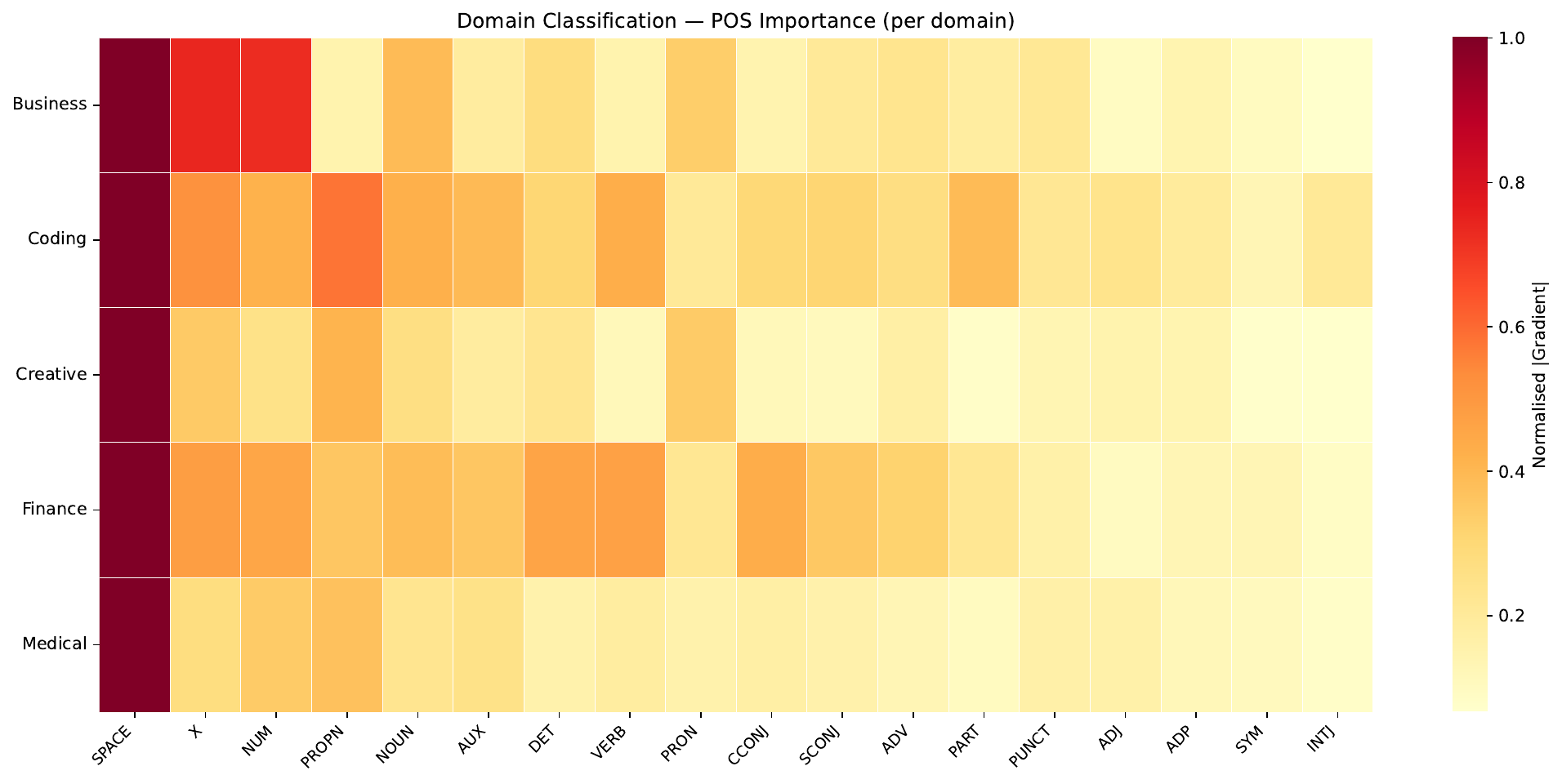}
        \caption{\small POS heatmap for routing.}
        \label{fig:imp_dom_heat}
    \end{subfigure}
    \hfill
    \begin{subfigure}[t]{0.245\textwidth}
        \centering
        \includegraphics[width=\linewidth]{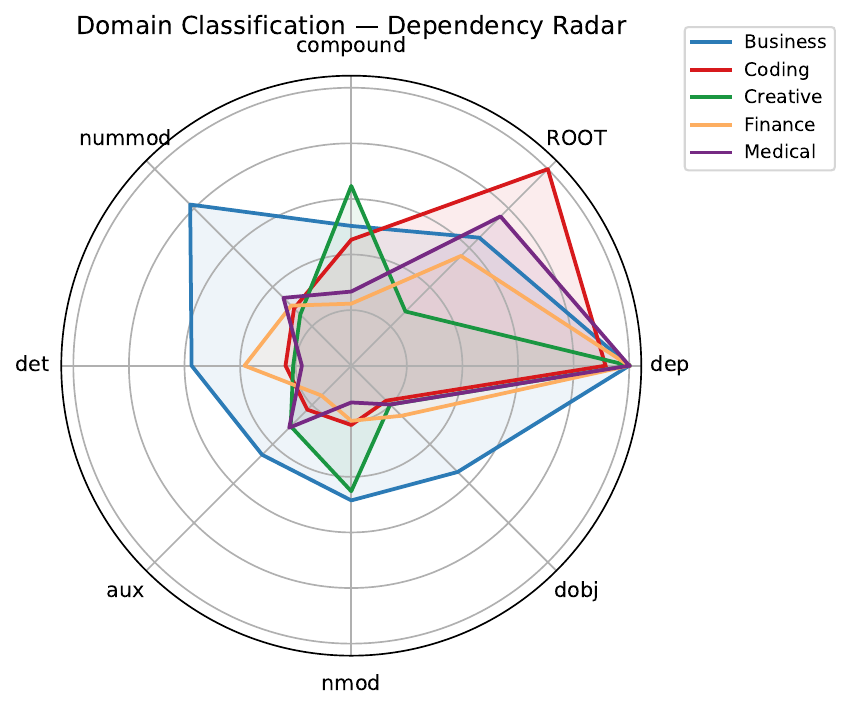}
        \caption{\small Dependency radar chart for routing.}
        \label{fig:imp_dom_radar}
    \end{subfigure}
    \hfill
    \begin{subfigure}[t]{0.245\textwidth}
        \centering
        \includegraphics[width=\linewidth]{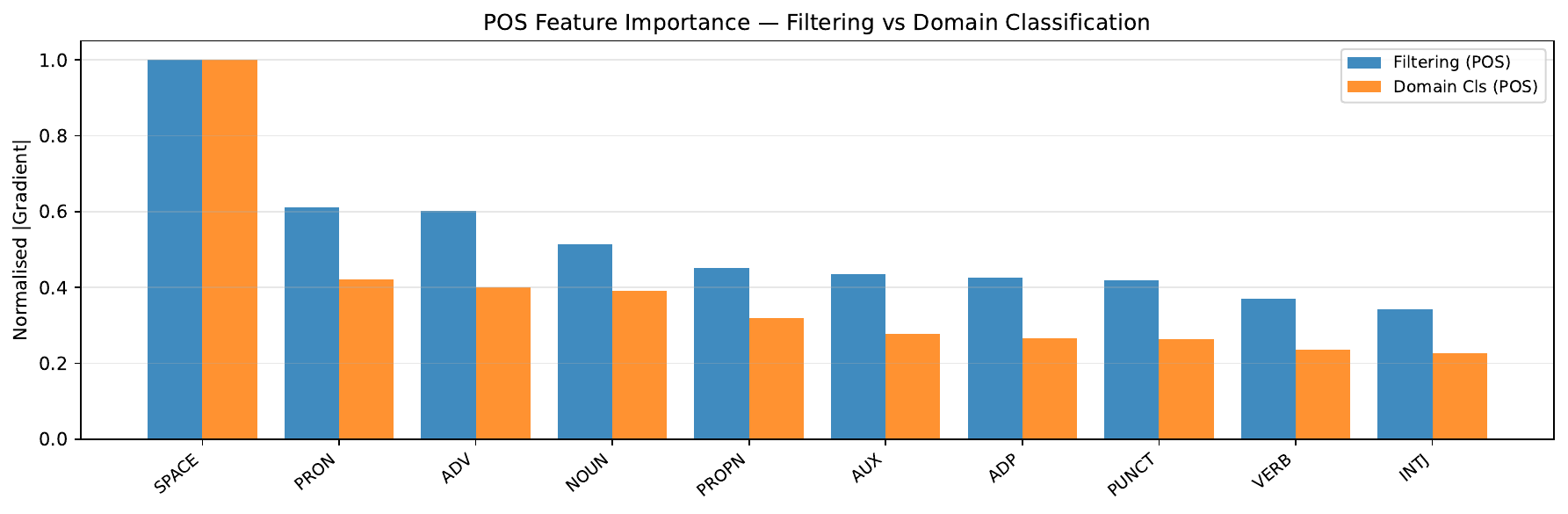}
        \caption{\small POS comparison across tasks.}
        \label{fig:imp_crosstask}
    \end{subfigure}
    \caption{\small Gradient importance for \textbf{source domain routing}. Panel (a) shows that the \texttt{X} tag is the second largest POS signal. This tag captures unknown vocabulary tokens such as code identifiers, medical abbreviations, and financial tickers. Panel (b) shows that each domain uses a distinct POS subset. Panel (c) shows that \texttt{dep}, \texttt{ROOT}, and \texttt{compound} are the main dependency features for domains. Panel (d) shows that POS gradient magnitudes are two to three times larger for source domain routing than for prompt filtering.}
    \label{fig:importance_domain}
\end{figure*}

Figure~\ref{fig:importance_domain} reports three views for source domain routing and compares the two tasks in a fourth panel. Panel (a) ranks POS tags across all domains. The \texttt{X} tag captures unknown vocabulary tokens such as the code identifier \texttt{np.array}, the medical abbreviation \emph{CBC}, and the financial ticker \emph{AAPL}. It is the second largest signal after \texttt{SPACE}. Panel (b) shows a distinct POS subset for each domain. \texttt{NUM} and \texttt{nummod} lead in Finance, \texttt{X} leads in Coding, and \texttt{ADJ} leads in Medical. Differences across cells show that one common prompt style cannot explain all five domains. Panel (c) shows \texttt{dep}, \texttt{ROOT}, \texttt{compound}, \texttt{nummod}, and \texttt{aux} in a radar chart. Together they capture technical noun phrases and imperative root verbs across domains. Panel (d) compares POS importance between the tasks. Gradient magnitudes are two to three times larger for source domain routing than for prompt filtering. This difference supports using the Syntactic representation first for routing and then for prompt detection when both tasks are needed.

\subsubsection{Elicited Response Quality}
\label{sec:quality-app}

\begin{figure*}[t]
    \centering
    \begin{subfigure}[t]{0.32\textwidth}
        \centering
        \includegraphics[width=\linewidth]{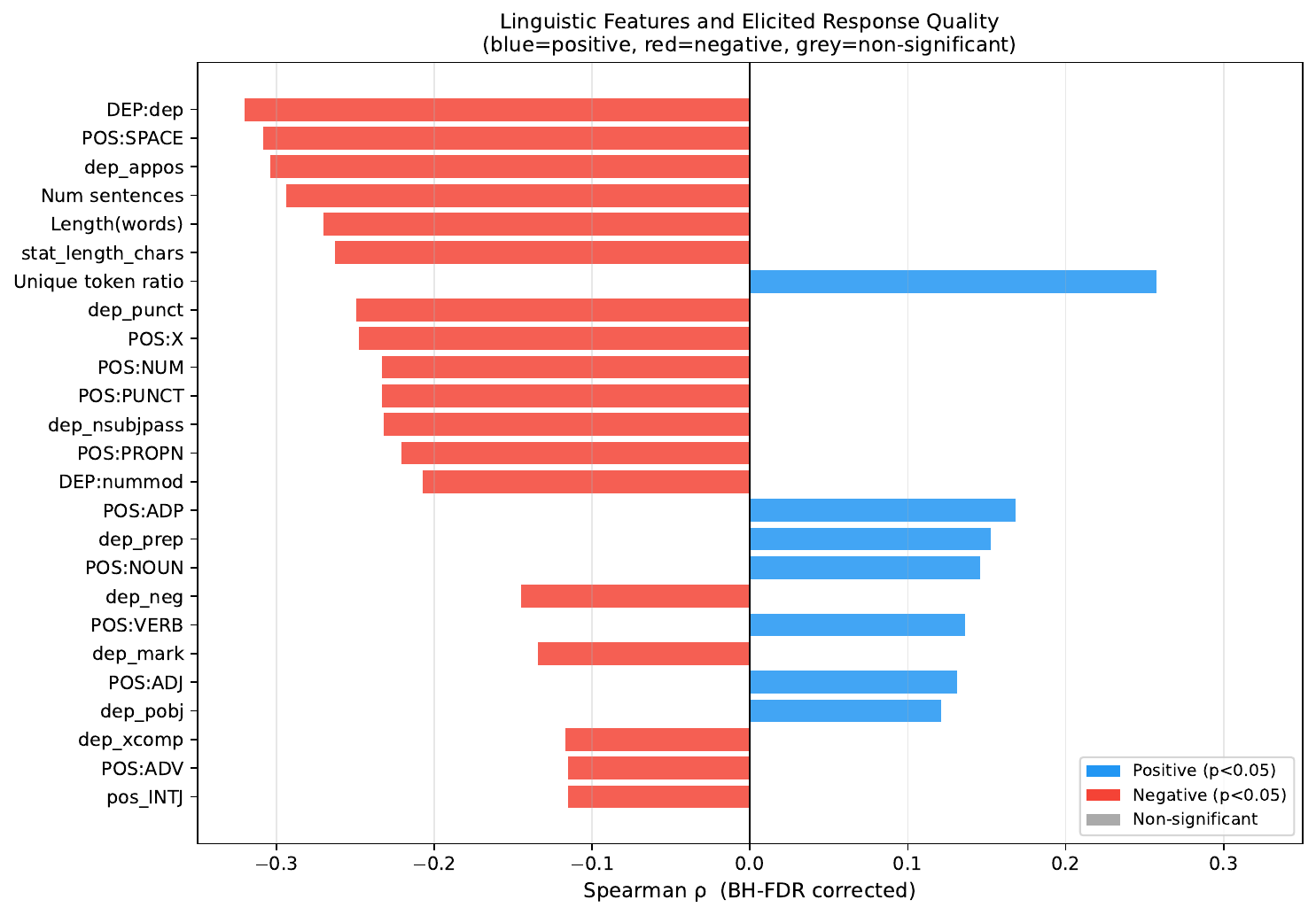}
        \caption{\small Spearman $\rho$ with elicited response quality.}
        \label{fig:qual_corr}
    \end{subfigure}
    \hfill
    \begin{subfigure}[t]{0.32\textwidth}
        \centering
        \includegraphics[width=\linewidth]{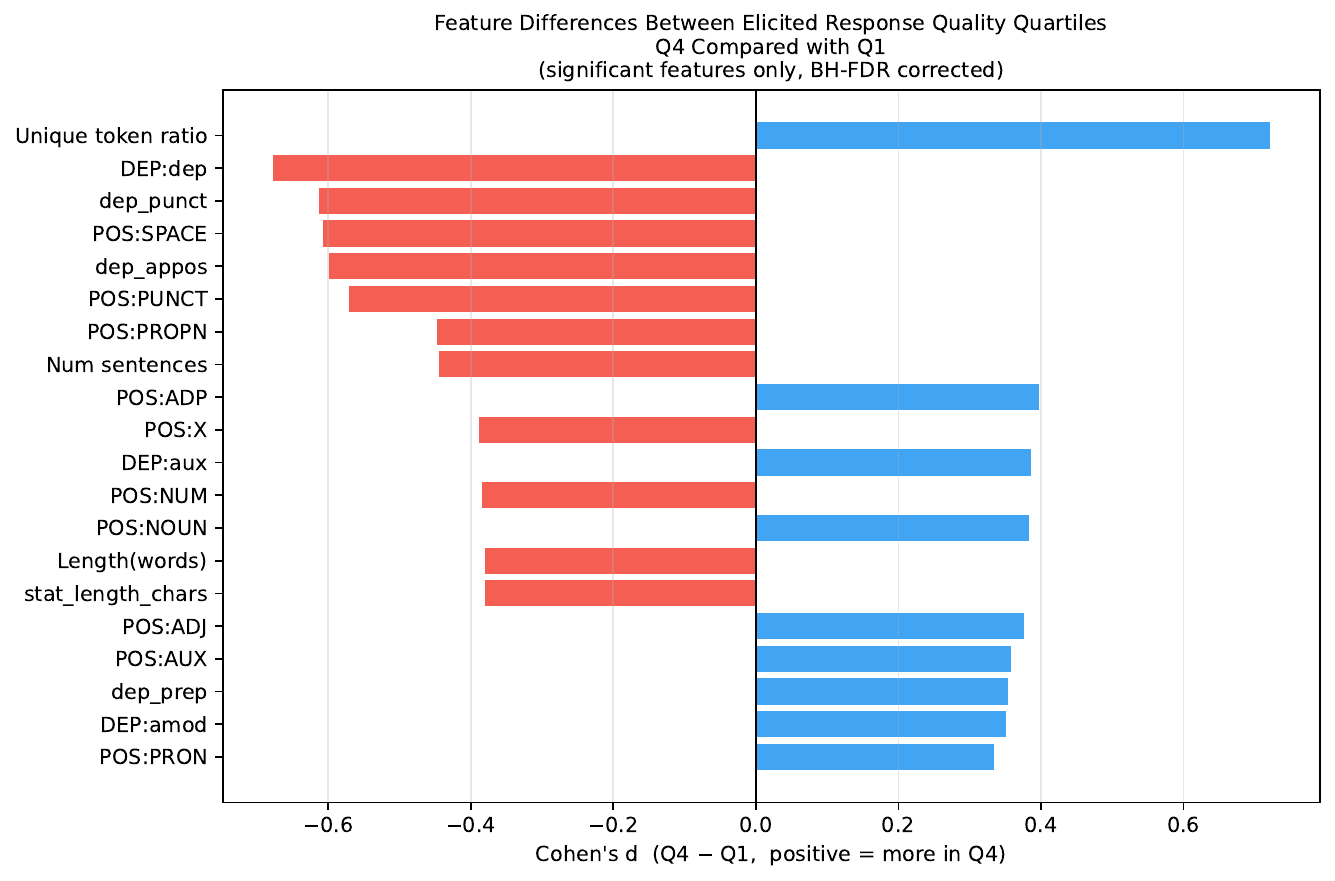}
        \caption{\small Cohen's $d$ for Q4 compared with Q1.}
        \label{fig:qual_cohen}
    \end{subfigure}
    \hfill
    \begin{subfigure}[t]{0.32\textwidth}
        \centering
        \includegraphics[width=\linewidth]{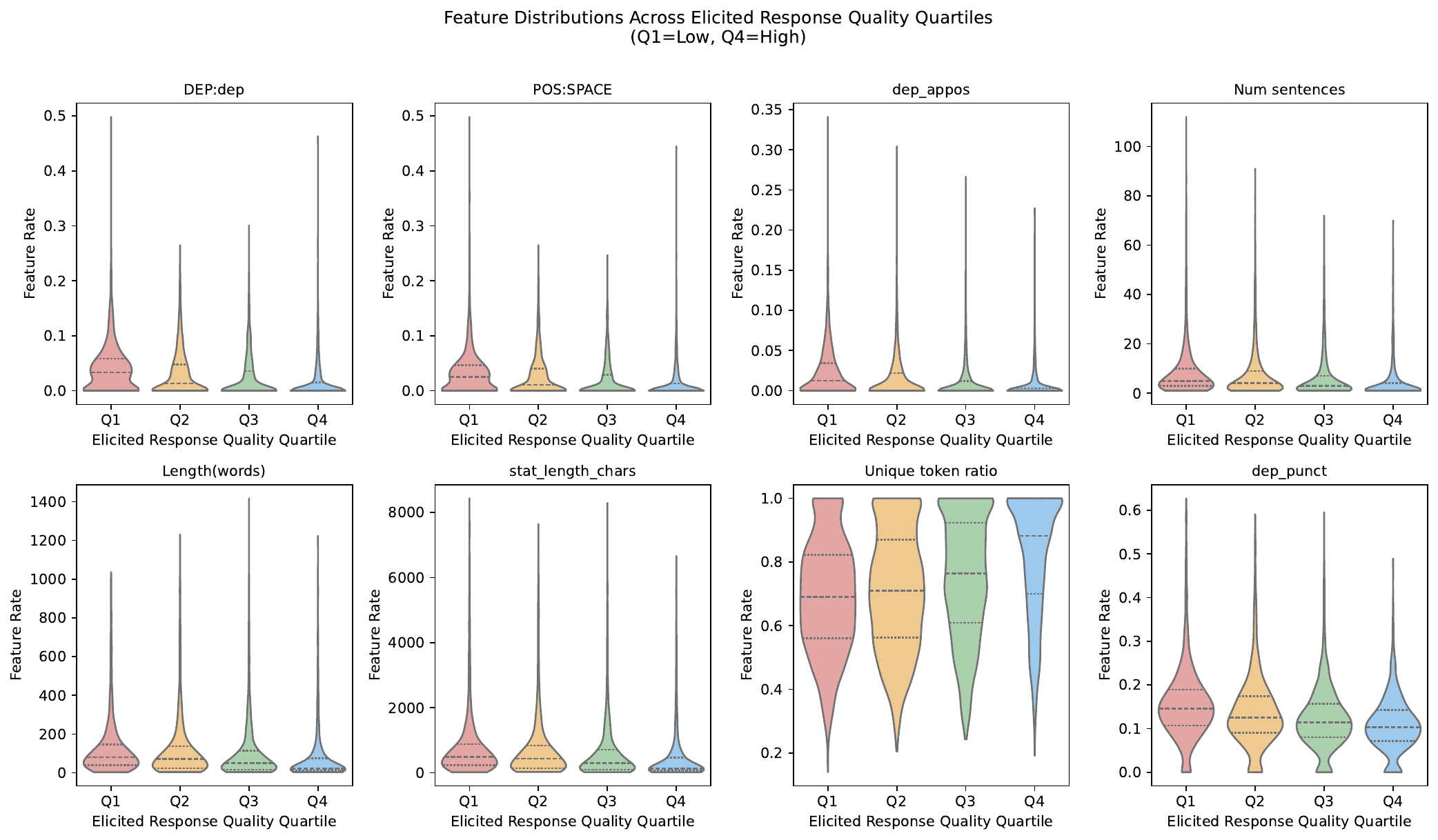}
        \caption{\small Feature distribution by elicited response quality quartile.}
        \label{fig:qual_violin}
    \end{subfigure}
    \\[.4em]
    \begin{subfigure}[t]{0.32\textwidth}
        \centering
        \includegraphics[width=\linewidth]{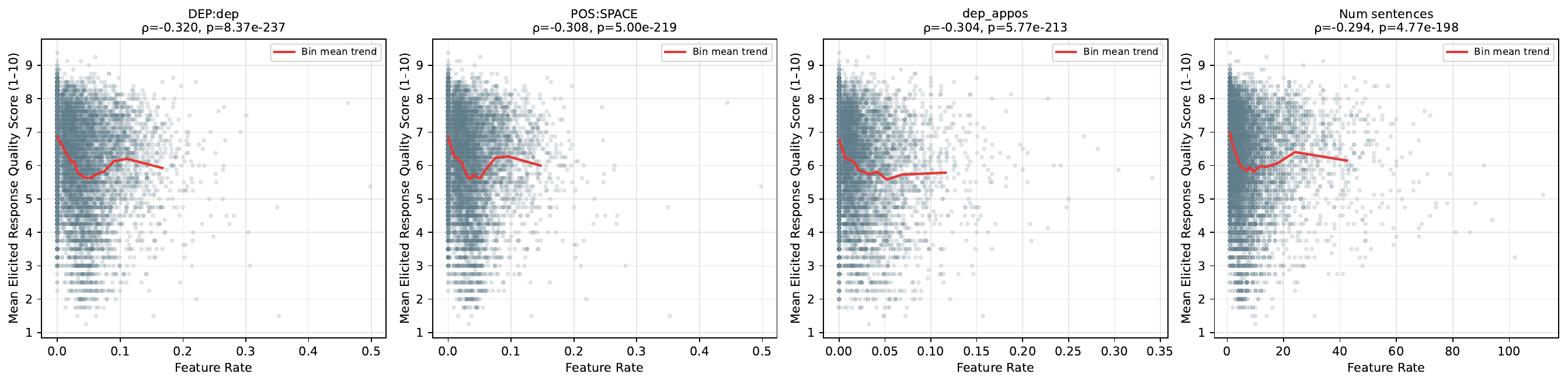}
        \caption{\small Four leading features and elicited response quality.}
        \label{fig:qual_scatter}
    \end{subfigure}
    \hfill
    \begin{subfigure}[t]{0.32\textwidth}
        \centering
        \includegraphics[width=\linewidth]{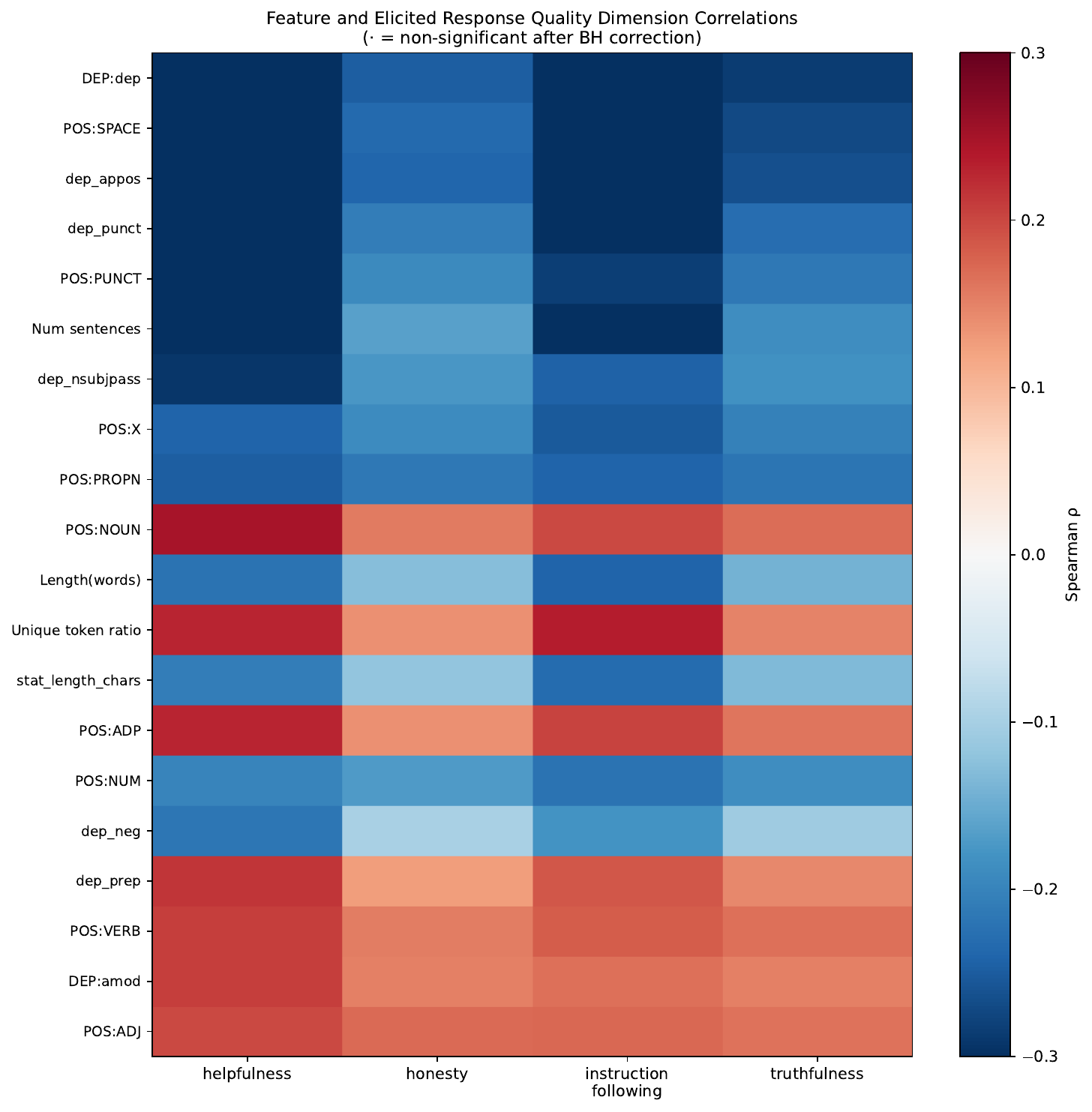}
        \caption{\small Correlation heatmap for each rating dimension.}
        \label{fig:qual_heat}
    \end{subfigure}
    \hfill
    \begin{subfigure}[t]{0.32\textwidth}
        \centering
        \includegraphics[width=\linewidth]{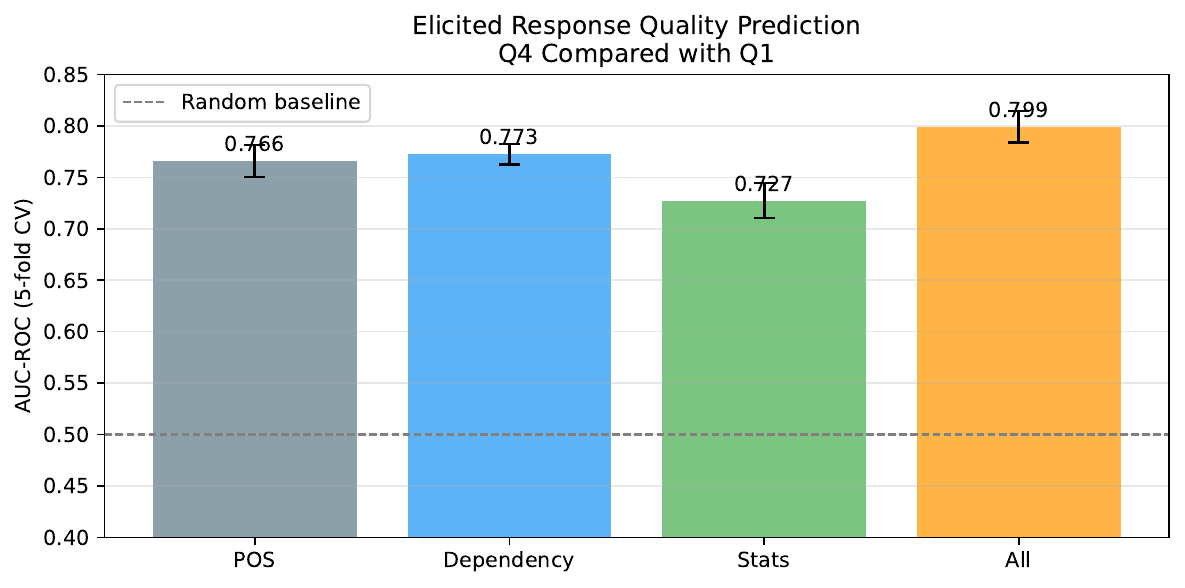}
        \caption{\small Q1 vs.\ Q4 prediction AUC.}
        \label{fig:qual_auc}
    \end{subfigure}
    \caption{\small \textbf{UltraFeedback diagnostics before length control.} Panels (a) through (c) show Spearman $\rho$, Cohen's $d$ for Q4 compared with Q1, and quartile distributions. Panels (d) through (f) show scatter plots, correlations across GPT-4 rating dimensions, and prediction AUC for Q1 compared with Q4 within the dataset. These panels do not control for prompt length.}
    \label{fig:quality_overview}
\end{figure*}

Figure~\ref{fig:quality_overview} presents the original UltraFeedback diagnostics before length control. The top row reports associations and differences between score groups. The bottom row reports prediction of Q1 compared with Q4 within UltraFeedback. These panels do not control for prompt length.

\textit{Top row.} Panel (a) shows Spearman $\rho$ with mean UltraFeedback scores before length control. Type to token ratio has a positive correlation of $+0.257$. \texttt{dep}, \texttt{appos}, \texttt{SPACE}, sentence count, and word count have the largest negative correlations. Panel (b) shows Cohen's $d$ for Q4 compared with Q1. Panel (c) shows the distribution of each feature by quartile. The inverse relation between type to token ratio and prompt length explains most of its visible separation. The partial correlations below show this result.

\textit{Bottom row.} Panel (d) shows scatter plots before length control. Panel (e) reports correlations for each GPT-4 rating dimension. Panel (f) reports logistic regression AUC for Q1 compared with Q4. POS, Dependency, and basic statistics reach AUC 0.766, 0.773, and 0.727. All features together reach 0.799. These results measure prediction within UltraFeedback. They provide no causal estimate or intrinsic prompt score.

\paragraph{Human ratings and length control.}
\label{sec:quality-robustness-app}
We sample 2{,}000 unique HelpSteer2 prompts with seed 42 and average human helpfulness scores for each prompt. For both datasets, we rank transform each variable. We remove the linear association of prompt word count from the feature and rating ranks. We then compute Spearman correlations between the residuals.

\begin{table*}[t]
    \centering
    \caption{Spearman correlations with elicited response quality before and after word count control. ``n.s.'' marks $p\geq .05$.}
    \label{tab:quality-length-control}
    \scriptsize
    \begin{tabular}{@{}lrr|rr@{}}
        \toprule
        & \multicolumn{2}{c|}{\textbf{UltraFeedback ($n=10{,}000$)}}
        & \multicolumn{2}{c}{\textbf{HelpSteer2 ($n=2{,}000$)}} \\
        \textbf{Feature} & \textbf{Before control} & \textbf{Partial} & \textbf{Before control} & \textbf{Partial} \\
        \midrule
        Type to token ratio & +.257 & +.024 & +.150 & $-$.017 n.s. \\
        \texttt{dep}      & $-$.320 & $-$.212 & $-$.128 & $-$.015 n.s. \\
        \texttt{SPACE}    & $-$.308 & $-$.197 & $-$.127 & $-$.010 n.s. \\
        \texttt{X}        & $-$.248 & $-$.153 & $-$.077 & +.052 \\
        \texttt{NUM}      & $-$.233 & $-$.136 & $-$.172 & $-$.116 \\
        Word count         & $-$.270 & --- & $-$.166 & --- \\
        \bottomrule
    \end{tabular}
\end{table*}

HelpSteer2 reproduces the directions before length control, so the pattern also appears in human ratings. Length control changes the interpretation. The association for type to token ratio becomes nearly zero in both datasets. The structural features remain negative in UltraFeedback. In HelpSteer2, only \texttt{NUM} remains negative, and \texttt{X} becomes positive after control. We therefore limit the claim to observational associations that depend on the dataset. Prompt length, task difficulty, response model capability, and judge preferences may all affect the measured outcome.

\subsection{Zero- and Few-shot LLM Baselines}
\label{sec:llm-baselines-app}

We query \texttt{openai/gpt-4o-mini} through the OpenRouter chat-completions endpoint with temperature 0, seed 42, and JSON-object output. The retry path uses a strict JSON schema when deterministic malformed output persists. The protocol fixes 500 evaluation examples per task. Prompt filtering is balanced between 250 prompts, source-stratified across the seven analysed datasets, and 250 general texts from blog, Reddit, and Wikipedia. Source domain routing contains 100 examples per domain from the five Phase D target collections, so evaluation remains cross-dataset relative to the original training sources. Elicited response quality uses a score-stratified 500-prompt subset of UltraFeedback. Few-shot prompts use four demonstrations for filtering, five for routing, and ten for score prediction; demonstrations are disjoint from evaluation examples and, for routing, come from the original rather than target collections. These subsets match the label spaces and source types of the corresponding applications but not their exact evaluation instances. The effectiveness values are therefore not paired comparisons with the main supervised results.

\begin{table*}[t]
    \centering
    \caption{Zero- and few-shot LLM baselines on fixed 500-example task samples. Latency is median end-to-end API time per example. Cost is estimated from recorded input/output tokens at \$0.15/\$0.60 per million tokens. For score prediction, Spearman $\rho$, Q1--Q4 AUC, and MAE are computed from the same API outputs. Latency and cost are reported once for this run; dashes denote that no additional inference is required.}
    \label{tab:llm-baselines}
    \scriptsize
    \resizebox{\textwidth}{!}{%
    \begin{tabular}{@{}llrrrrrr@{}}
        \toprule
        & & \multicolumn{3}{c}{\textbf{Zero-shot}} & \multicolumn{3}{c}{\textbf{Few-shot}} \\
        \textbf{Task} & \textbf{Effectiveness metric} & \textbf{Value} & \textbf{Latency (s)} & \textbf{Cost (\$)} & \textbf{Value} & \textbf{Latency (s)} & \textbf{Cost (\$)} \\
        \midrule
        Prompt filtering & F1 & .919 & 1.600 & .0229 & .943 & 1.537 & .0749 \\
        Source domain routing & Macro-F1 & .921 & 1.550 & .0190 & .924 & 1.571 & .0612 \\
        Elicited response score & Spearman $\rho$ & $-$.001 & 1.579 & .0188 & $-$.024 & 1.583 & .0524 \\
                                & Q1--Q4 AUC & .479 & --- & --- & .457 & --- & --- \\
                                & MAE & 1.723 & --- & --- & 1.608 & --- & --- \\
        \bottomrule
    \end{tabular}}
\end{table*}

\textbf{Effectiveness.} Table~\ref{tab:llm-baselines} shows that the LLM performs strongly on prompt filtering and source domain routing. Few-shot prompting raises filtering F1 from .919 to .943 but changes routing Macro-F1 only from .921 to .924. For elicited response scores, it reduces MAE from 1.723 to 1.608 but does not recover score ordering: Spearman $\rho$ remains indistinguishable from zero and Q1--Q4 AUC remains below .5. On the matched but non-paired samples, the LLM therefore exceeds the local models for routing, is only modestly stronger for filtering, and is substantially weaker for within-dataset quality-quartile discrimination.

\textbf{Efficiency.} All 3{,}000 baseline API outputs parse successfully after retry. The six 500-example runs consume 1{,}531{,}480 tokens and cost an estimated \$0.249, or \$0.000083 per evaluated example. Median API latency is 1.54--1.60\,s per prompt, about 270 times the 5.7\,ms embedding pipeline and 510 times the 3.0\,ms Syntactic pipeline. These ratios include network and provider time, whereas the local measurements do not, and the local figures omit hardware amortisation. The local methods therefore provide competitive filtering, stronger quality discrimination, interpretable features, and substantially lower latency without a per-query API call. The release includes the evaluation script, aggregate metrics, token counts, and cost calculations. Raw prompt text and per-example prediction records remain in git-ignored local caches and are not redistributed.

\end{document}